\documentclass[10pt,twocolumn,letterpaper]{article}
\usepackage[pagenumbers]{cvpr}

\usepackage{booktabs}
\usepackage[dvipsnames]{xcolor}

\definecolor{cvprblue}{rgb}{0.21,0.49,0.74}

\usepackage[pagebackref,breaklinks,colorlinks,citecolor=cvprblue]{hyperref}

\title{Rotated Lights for Consistent and Efficient 2D Gaussians Inverse Rendering}

\author{Geng Lin
\hspace{2em}
Matthias Zwicker \\
[0.4em]
\textnormal{University of Maryland, College Park} \\
[0.1em]
\url{https://rotlight-ir.github.io/}  
}

\newcommand{\gset}{\mathcal{G}}

\newcommand{\loss}{\mathcal{L}}
\renewcommand{\vec}[1]{\ensuremath{\mathbf{#1}}}

\makeatletter
\newcommand{\providelength}[1]{%
  \@ifundefined{\expandafter\@gobble\string#1}
   {% if #1 is undefined, do \newlength
    \typeout{\string\providelength: making new length \string#1}%
    \newlength{#1}%
   }
   {% else do nothing
   }%
}
\makeatother
\begin{document}

\maketitle

\begin{abstract}

Inverse rendering aims to decompose a scene into its geometry, material properties and light conditions under a certain rendering model. It has wide applications like view synthesis, relighting, and scene editing.
In recent years, inverse rendering methods have been inspired by view synthesis approaches like neural radiance fields and Gaussian splatting, which are capable of efficiently decomposing a scene into its geometry and radiance. They then further estimate the material and lighting that lead to the observed scene radiance.
However, the latter step is highly ambiguous and prior works suffer from inaccurate color and baked shadows in their albedo estimation albeit their regularization.
To this end, we propose RotLight, a simple capturing setup, to address the ambiguity.
Compared to a usual capture, RotLight only requires the object to be rotated several times during the process.
We show that as few as two rotations is effective in reducing artifacts.
To further improve 2DGS-based inverse rendering, we additionally introduce a proxy mesh that not only allows accurate incident light tracing, but also enables a residual constraint and improves global illumination handling.
We demonstrate with both synthetic and real world datasets that our method achieves superior albedo estimation while keeping efficient computation.

\end{abstract}
\section{Introduction}

Inverse rendering solves the decomposition of a scene into its geometry, surface material properties, and lighting, usually from a sequence of images of the scene.
It has wide applications like scene relighting and digital asset creation.

Usually, inverse rendering is achieved by implementing a differentiable forward rendering model, such as rasterization and physically-based path tracing.
Traditionally, differentiable path tracing is considered computationally infeasible until adjoint and path reusing approaches~\cite{NimierDavid2020Radiative,vicini2021prb} and neural radiance cache~\cite{Hadadan2023inverse} were proposed to significantly lower the cost. However, they still assume known geometry.
Recent developments in view synthesis methods have shred much light on inverse rendering research.
This is largely because recent view synthesis approaches like NeRF~\cite{mildenhall2020nerf}, SDF~\cite{wang2021neus} and 3D Gaussian Splatting~\cite{kerbl20233dgs} all decompose the scene into high-quality geometry and radiance representations efficiently.
Geometry is essential for inverse rendering, and the radiance field can be useful to model indirect illumination.

In particular, Gaussian splatting approaches \cite{kerbl20233dgs,huang20242dgs} have encouraged works such as GS-IR~\cite{liang2024gs}, RefGaussians~\cite{yao2025refGS}, and the state-of-the-art IRGS~\cite{gu2024IRGS}. These approaches are much faster than previous methods, due to the fast rasterization of Gaussian splatting. In particular, IRGS proposes a two-stage pipeline by first reconstructing the scene geometry and radiance with 2DGS, and then recovering the surface material and lighting using differentiable physically-based rendering.
To better handle global illumination, they propose to ray trace the 2D Gaussian primitives and use the integrated radiance as indirect illumination.

While IRGS achieve fast and state-of-the-art quality inverse rendering, the recovered albedo often still contains baked shadows and inconsistent shifts in colors. This is largely due to the ambiguity between light and surface color.
To this end, we propose to capture the object under several rotated environment lighting. In practice, this can be easily achieved by rotating the object itself and taking a set of photos at each rotation.
We demonstrate that with this simple addition, a significant portion of shadows and artifacts in the estimated albedo can be removed.

In order to handle multiple radiance fields from rotated lights, we use MLPs as scene radiance caches for indirect illumination queries. While directly ray-tracing incident rays on the 2D Gaussians, as proposed in IRGS, is an intuitive approach, we find that it is geometrically inconsistent and may harm incident light integration. To alleviate this, we propose to make use of an extracted mesh as a proxy geometry for accurate and efficient incident queries.

To further improve global illumination handling, we incorporate a residual constraint on the radiance cache, inspired by Inverse Neural Radiosity \cite{Hadadan2023inverse}. The constraint enforces the consistency between cached and rendered radiance and effectively improves albedo in regions covered in shadow where indirect lights dominate. To enable residual evaluation at arbitrary scene locations, we utilize the mesh to randomly sample on the scene surface. Our experiments show that baked shadows in the estimated albedo are further reduced with the residual constraint.

In summary, our contributions are:

\begin{enumerate}
    \item We propose \textit{RotLight}, a simple but effective capture setup to improve albedo estimation.
    \item We analyze the drawbacks of using 2DGS as a scene representation and propose mitigation with a proxy mesh.
    \item We incorporate the residual constraint into 2DGS-based inverse rendering to better handle global illumination effects.
\end{enumerate}

Our code and data will be released to the public.
\section{Related Works}

\subsection{Novel View Synthesis}

In recent years, neural radiance fields (NeRFs) \cite{mildenhall2020nerf} have gained popularity in novel view synthesis.
It predicts view-dependent colors and volume densities for each 3D location along the ray with multi-layer perceptrons (MLPs) and applies volume rendering \cite{kajiya1984ray} to render images from novel views.
However, NeRF-based methods suffer from slow training and rendering speeds due to the huge amount of MLP queries for volumetric integration, despite a good amount of effort~\cite{garbin2021fastnerf,yu2021plenoctrees,fridovich2022plenoxels,liu2020nsvf,muller2022instantngp,chen2023mobilenerf,SunSC22dvgo,yariv2023bakedsdf,reiser2023merf,reiser2021kilonerf} to improve NeRF rendering efficiency.
As a result, it remains challenging for the NeRF-based method to achieve real-time high-quality rendering.

Another method to represent scene geometry and radiance is using signed distance fields (SDFs)~\cite{wang2021neus,jeong2024esr,yariv2021volumerendering}.
These methods are capable of representing accurate and consistent geometry thanks to the regularization derived from their mathematical definitions of SDFs.
However, they require iteratively stepping along each ray and evaluating the SDF value to determine if a surface is hit as determined by a radius threshold.
Often, many small steps per pixel are required ensure accuracy, making it computationally expensive. Similarly, UniSURF \cite{oechsle2021unisurf} uses expensive root finding to extract surfaces.

Finally, an increasingly popular branch of works utilize the point-based rendering model~\cite{wiles2020synsin,yifan2019differentiable,aliev2020neural,kopanas2021point,ruckert2022adop,ost2022neural,xu2022point,lassner2021pulsar}, which developed combinations of differentiable point-based rendering potentially and learned neural networks. More recently, 3D Gaussian Splatting~\cite{kerbl20233dgs} has attracted much attention. It represents a scene using anisotropic 3D Gaussian primitives, and achieves a new state-of-the-art in both rendering efficiency and quality with its efficient tile-based rasterization method. However, the rasterizer makes approximations that result in multi-view inconsistencies in its renderings, often referred to as \textit{popping artifacts}. StopThePop~\cite{radl2024stopthepop} proposes a more accurate sorting algorithm to mitigate the issue, and 3D Gaussian Ray Tracing~\cite{moenne20243dgsrt} implements a ray tracing-based renderer of 3D Gaussian primitives that achieves multi-view consistent renderings at the cost of slower speed. Nevertheless, their abilities to represent accurate geometry are still limited by the lack of a good depth definition for 3D Gaussian primitives. To this end, SuGaR~\cite{guedon2024sugar} regularizes Gaussians to align with the surface and a method to extract meshes from the Gaussians using Poisson reconstruction. 2D Gaussian Splatting~\cite{huang20242dgs} further replaces 3D Gaussian volumes with 2D Gaussian disks, resulting in scene reconstruction with consistent depths, from which high-quality meshes can be extracted with truncated signed distance fields~\cite{brian96volume}.

In our work, we leverage 2D Gaussians for its accuracy in geometry representation, which is crucial for inverse rendering. In addition, we make use of the mesh extraction technique to obtain a less accurate but more consistent model for better incident light tracing.

\subsection{Inverse Rendering}

Inverse rendering can be achieved by implementing a differentiable rendering pipeline and running gradient descent on the material parameters. However, as physically-based rendering pipelines often rely on path tracing, they can have prohibitive computational requirements for complicated problems.
Adjoint approaches~\cite{NimierDavid2020Radiative,vicini2021prb} reduce the cost by propagating gradients and reusing paths. Inverse Neural Radiosity (InvNeRad)~\cite{Hadadan2023inverse} proposes to use a neural network as a radiance cache to record incident lights everywhere, with a residual constraint to reduce bias. These methods are accurate, but require known geometry and light.

Recent developments of inverse rendering are closely related and inspired by the advancements in novel view synthesis, as it can be seen as a downstream task that relies on scene geometry being available.
In addition, recent view synthesis methods often include certain forms of radiance fields that are convenient to use as radiance caches.
For example, NeRF inspired NeRV~\cite{srinivasan2021nerv}, NeRD~\cite{boss2021nerd}, Ref-NeRF~\cite{verbin2022refnerf}, and many more~\cite{boss2022samurai,yao2022neilf,liu2023nero,jin2023tensoir,yang2023sireir,attal2024flash}. Similarly, signed distance fields have been used for modeling geometry in NeILF++~\cite{zhang2023neilf++}, InvRender~\cite{zhang2022modeling}, and WildLight~\cite{cheng2023wildlight}.

Gaussian splatting is no exception in being integrated into inverse rendering pipelines~\cite{liang2024gs,shi2025gir,guo2024prtgs,yao2025refGS,gu2024IRGS}. It is particularly preferred because of its fast convergence and rendering speed compared to NeRF and SDF-based methods. The point-based nature of Gaussian splatting also makes it straightforward to attach material properties as additional data to every Gaussian primitive, resulting in natural and fast rendering of material maps. However, a major drawback of using Gaussians as a radiance field is that the splatting rasterization is image-centric and does not provide an easy method to trace incident radiance of arbitrary rays in space. It prevents prior works from properly handling global illumination effects and limits their performance.

To alleviate this weakness, IRGS~\cite{gu2024IRGS} proposes to incorporate global illumination by tracing 2D Gaussian primitives as incident light samples, effectively using them as a radiance cache. However, their work still suffers from unwanted effects like baked shadows and color shifts, due to the ambiguous nature of the problem.

Our work takes a step in resolving the ambiguity by capturing objects with multiple rotations, creating observations of the object under different but coherent lighting. This scheme introduces little additional effort, yet is effective in improving the consistency of the estimated albedo. We also propose a proxy mesh and include a residual constraint to further improve the accuracy of material estimation.

\section{Background}

Our inverse rendering pipeline consists of two stages: \textit{scene reconstruction} that uses novel view synthesis approaches to model the scene with a geometry field and a radiance field, and \textit{material decomposition} that estimates material and environment light given the model from the first stage. This section covers the related background of both stages.

\subsection{3D Gaussian Splatting}

Given a set of images $\{I_i\}$ and their associated camera parameters $\{P_i\}$, Kerbl et al.~\cite{kerbl20233dgs} propose to model the scene with 3D Gaussian primitives $\gset = \{G_i\}$. Each Gaussian is parameterized by its 3D location $\mu_i\in \mathbb{R}^3$, rotation $q_i\in\mathbb{R}^4$, and scale $s_i\in\mathbb{R}^3$. For view-dependent color rendering, it is also associated with an opacity value $\sigma_i$ and a set of spherical harmonics (SH) parameters. To render the scene from a camera, all 3D Gaussians are first projected onto the image plane using the camera view matrix $W$ and an affine transformation $J$ that approximates the camera projection. A 3D Gaussian's covariance matrix $\Sigma$ and the projected $\Sigma'$ are (dropping index for clarity):

\begin{align}
\Sigma &= RSS^TR^T \\
\Sigma' &=JW\Sigma W^TJ^T
\end{align}

where $S$ and $R$ are the scale and rotation matrices derived from $s$ and $q$, respectively. The projected Gaussian on the image plane is obtained by skipping the third row and column of $\Sigma'$~\cite{zwicker2001ewa}.

3DGS then renders each pixel $\vec{x}$ by alpha blending all projected Gaussians evaluated at that pixel:

\begin{align}
\vec{c} &= \sum_{i=1}^{N}{\left(c_i\sigma_i G_i(\vec{x})\prod_{j=1}^{i-1}{(1-\sigma_j G_j(\vec{x})})\right)},  \label{eq:gauss_render} \\
G_i(\vec{x}) &= \exp\left({-\frac{1}{2}(\vec{x}-\vec{x}_i)^T{\Sigma'}^{-1}(\vec{x}-\vec{x}_i)}\right)
\end{align}

where the Gaussian are sorted by the distances between their centers and the camera. $\vec{x}_i$ is the projected pixel location of the Gaussian center, and $c_i$ is the view-dependent color of the Gaussian derived from its SH parameters evaluated at the direction from its center to the camera.

\subsection{2D Gaussian Splatting}

Although 3D Gaussians provide good image reconstruction quality, the method imposes challenges to surface geometry reconstruction due to the ambiguity of depth. To this end, Huang et al.~\cite{huang20242dgs} propose to replace the primitives with 2D Gaussians, which are similar to 3D Gaussians, except that their scale in the local z direction is zero. Formally, a 2D Gaussian centered at $\mu$ lies on a plane defined by two orthogonal unit vectors $\vec{t}_u$ and $\vec{t}_v$. Its normal is then $\vec{n}=\vec{t}_u\times \vec{t}_v$. These three vectors form a rotation matrix $R=[\vec{t}_u, \vec{t}_v, \vec{n}]$ that characterizes the orientation of the primitive. Similarly to 3DGS, the variance of the Gaussian along these directions is defined by a scale vector $s=(s_u,s_v,0)$. Then, a point lying on the plane can be parameterized in the Gaussian's local space:

\begin{align}
    P(u,v)=\mu+s_u\vec{t}_u u+s_v\vec{t}_v v.
    \label{eq:gauss_point}
\end{align}

The value of the Gaussian at the point is the standard Gaussian:

\begin{align}
    G(u,v)=\exp\left(-\frac{u^2+v^2}{2}\right).
    \label{eq:gauss_func}
\end{align}

To rasterize a 2D Gaussian, they follow the formulation of~\cite{zwicker04perspective} to approximate the projection between Gaussian space and screen space points, and further derive an efficient method to solve for $(u,v)$ given image plane $\vec{x}$. To compose the color of a pixel from multiple Gaussians, 2DGS follows the same alpha blending formulation as Equation \ref{eq:gauss_render}. Note that the depth and normal of a 2D Gaussian intersecting a ray is uniquely defined, and the depth and normal of the pixel can be composited in the same way.

\subsection{Gaussian Splatting in Inverse Rendering}

With the scene geometry and radiance recovered from the reconstruction methods, inverse rendering poses a further challenge to decompose the radiance into surface materials and environment lighting.
Similarly to IRGS~\cite{gu2024IRGS}, we use a physically based rendering model. To render a pixel, its camera ray $\vec{r}(t)$ towards direction $\vec{d}$ is defined by the camera parameters, and a surface point $\vec{x}$ can be determined by using the depth map value $d$ from splatting, i.e. $\vec{x}=\vec{r}(d)$. The color of the pixel is the outgoing radiance at $\vec{x}$ towards $\omega_o=-\vec{d}$, defined as:

\begin{equation}
    L_o(\vec{x},\omega_o)=\int_{\Omega}f(\vec{x},\omega_i,\omega_o)L_i(\vec{x},\omega_i)(\omega_i\cdot \vec{n})\text{d}\omega_i,
    \label{eq:render}
\end{equation}

where $\omega_i$ is the incident direction over the hemisphere $\Omega$ defined by the surface normal $\vec{n}$ at point $\vec{x}$, $L_i$ is the incident radiance function, and $f$ is the bidirectional reflectance distribution function (BRDF).

We follow prior works and use a simplified Disney BRDF model that has two parameters: albedo $\vec{a}\in \mathbb{R}^3$ and roughness $r\in \mathbb{R}$. The parameters are attached to every Gaussian primitive and are rendered with the same method as color and depth.

To render an image, we first splat the 2D Gaussians with the 2DGS rasterizer to obtain maps of per-pixel depth, normal, albedo, and roughness. For every pixel, its corresponding 3D point is derived from its depth $d$ by $\vec{x}=\vec{r}(d)$. Then, a set of incident directions $\{\omega_i\}$ is randomly sampled to compute $L_i(\vec{x},\omega_i)$ and evaluate Equation \ref{eq:render} with Monte Carlo integration.

The computation of $L_i(\vec{x},\omega_i)$ can be challenging, as it requires tracing incident rays from $\vec{x}$ to determine if it hits scene geometry or the environment lighting.
However, 2DGS rasterization is not built for integrating individual rays in arbitrary directions.
To this end, IRGS~\cite{gu2024IRGS} proposes 2D Gaussian Ray Tracing, which follows 3DG-RT~\cite{moenne20243dgsrt} to utilize Nvidia Optix for efficient implementation. The ray tracing operator $\text{Ray-Trace}(\vec{x}, \vec{d})$ integrates the 2D Gaussian primitives along the ray and composite the same quantities as rasterization, including color $\vec{c}_{\text{rt}}$ and alpha $\alpha_{\text{rt}}$.
Note that the integrated color corresponds to the \textit{indirect illumination} towards $\vec{x}$. Intuitively, ray tracing allows the use of the 2D Gaussians as a cache of the scene radiance. Environment visibility can be derived as $(1-\alpha)$, and the \textit{direct illumination} can be queried from the optimizable environment map $\textbf{E}(\omega_i)$. The value of $L_i$ is then the sum of both:
\begin{equation}
    L_i(\vec{x},\omega_i)=\vec{c}_{\text{rt}} + (1-\alpha_{\text{rt}})\textbf{E}(\omega_i).
    \label{eq:li}
\end{equation}

\section{Method}

\begin{figure*}[t]
\centering

\includegraphics[width=0.9\textwidth]{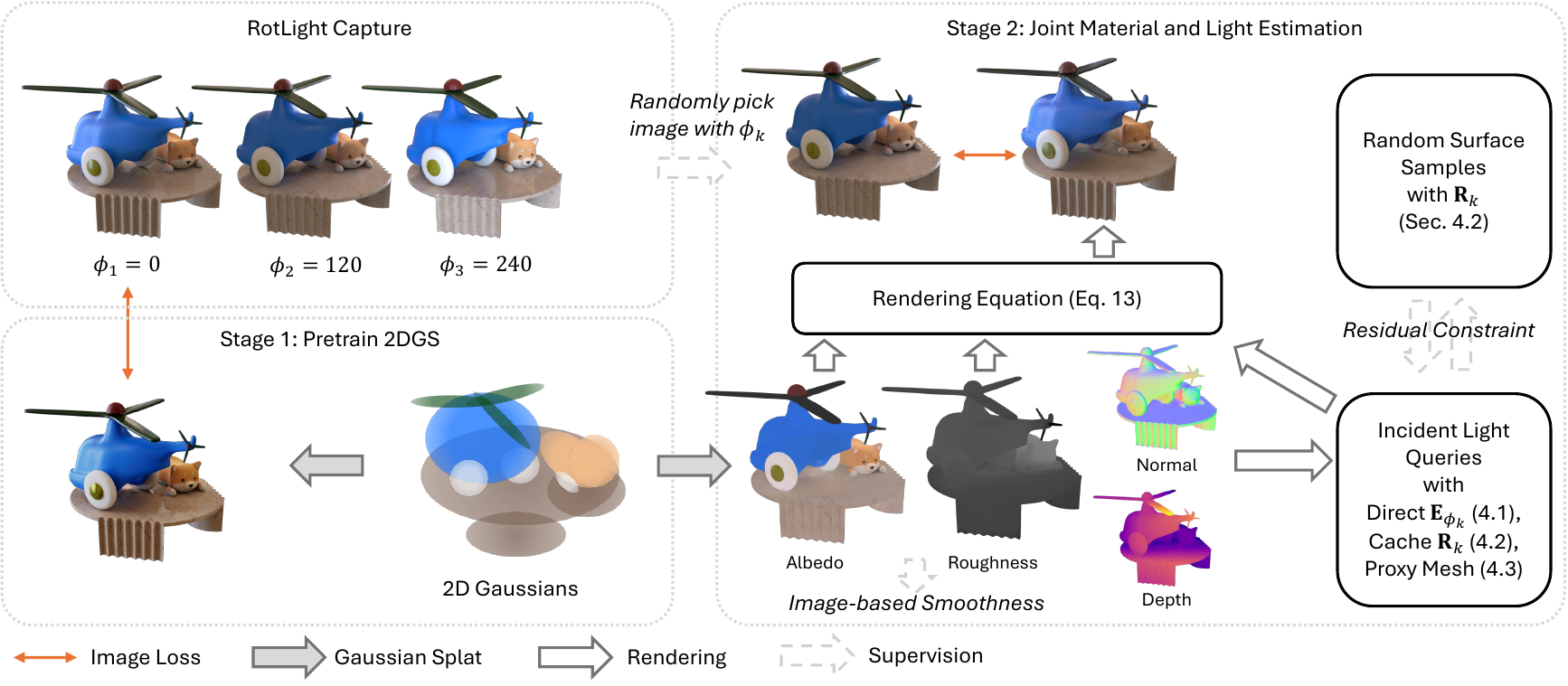}
\caption{Overview of our pipeline. \textbf{RotLight:} we capture the same object with $K=3$ different rotations. \textbf{Stage 1:} we pretrain a 2DGS model with a set of images under the same lights (e.g. $\phi_1$) to obtain an initial geometry reconstruction of the scene. \textbf{Stage 2:} we obtain pixel material, depth and surface normal by splatting Gaussians. The depth and normal are used to query incident lights, with direct illumination from the environment map $\textbf{E}_{\phi_k}$ (Section \ref{sec:method_rotlight}), and indirect illumination from a radiance cache (Section \ref{sec:method_radiance_cache}). The radiance cache is further supervised with a residual constraint evaluated at random surface points. For accurate incident sampling, we also use a proxy mesh instead of ray tracing the Gaussians (Section \ref{sec:method_incmesh}).}
\label{fig:inverse_overview}
\end{figure*}

Our proposed inverse rendering approach consists of a simple yet effective capturing setup, \textit{RotLight}, and a two-stage reconstruction pipeline. \textit{RotLight} captures an object under rotated lighting, reducing the ambiguity of the problem (Section \ref{sec:method_rotlight}). During scene reconstruction, we first train a 2D Gaussian model to represent the scene geometry. Similarly to previous work~\cite{gu2024IRGS,yao2025refGS}, the Gaussian primitives are then augmented with material properties $\vec{a},r$ to represent the surface material. Then, jointly estimate surface material properties and environment lighting. To handle global illumination effects, we use a neural radiance cache represented by several MLPs (Section \ref{sec:method_radiance_cache}). Finally, we propose a proxy mesh geometry for consistent and efficient incident radiance queries (Section \ref{sec:method_incmesh}).

\subsection{RotLight Capturing}
\label{sec:method_rotlight}

Although prior work \cite{gu2024IRGS} achieves plausible material reconstruction by taking global illumination effects into account, they still suffer from the ambiguity of the problem.
As every observed color can be explained by an arbitrary combination of albedo color, roughness, direct illumination, and indirect illumination, the optimization process often converges to sub-optimal results despite regularization efforts.
To this end, we propose a simple modification to the capturing setup that drastically reduces the amount of ambiguity and improves albedo recovery.
First, the object is captured as usual.
Then, we rotate the object by a predefined angle and capture an additional set of images.
The process can be repeated as needed, and we use two rotations of 120 degrees in our experiments.
We call this approach \textit{RotLight}, referring to the fact that the light on the object is effectively rotated.

In this capture setup, each image in the image set is also tagged with an environment rotation angle $\phi_k$. When querying direct illumination of the direction $\omega_i$ for a pixel, the angle of its image is taken into account, and $\vec{E}_{\phi_k}(\omega_i)$ is obtained by rotating the queried direction along the up axis:

\begin{align}
    \vec{E}_{\phi_k}(\omega_i) &= \vec{E}(R_{\phi_k}w_i), \text{where} \\
    R_{\phi_k} &= \begin{pmatrix}
\cos(\phi_k) & 0 & \sin(\phi_k) \\
0 & 1 & 0 \\
-\sin(\phi_k) & 0 & \cos(\phi_k) \\
\end{pmatrix}.
\end{align}

\subsection{Neural Radiance Cache with Residual Constraint}
\label{sec:method_radiance_cache}

The idea of modeling indirect illuminations with neural networks has been seen in several inverse rendering approaches~\cite{Hadadan2023inverse,zhang2023neilf++,gu2024IRGS}. In particular, IRGS effectively uses the 2D Gaussians as the cache whose radiance information is trained in the first stage. They query the cache with the $\text{Ray-Trace}$ operator that integrates the radiance along the ray and use the integrated color $\vec{c}_{\text{tr}}$ for indirect illuminations.

The idea of using a neural radiance cache in forward rendering was first proposed in Neural Radiosity (NeRad)~\cite{Hadadan2021NeRad}.
Their cache is modeled with a spatial feature grid and a multi-layer perceptron (MLP) decoder.
It records radiance everywhere in the scene, so that lengthy path-tracing operations can be replaced by a single query to the cache.
Importantly, NeRad proposes a \textit{residual constraint} to minimize the following difference, when tracing a ray $\text{Ray-Trace}(\vec{x},\vec{d})$:

\begin{equation}
    \loss_{\text{residual}} = \left| \vec{c}_{\text{tr}}, L_o(\vec{y},-\vec{d}) \right|,
\end{equation}

where $\vec{c}_{\text{tr}}$ is the cached value from ray tracing, $\vec{y}$ is the hit point of the ray, and $L_o$ is defined in Equation \ref{eq:render}.
Intuitively, the constraint requires that the \textit{cached radiance} to match the \textit{rendered radiance}.
Without this constraint, other works~\cite{zhang2023neilf++,gu2024IRGS} constrain their radiance cache (SDF and 2D Gaussians) solely with input images and camera rays.
As shown in InvNeRad~\cite{Hadadan2023inverse}, the residual loss improves cache queries from unseen locations and directions and makes the handling of global illumination more accurate.

In this work, we use MLPs to model the radiance cache, instead of re-using the 2D Gaussians like IRGS.
Although the residual constraint can be applied to the Gaussians-based cache, we made this choice for the following considerations:

\begin{enumerate}
    \item With our \textit{RotLight} capture, a cache is required for each unique light angle.
    It is faster in computation and simpler in implementation to train several MLPs than to optimize several sets of Gaussians.
    \item For a given point and direction in space, querying an MLP is more efficient than tracing the Gaussians.
\end{enumerate}

For every light angle $\phi_k$ in the captured images, we define a radiance cache $\textbf{R}_{k}(\vec{x},\vec{d})$. The indirect illumination at a point $\vec{x}$ in direction $\omega_i$ is found by:

\begin{equation}
    \vec{c}_{\text{tr}}=\textbf{R}_{k}(\vec{y},-\omega_i),
    \label{eq:mlp_query}
\end{equation}

where $\vec{y}$ is the hit point of the ray from $\vec{x}$ towards $\omega_i$. We discuss how to find $\vec{y}$ in Section~\ref{sec:method_incmesh}.

\subsubsection{Sampling Scene Geometry}

To evaluate residual loss $\loss_{\text{residual}}$, random surface points and directions need to be sampled from the scene.
However, since depths of 2DGS are defined with volumetric integration, individual Gaussian primitives do not provide explicit information about surface locations.
Instead, surface information is encoded with depth maps, which has been shown in 2DGS~\cite{huang20242dgs} to be capable of supporting high-quality mesh extraction with TSDF~\cite{brian96volume}.
Therefore, we choose to extract a mesh in the same manner.
Then, at each optimization step, we uniformly randomly choose $N_{\text{residual}}$ triangles from the mesh. For each triangle, we pick a random point and a random view direction over the hemisphere defined by its normal.

While the mesh should theoretically be updated periodically during the second stage to reflect geometry changes, we find empirically that the changes in geometry are often very small and do not require the mesh to be regenerated.

\subsection{Proxy Mesh for Incident Queries}
\label{sec:method_incmesh}

For an arbitrary ray from a scene surface $\vec{x}$ in direction $\vec{d}$, we need to find its intersection point $\vec{y}$ to query the MLP radiance cache in Equation \ref{eq:mlp_query}.
We propose to use the extracted mesh discussed previously for incident ray queries.
For efficient queries, we implement an Optix-based ray tracing extension for Python, which reports the ray distance, hit triangle index, and barycentric coordinate.
Compared to the naive alternative of re-using IRGS's 2D Gaussian ray tracer, the mesh is not only more view consistent but also faster for incident queries, since they are opaque and do not require integration.
Our approach, as demonstrated in Figure \ref{fig:ambient}, results in accurate and smooth ambient occlusion renderings.

One catch of this approach is the primary hit points of camera rays, where the pixel colors are computed. For a camera ray, its hit point $\vec{x}$ from splatting may not be exactly on the mesh, causing issues in subsequent incident ray tracing. We use a simple workaround to also trace the ray with the mesh and obtain an alternative hit point $\vec{x}_m$. Then, Equation~\ref{eq:render} is modified to find incident lights for $\vec{x}_m$:

\begin{equation}
        L_o(\vec{x},\omega_o)=\int_{\Omega}f(\vec{x},\omega_i,\omega_o)L_i(\vec{x}_m,\omega_i)(\omega_i\cdot \vec{n})\text{d}\omega_i,
    \label{eq:render_incmesh}
\end{equation}

\subsubsection{Comparing 2D Gaussian Ray Tracing}

Since we use 2D Gaussians to model the scene, an obvious alternative is to follow IRGS: ray-trace the Gaussian primitives and obtain per-pixel depth.
However, this method suffers from inconsistencies when the Gaussians are rendered from different views.
This issue of Gaussian splatting has also been discussed in prior work~\cite{radl2024stopthepop}.
In addition, as Gaussians form thick clouds on model surfaces, the origin of an incident ray must be shifted along its direction by a significant amount.
This amount can be tricky to handle and will affect the accuracy of incident queries.
We demonstrate this problem by rendering ambient occlusion images that show the amount of direct light that is deemed visible at every point during rendering.
In Figure \ref{fig:ambient}, it can be seen that ray-tracing 2D Gaussians results in visible artifacts in ambient occlusion.

\subsection{Optimization Objective}

In summary, our optimization objective is to minimize the following losses.

\begin{itemize}
    \item $\loss_{data}=\left|L_o(\vec{x},\omega_o) - \vec{c}_{\text{gt}} \right|$, the difference between the rendered color and the ground truth color.
    \item $\loss_{\text{cache}}=\left|\textbf{R}_{k}(\vec{x},\omega_o) - \vec{c}_{\text{gt}} \right|$, which optimizes the radiance cache MLPs for camera rays.
    \item $\loss_{\text{residual}}$, the residual loss discussed in Section \ref{sec:method_radiance_cache}.
    \item $\loss_{\text{reg}}$, regularization loss terms that include mask entropy, albedo smoothness, light smoothness, roughness smoothness, and light whiteness. All of these regularization terms follow IRGS \cite{gu2024IRGS}.
\end{itemize}

So, the final loss is the sum:

\begin{equation}
    \loss=\loss_{data}+\loss_{\text{cache}}+\lambda_{\text{residual}}\loss_{\text{residual}}+\loss_{\text{reg}}
\end{equation}

In all experiments, we use $\lambda_{\text{residual}}=10$.

\section{Experiments}

\subsection{SynRotLight Dataset}

\providelength\width
\setlength\width{1.8cm}

\providelength\oldtabcolsep
\setlength{\oldtabcolsep}{\tabcolsep}
\setlength{\tabcolsep}{1pt}

\begin{figure}[t]
\centering
\footnotesize

\begin{tabular}{cccc}
Counter & Toy & Table & Hotdog \\
\midrule
\includegraphics[width=\width]{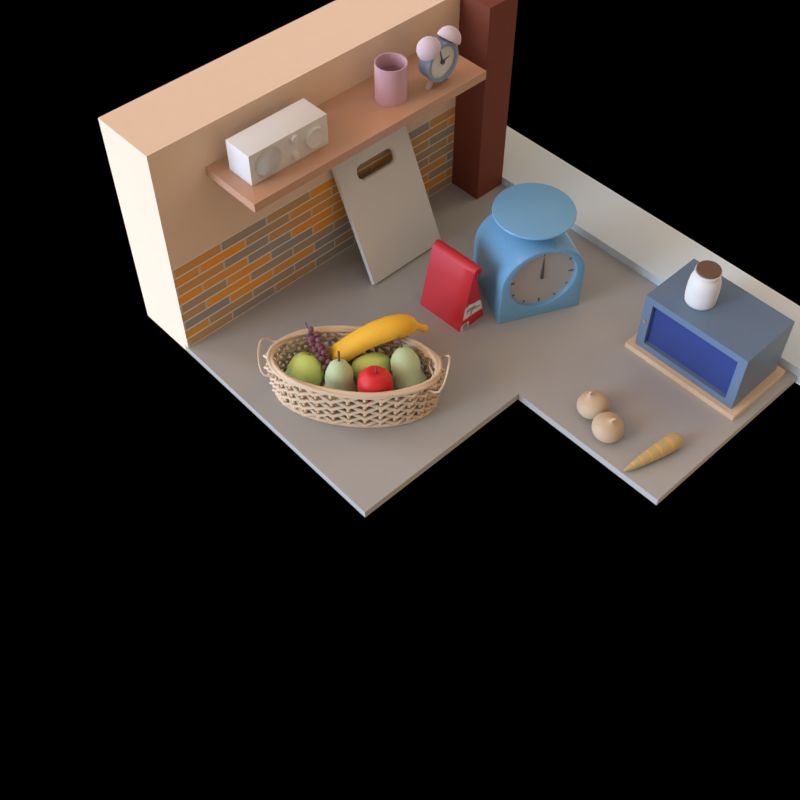} &
\includegraphics[width=\width]{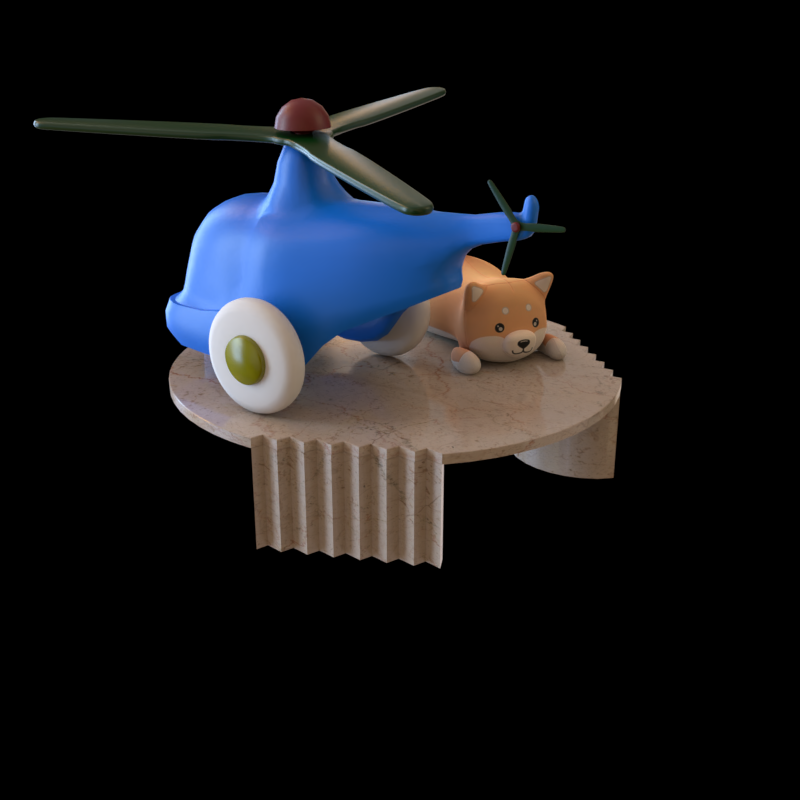} &
\includegraphics[width=\width]{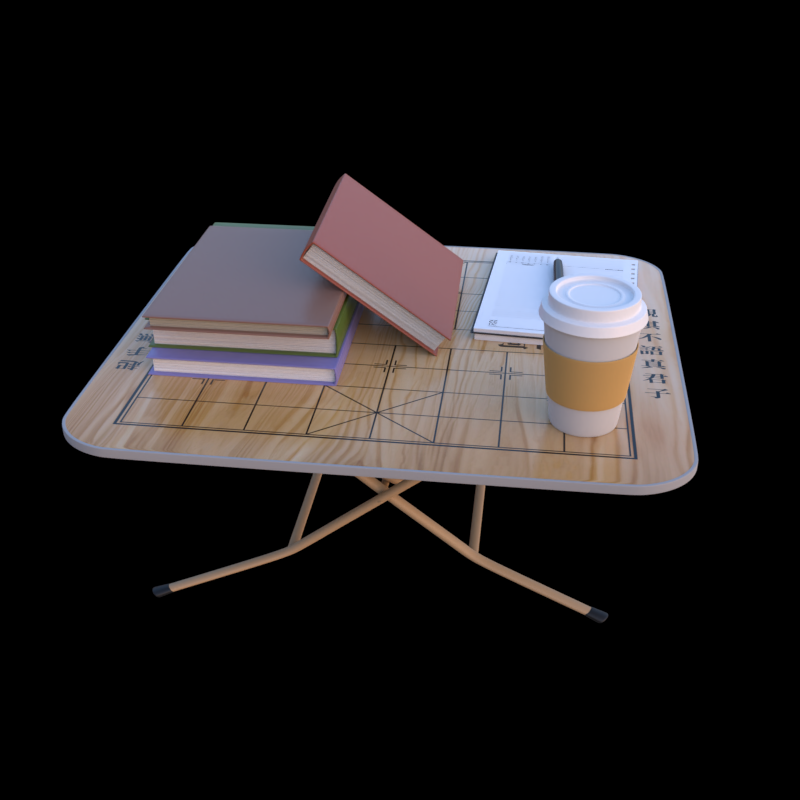} &
\includegraphics[width=\width]{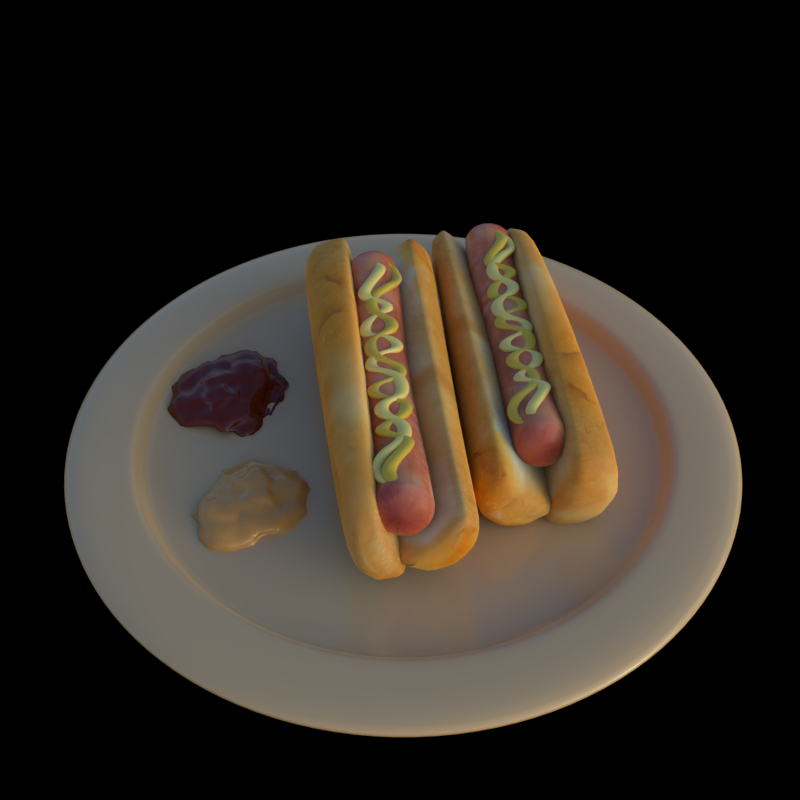} \\
\includegraphics[width=\width]{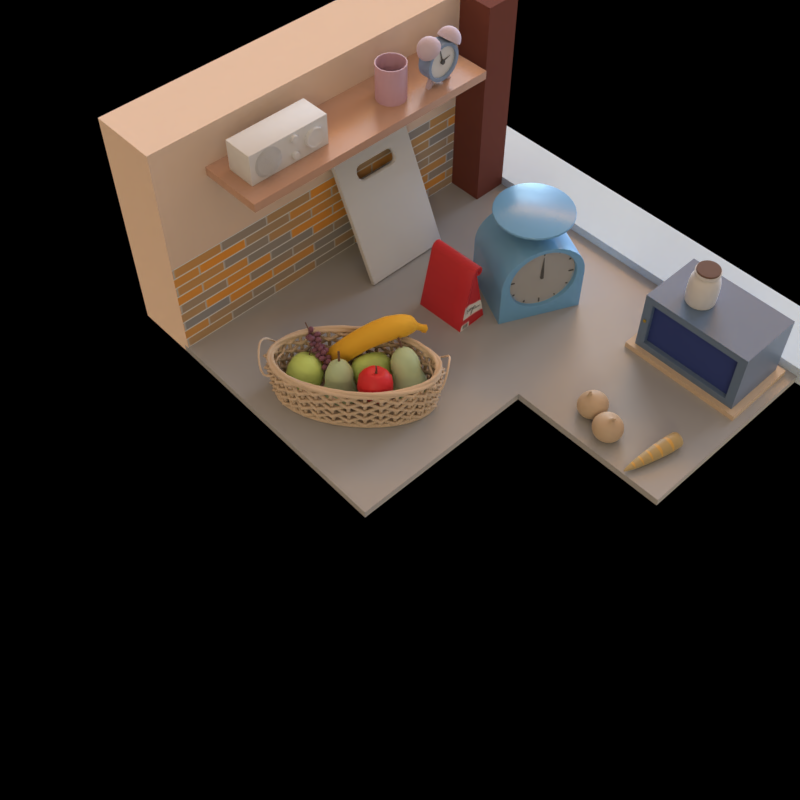} &
\includegraphics[width=\width]{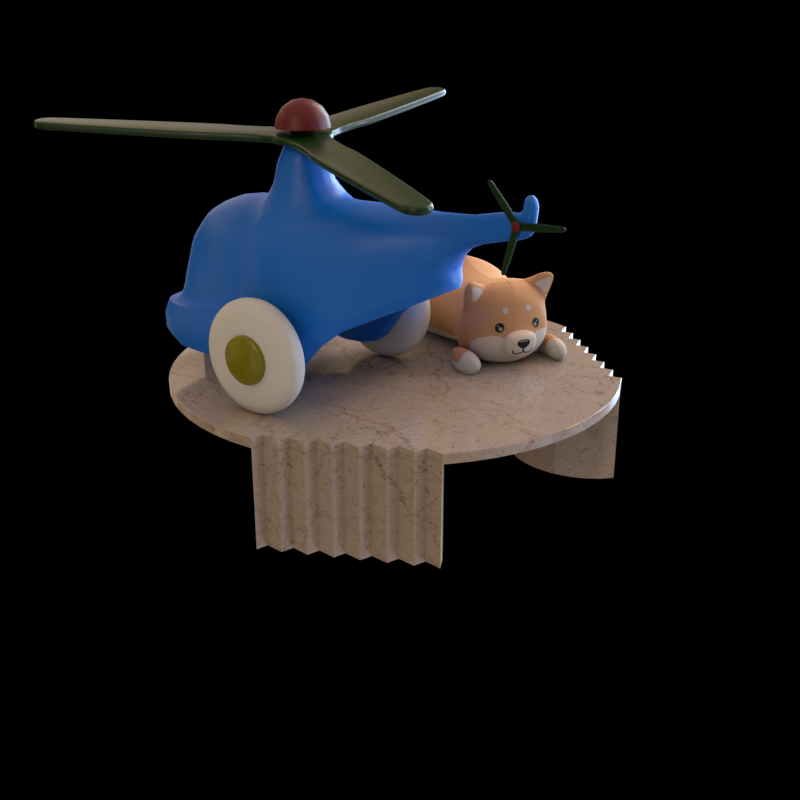} &
\includegraphics[width=\width]{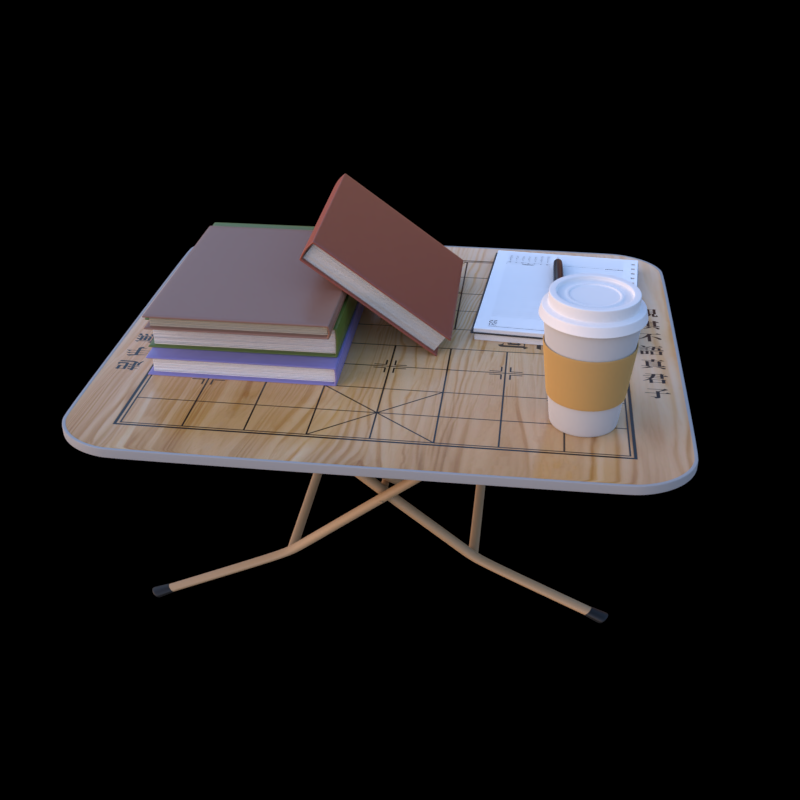} &
\includegraphics[width=\width]{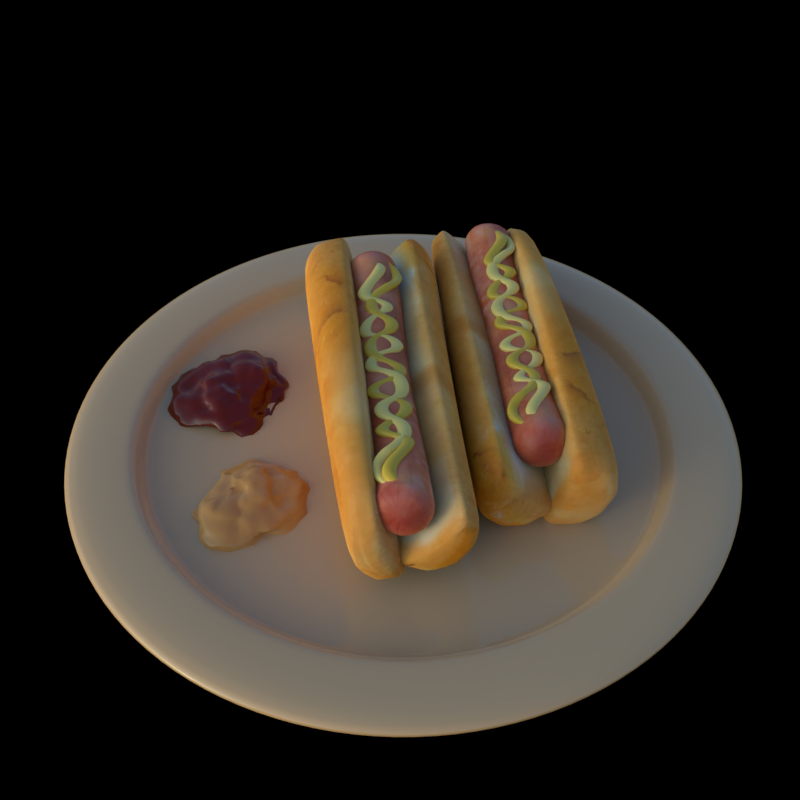} \\
\includegraphics[width=\width]{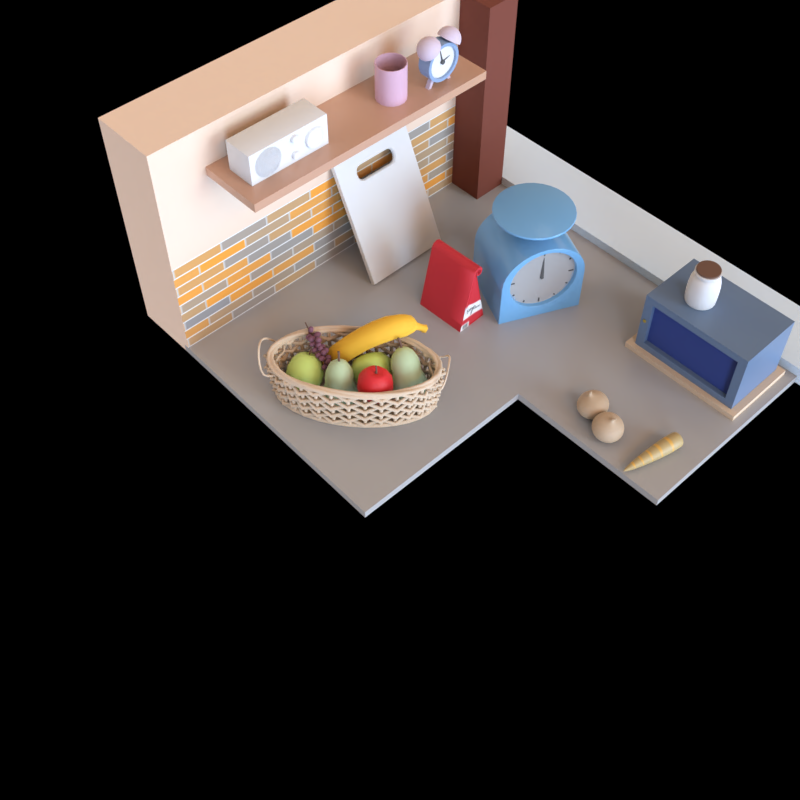} &
\includegraphics[width=\width]{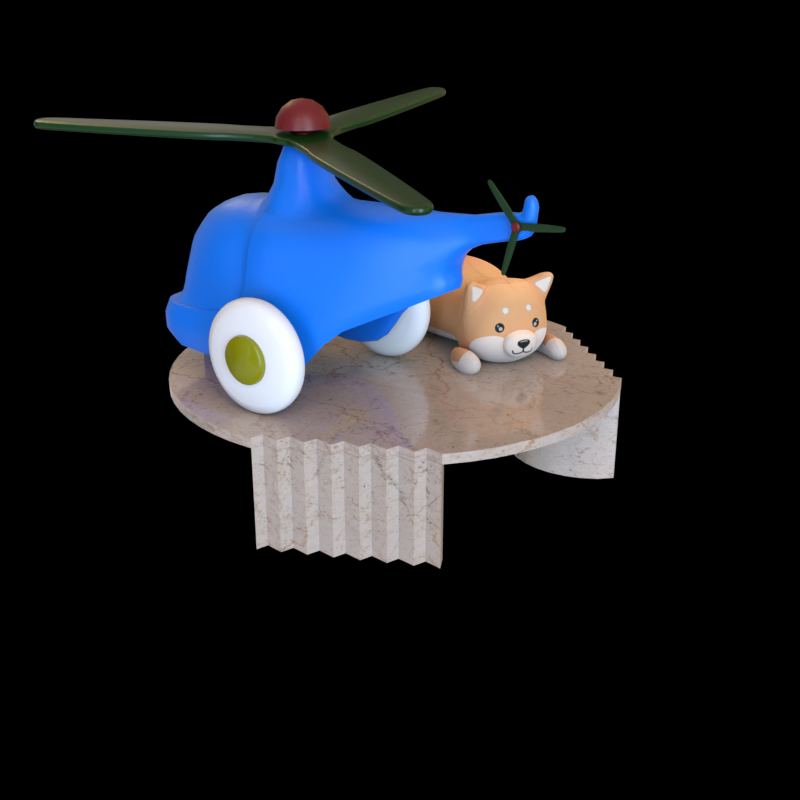} &
\includegraphics[width=\width]{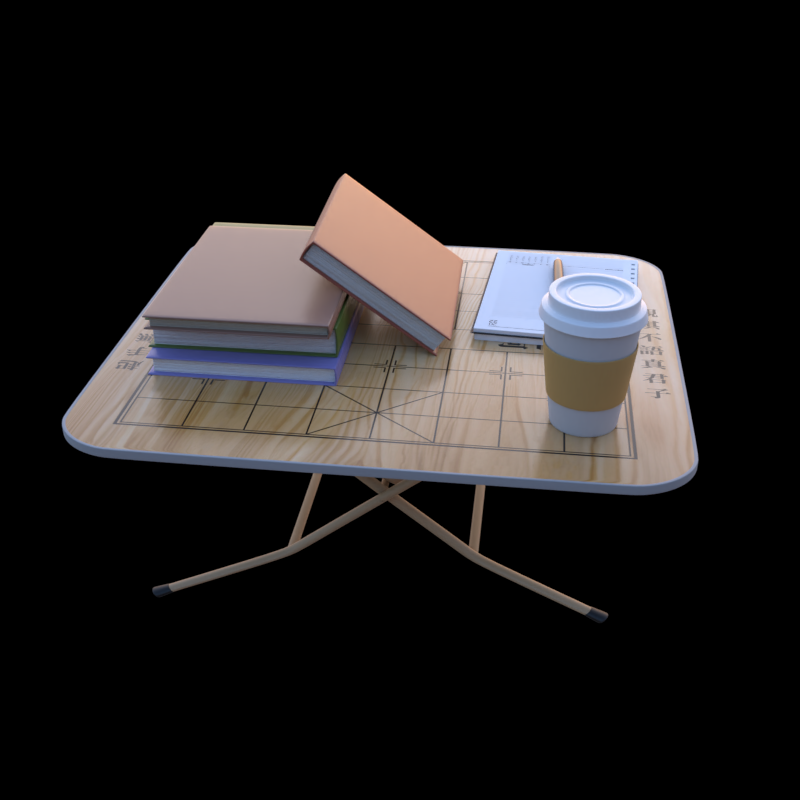} &
\includegraphics[width=\width]{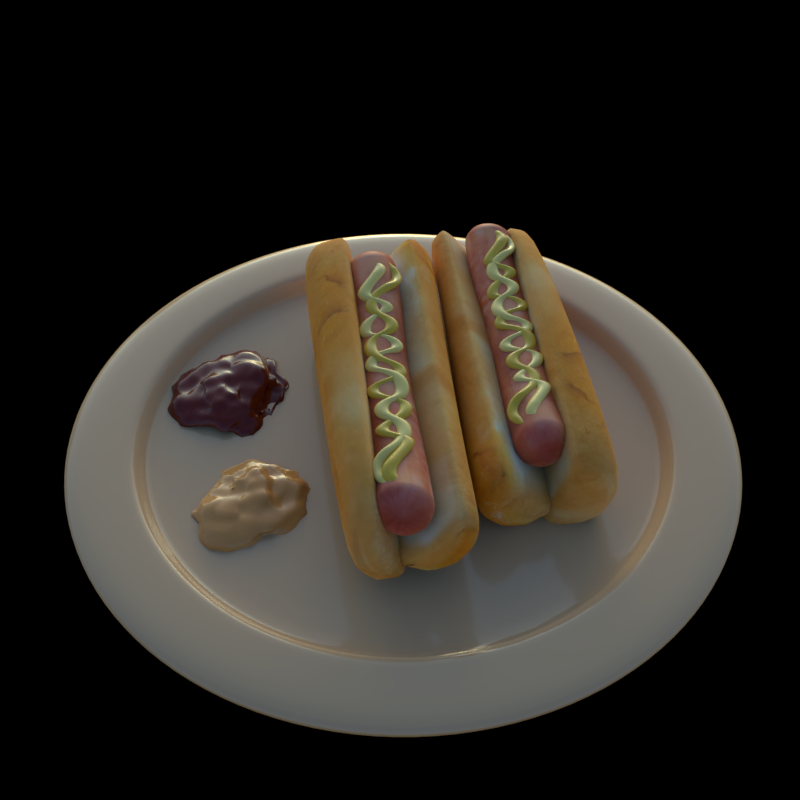}
\end{tabular}

\caption{Examples from our \textit{RotLight} dataset. The three rows are light rotations at 0, 120 and 240 degrees respectively.}
\label{fig:inverse_dataset_examples}

\end{figure}

\setlength\tabcolsep{\oldtabcolsep}

To perform fair quantitative evaluations of our proposed method, we create and release a dataset \textit{SynRotLight}, examples shown in Figure \ref{fig:inverse_dataset_examples}. It consists of four object-centric scenes with varying complexity, each containing 100 training views and 50 test views. The scenes were created and rendered with Blender. The model of the \textit{hotdog} scene is modified from the NeRF dataset, while the rest are adopted from online royalty free and CC0 resources. Following NeRF~\cite{mildenhall2020nerf}, the images have resolution 800$\times$800.

\subsection{Implementation}

We implement our pipeline with PyTorch based on 3DGS, RefGaussians, and IRGS.
In addition, to achieve fast ray tracing on meshes, we implement a PyTorch extension with Nvidia Optix. For a fair comparison, our method shares the same first stage 2DGS model with IRGS, and our second stage is trained for the same number of iterations. Our radiance cache MLPs are pre-trained for 20,000 steps regardless of the number of lights. Details are in the appendix.

\begin{table*}[t]
\footnotesize
\centering
\caption{Quantitative comparison of material estimation. All experiments are conducted in linear color space. \textbf{Top:} our method outperforms prior works by a large margin thanks to the capturing setup. \textbf{Bottom:} ablation studies show that our method gains the most improvement from \textit{RotLight}.}
\label{tab:main_quant}

\resizebox{0.9\textwidth}{!}{

\begin{tabular}{l|ccc|ccc|ccc|ccc}
\toprule
 &
 \multicolumn{3}{c|}{Hotdog} &
 \multicolumn{3}{c|}{Table} &
 \multicolumn{3}{c|}{Toy} &
 \multicolumn{3}{c}{Counter} \\
 &
 \multicolumn{2}{c}{Albedo} & Roughness &
 \multicolumn{2}{c}{Albedo} & Roughness &
 \multicolumn{2}{c}{Albedo} & Roughness &
 \multicolumn{2}{c}{Albedo} & Roughness \\
Method & PSNR & SSIM & MSE & PSNR & SSIM  & MSE & PSNR & SSIM  & MSE & PSNR & SSIM  & MSE \\
\midrule
GS-IR &
18.039 & 0.858 & 0.111 & 22.165 & 0.925 & 0.084 & 23.556 & 0.933 & 0.096 & 18.971 & 0.906 & 0.053\\
RefGS &
17.282 & 0.793 & 0.017 & 15.960 & 0.832 & 0.008 & 15.107 & 0.834 & 0.007 & 15.978 & 0.842 & 0.056\\
IRGS &
25.157 & 0.931 & 0.017 & 25.667 & 0.952 & 0.009 & 30.524 & 0.930 & 0.008 & 21.971 & 0.943 & 0.053\\
Ours &
29.227 & 0.952 & 0.015 & 28.621 & 0.962 & 0.008 & 34.073 & 0.937 & 0.007 & 27.098 & 0.965 & 0.019\\
\midrule
w/o residual &
29.037 & 0.952 & 0.015 & 28.529 & 0.962 & 0.008 & 33.368 & 0.936 & 0.008 & 27.071 & 0.966 & 0.018\\
w/o proxy mesh &
27.226 & 0.946 & 0.014 & 28.329 & 0.961 & 0.007 & 33.083 & 0.936 & 0.007 & 26.560 & 0.965 & 0.015\\
w/o \textit{RotLight} &
25.357 & 0.930 & 0.018 & 26.289 & 0.954 & 0.009 & 31.774 & 0.932 & 0.008 & 22.907 & 0.944 & 0.048\\
\bottomrule

\end{tabular}

}

\end{table*}

\begin{figure*}[t]
\centering

\includegraphics[width=0.8\textwidth]{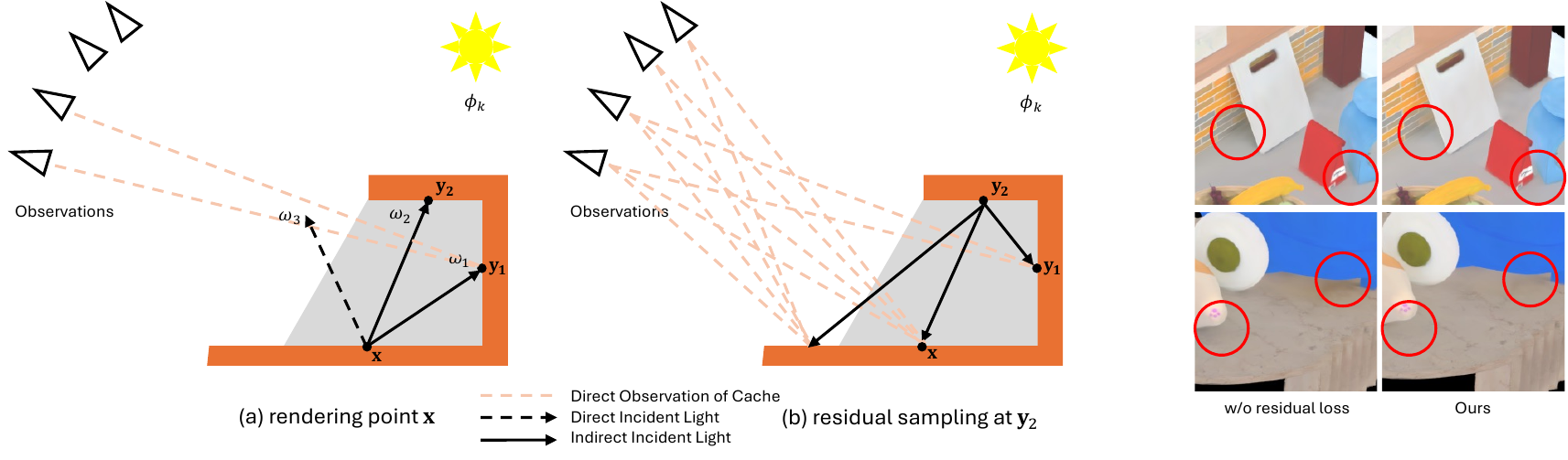}
\caption{Analysis of residual constraint. In the above example, consider the albedo estimation of point $\vec{x}$.
\textbf{(a)} when querying incident radiance $L_i(x,\omega_i)$, we show three possibilities of the sampled directions. $\omega_3$ does not hit scene geometry and queries the environment.
$\omega_1$ and $\omega_2$ hits the scene at $\vec{y_1},\vec{y_2}$ respectively, and will query the radiance cache $\vec{R}_k$.
Note that $\vec{y_1}$ can be observed from several cameras, so the cached radiance $\vec{R}_k(\vec{y_1},-\omega_1)$ may be relatively accurate, in contrast to $\vec{R}_k(\vec{y_2},-\omega_2)$.
\textbf{(b)} in this case, the residual constraint can be useful to correct $\vec{R}_k(\vec{y_2},-\omega_2)$ by rendering $L_o(\vec{y_2},-\omega_2)$ with Equation \ref{eq:render_incmesh}, which in turns queries incident radiance from more directly observable regions (like $\vec{x}$ and $\vec{y}_1$) towards $\vec{y_2}$.
As a result, the cached value $\vec{R}_k(\vec{y_2},-\omega_2)$ will be more accurate, which benefits the material estimation of $\vec{x}$.}
\label{fig:residual}
\end{figure*}

\providelength\width
\setlength\width{2.6cm}

\providelength\oldtabcolsep
\setlength{\oldtabcolsep}{\tabcolsep}
\setlength{\tabcolsep}{1pt}

\begin{figure}[t]
\centering
\footnotesize

\begin{tabular}{ccc}
IRGS & Ours & GT \\
\midrule
\includegraphics[width=\width]{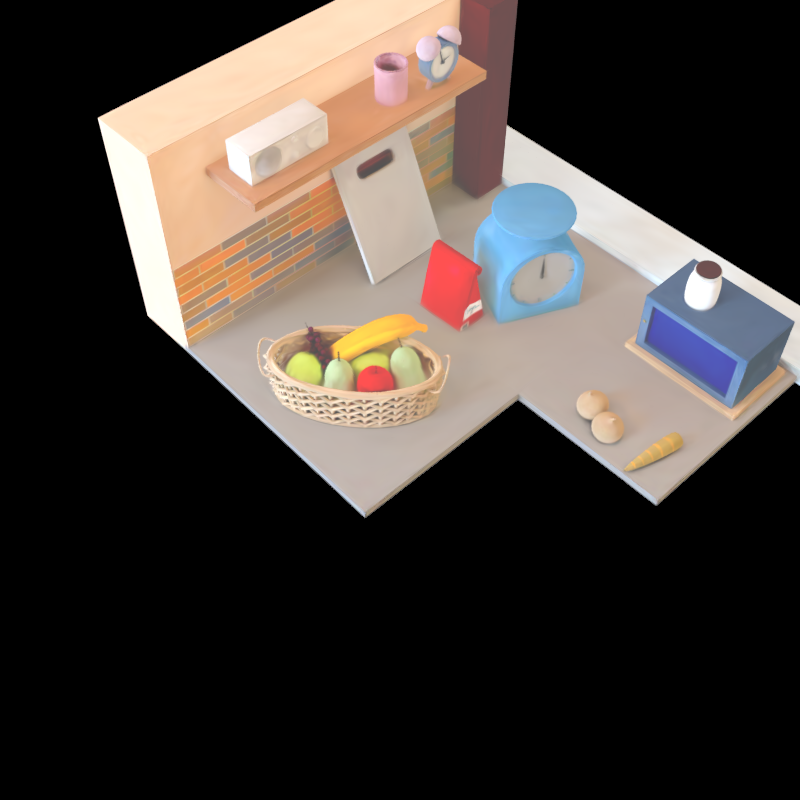} &
\includegraphics[width=\width]{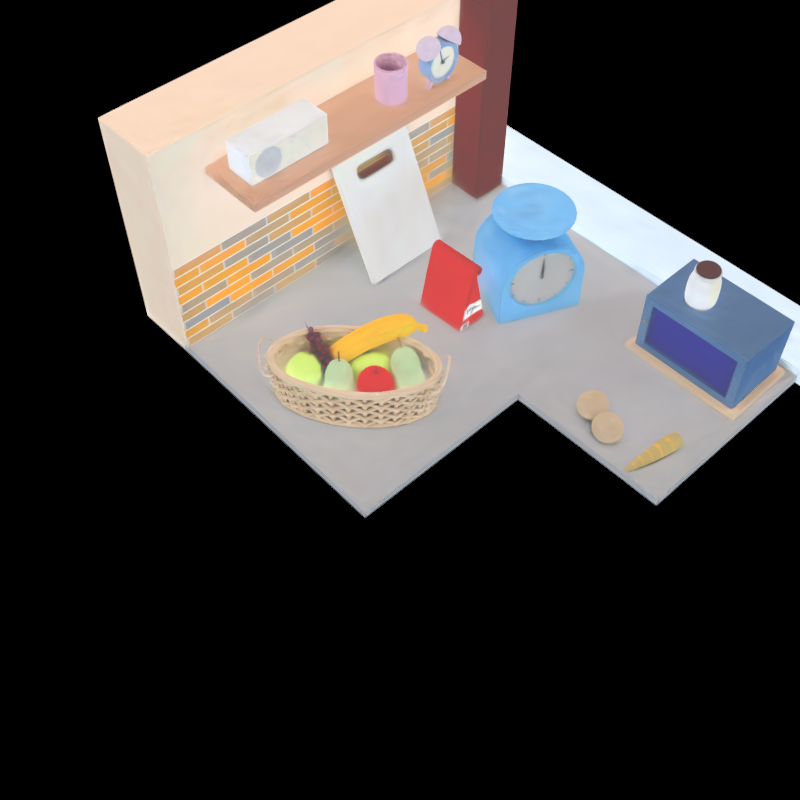} &
\includegraphics[width=\width]{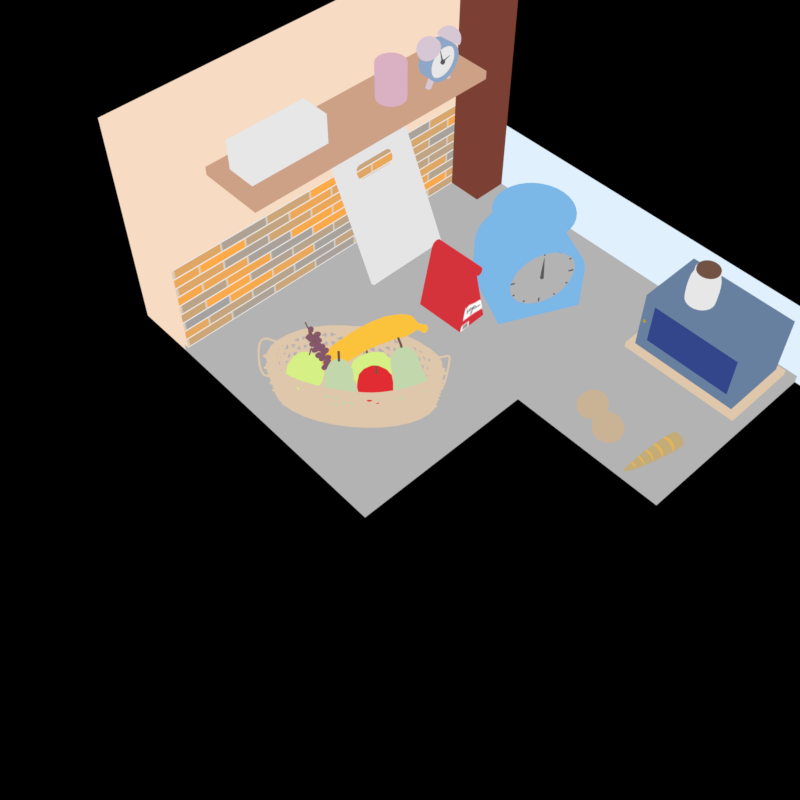} \\
20.589 / 0.916 &  27.722 / 0.955 &   \\
\includegraphics[width=\width]{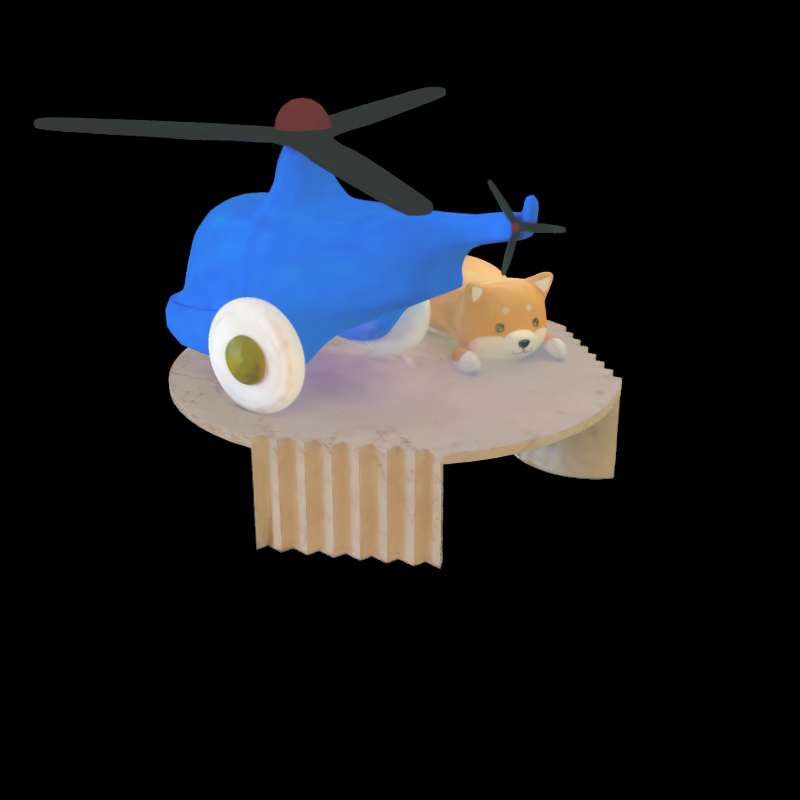} &
\includegraphics[width=\width]{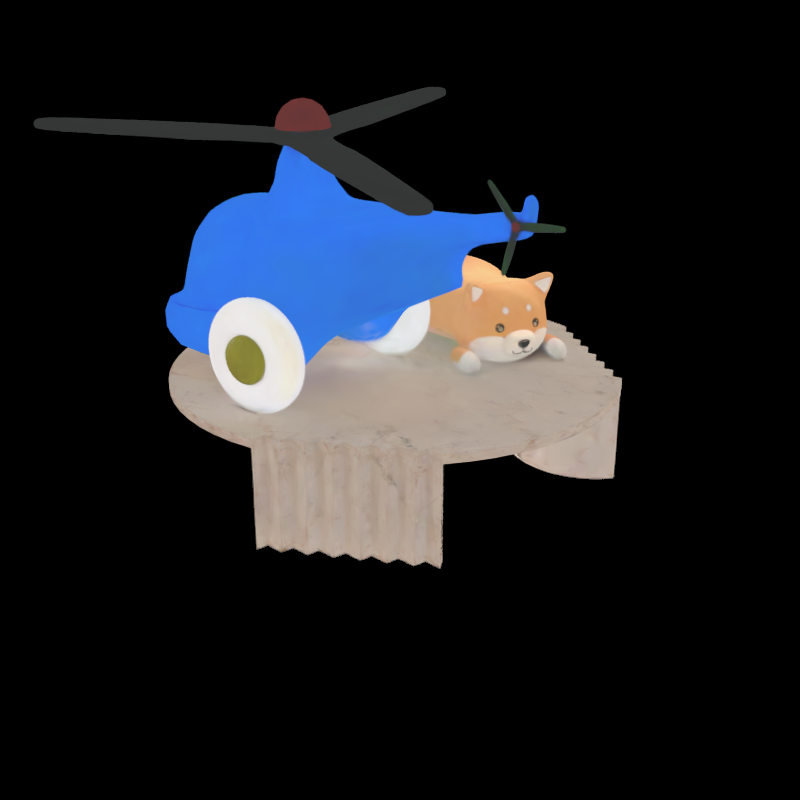} &
\includegraphics[width=\width]{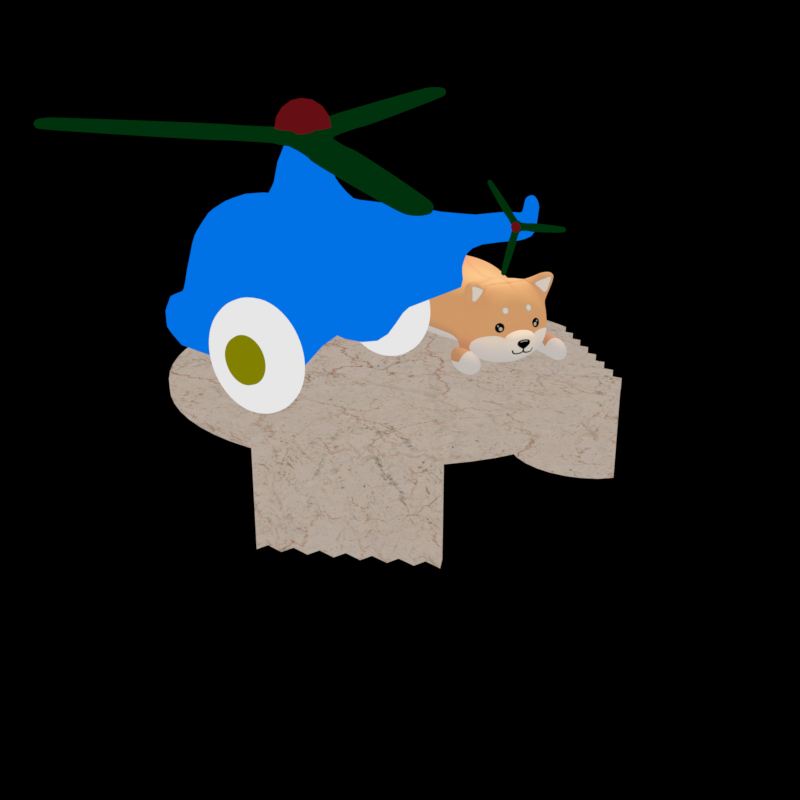} \\
28.293 / 0.927 &  34.206 / 0.941 &   \\
\includegraphics[width=\width]{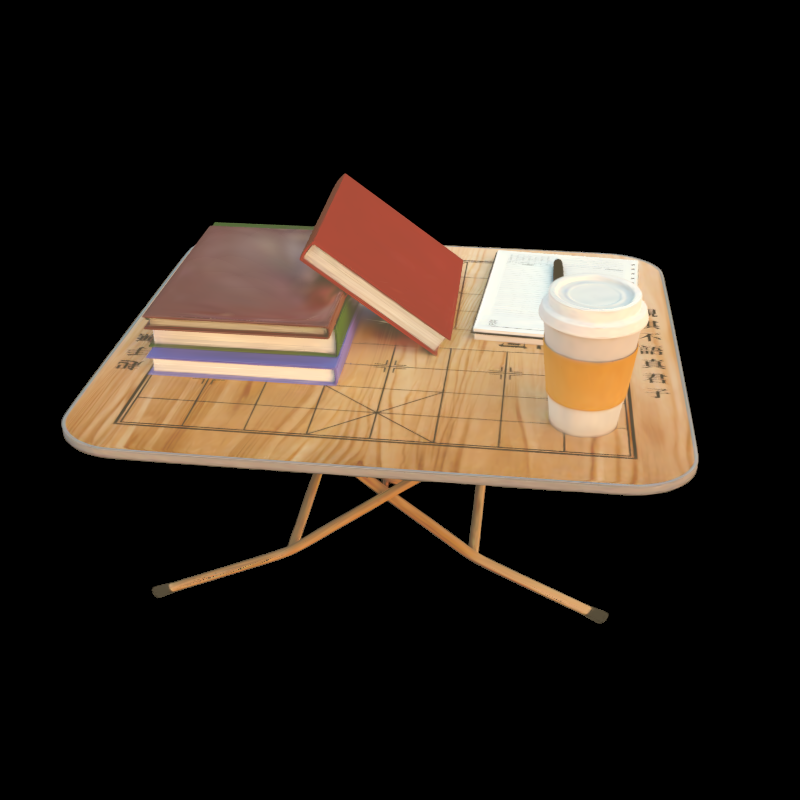} &
\includegraphics[width=\width]{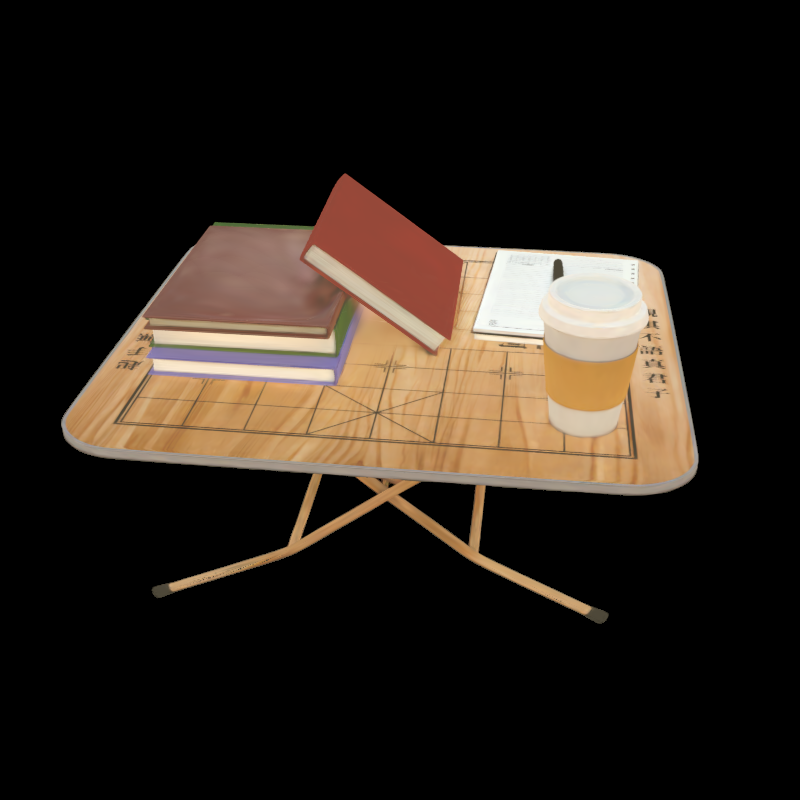} &
\includegraphics[width=\width]{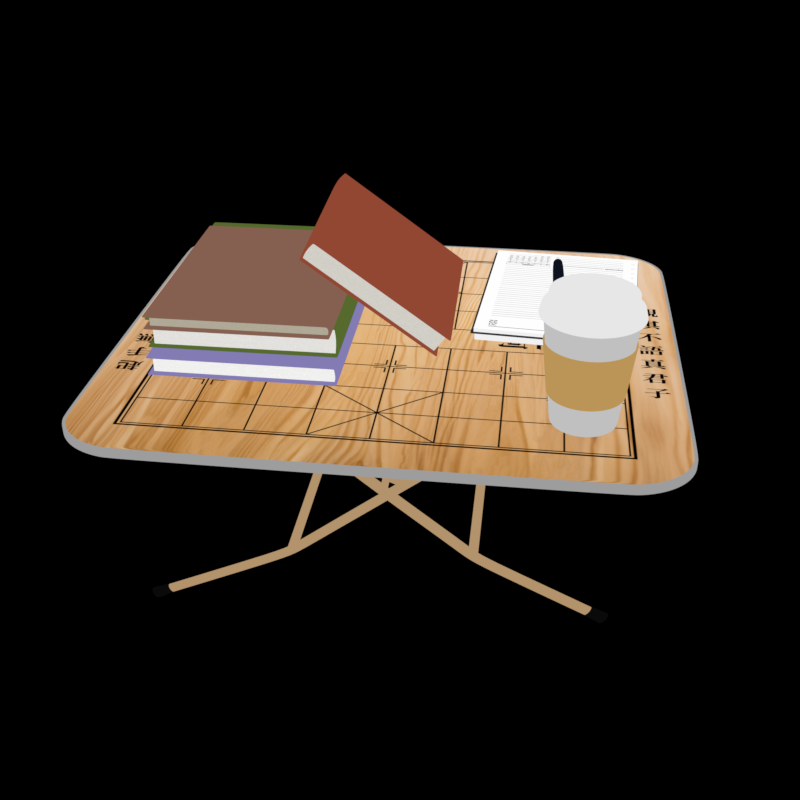} \\
26.567 / 0.960 &  29.664 / 0.968 &   \\
\includegraphics[width=\width]{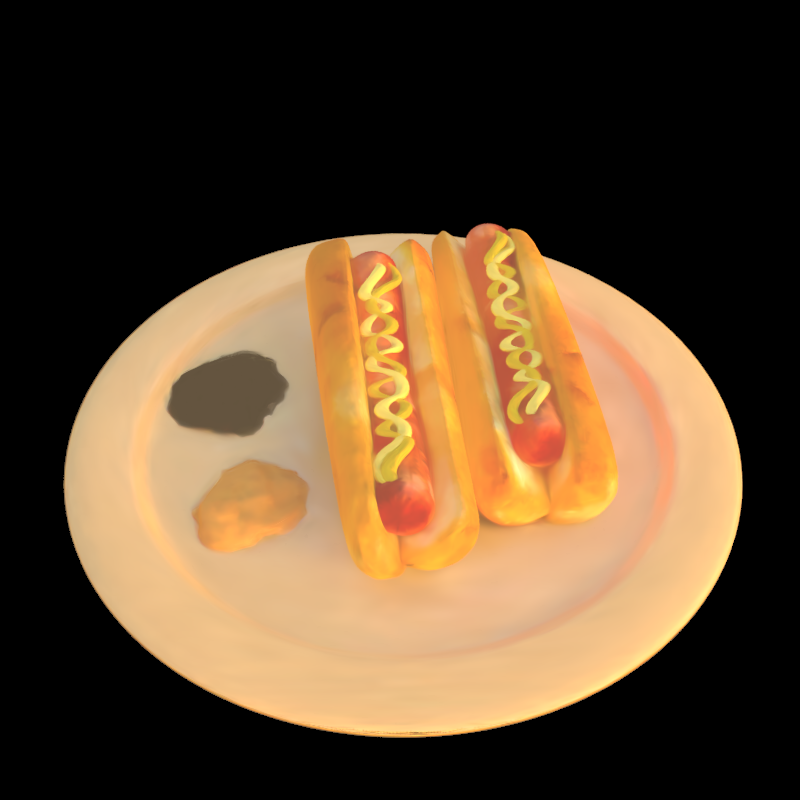} &
\includegraphics[width=\width]{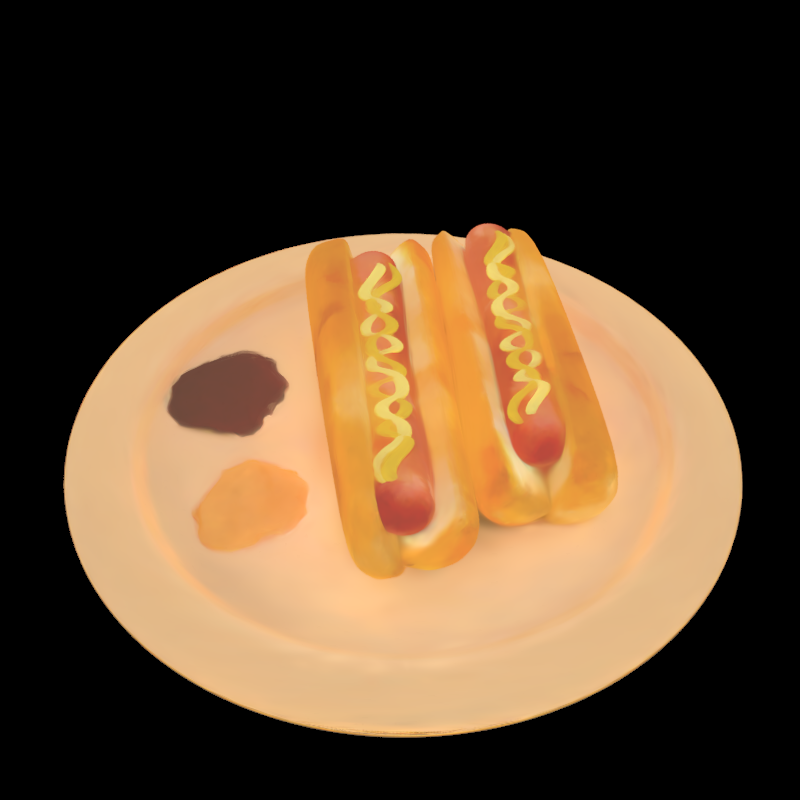} &
\includegraphics[width=\width]{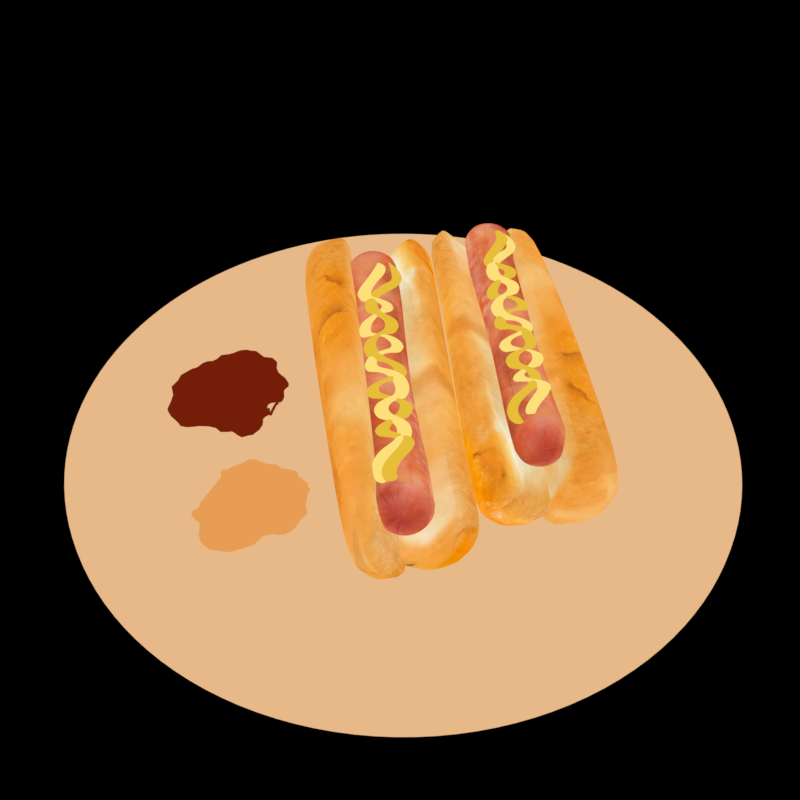} \\
24.796 / 0.922 &  29.116 / 0.949 &   \\
\end{tabular}

\caption{Qualitative results of albedo estimation. Thanks to our \textit{RotLight} setup, our method is able to better factor out environment illumination from albedo, resulting in more consistent and accurate estimations. Numbers are PSNR and SSIM metrics.}
\label{fig:inverse_qualitative}

\end{figure}

\setlength\tabcolsep{\oldtabcolsep}

\subsection{Material Recovery}

We evaluated and compared the performance of the material reconstruction of our proposed method in Table \ref{tab:main_quant}. We also show qualitative comparisons of albedo in Figure \ref{fig:inverse_qualitative}. Our method achieves on-par performance in roughness estimation and environment light estimation, and qualitative examples are included in the supplementary.

\subsection{Ablation Study}

We perform an ablation study of our proposed components in Table \ref{tab:main_quant}.
Qualitative examples are shown in the supplementary material.
The use of the \textit{RotLight} capture setup drastically improves the recovery of albedo.
With the proposed proxy mesh geometry for incident queries, incident lights are more accurately integrated, which also benefits material estimation.
The residual constraint, while not significant in metrics, is shown to be effective in improving regions covered in shadows. We provide a more detailed analysis below.

\subsection{Residual Constraint}

As depicted in Figure \ref{fig:residual}, the residual constraint is most effective in regions where indirect illumination comes from points and directions that are not directly visible in any input views. In practice, this often means regions that are covered or surfaces that are close to each other.

\providelength\width
\setlength\width{2.6cm}

\providelength\oldtabcolsep
\setlength{\oldtabcolsep}{\tabcolsep}
\setlength{\tabcolsep}{1pt}

\begin{figure}[t]
\centering
\footnotesize

\begin{tabular}{ccc}
IRGS & Ours & GT \\
\midrule
\includegraphics[width=\width]{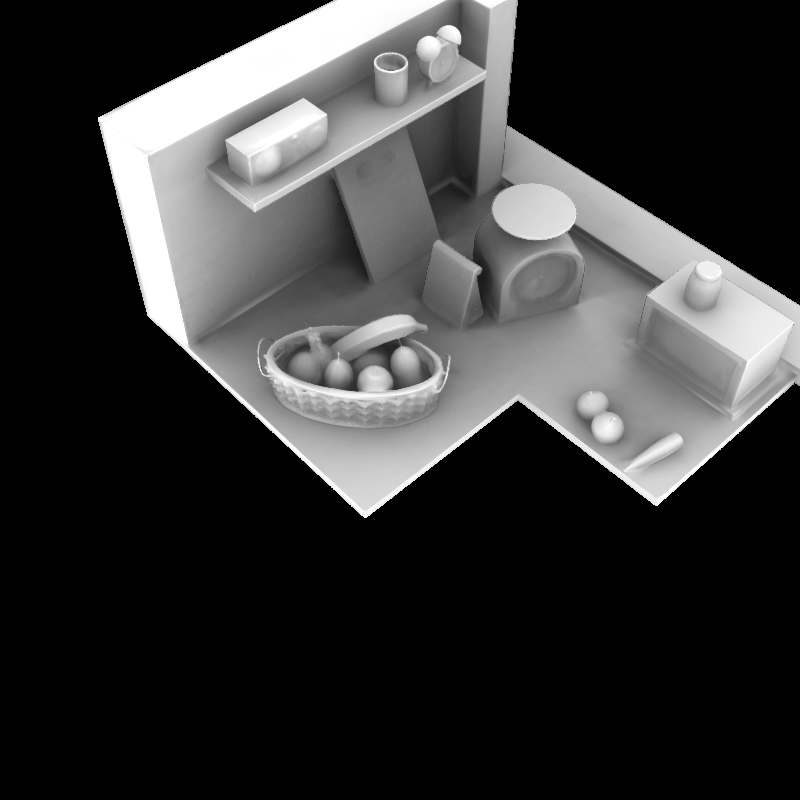} &
\includegraphics[width=\width]{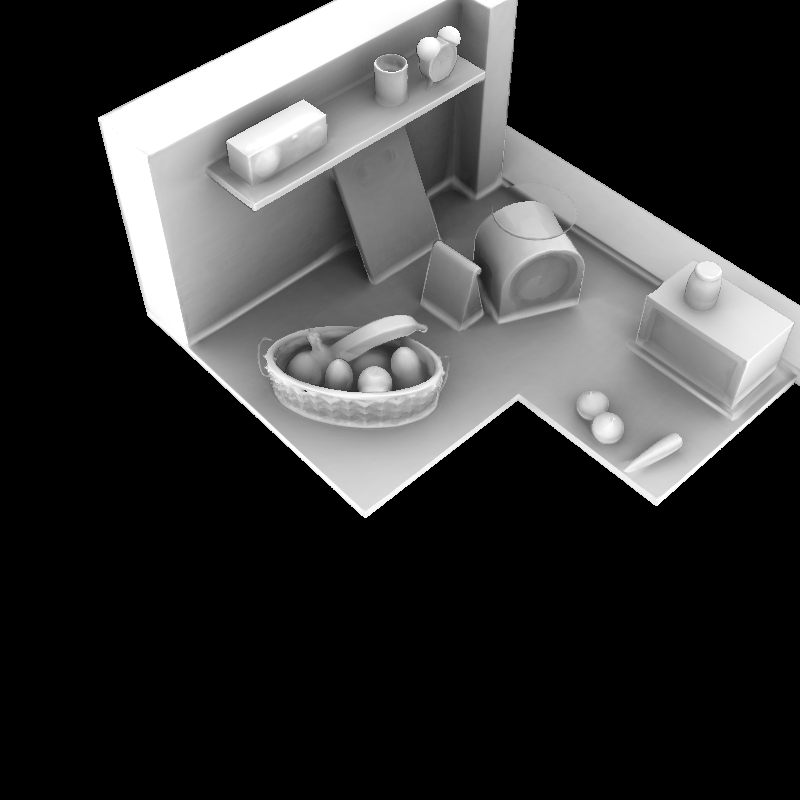} &
\includegraphics[width=\width]{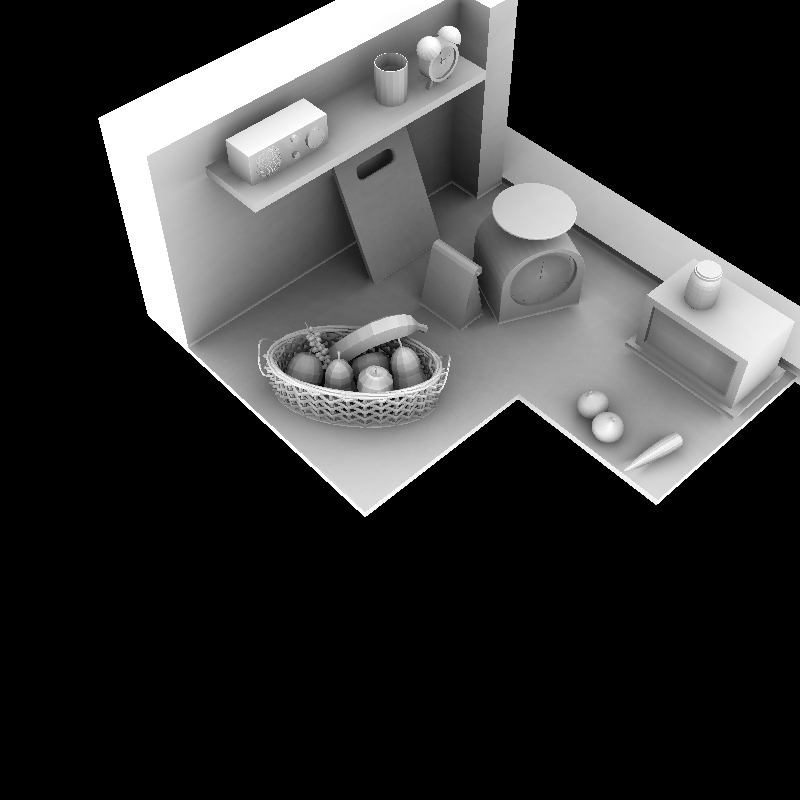} \\
\includegraphics[width=\width]{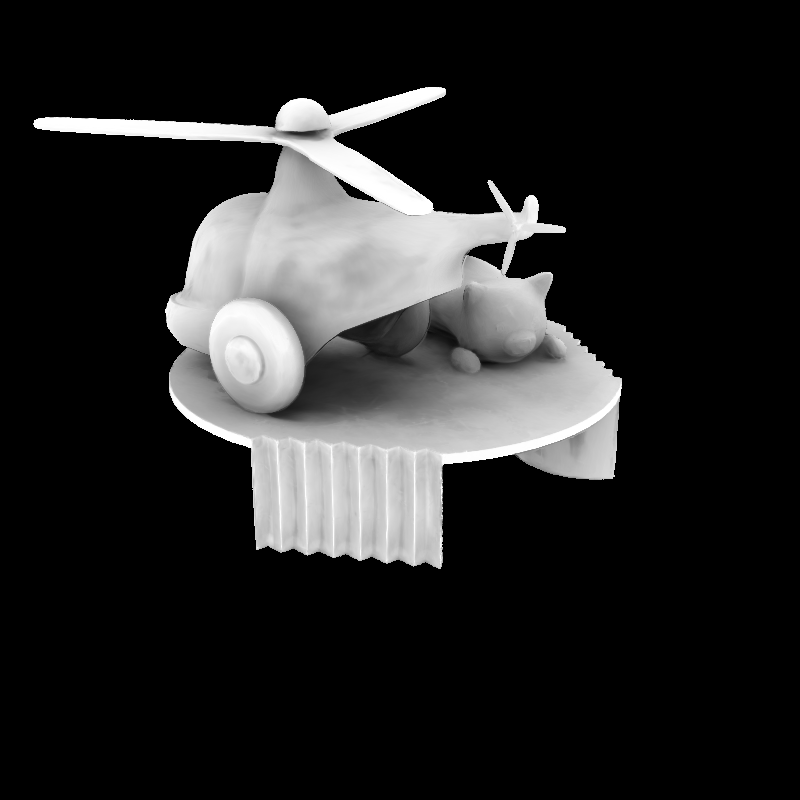} &
\includegraphics[width=\width]{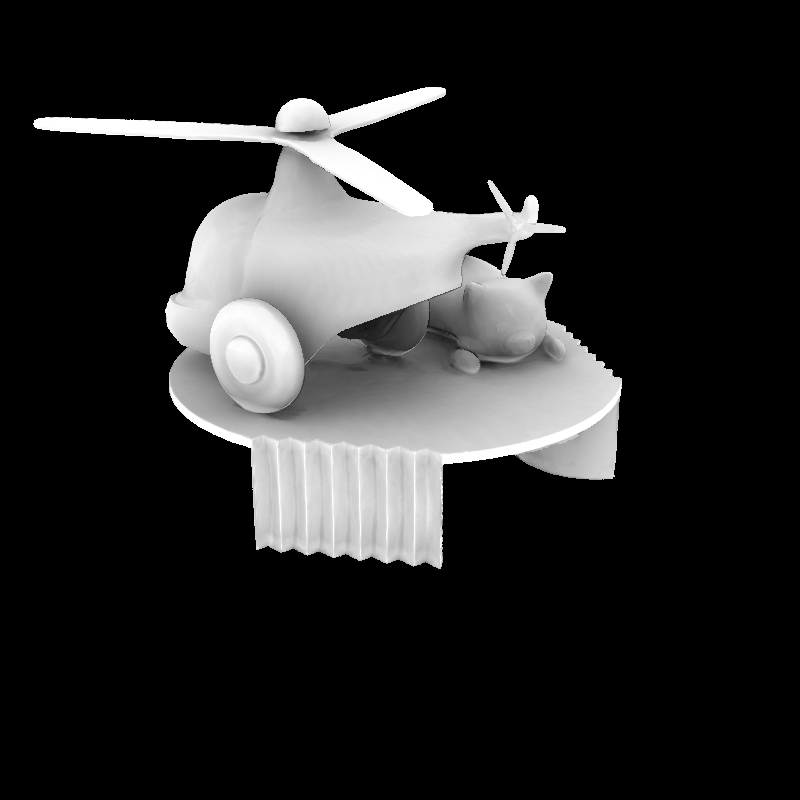} &
\includegraphics[width=\width]{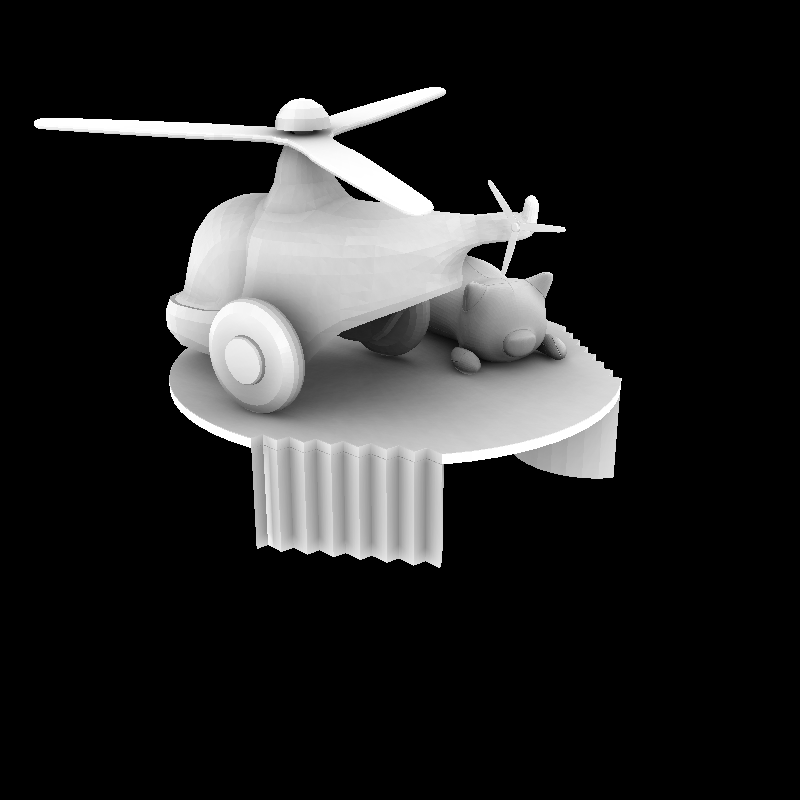} \\
\includegraphics[width=\width]{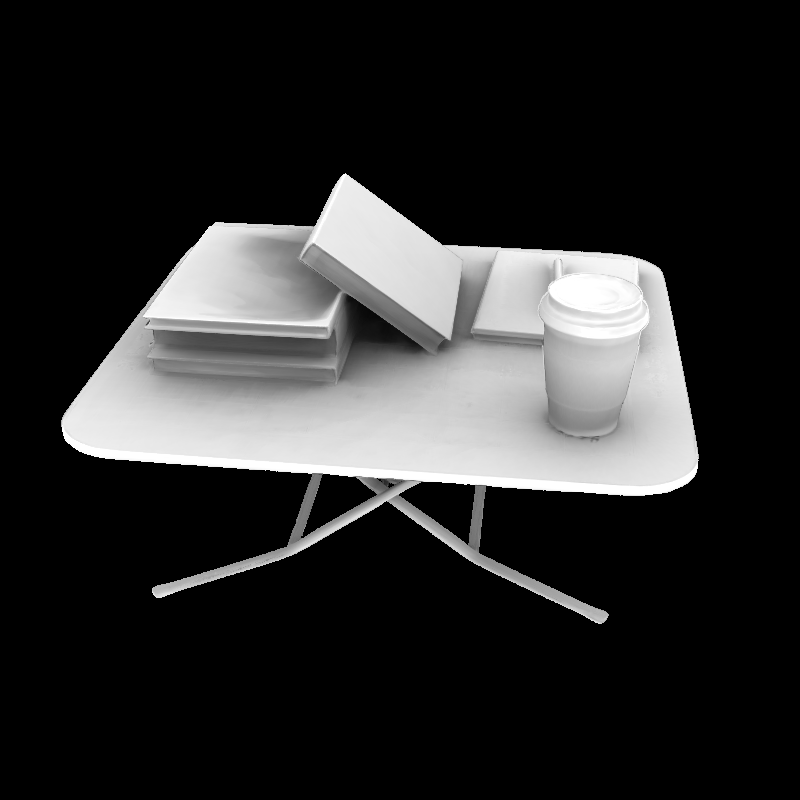} &
\includegraphics[width=\width]{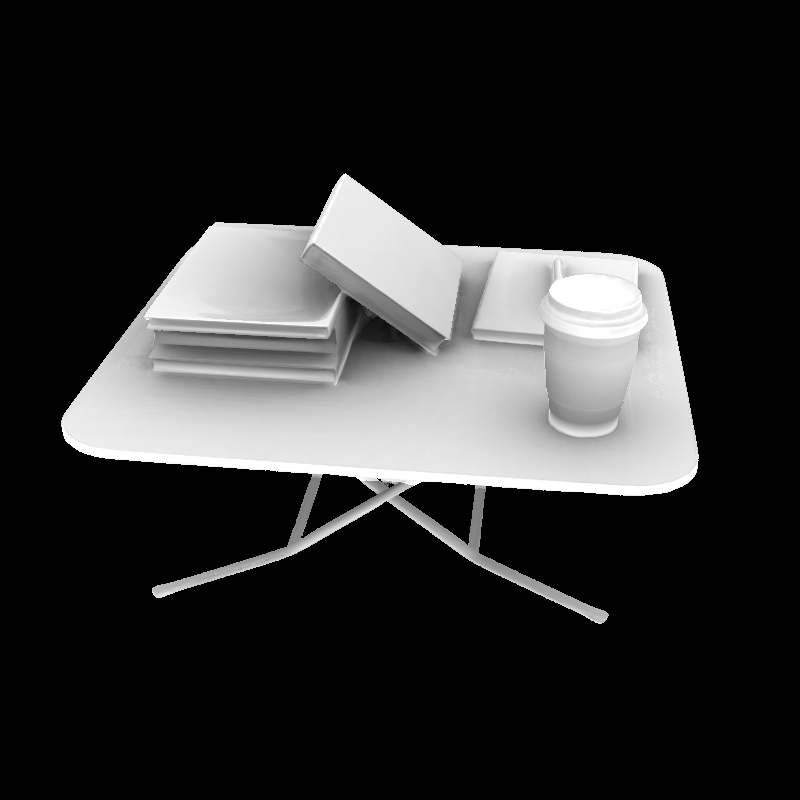} &
\includegraphics[width=\width]{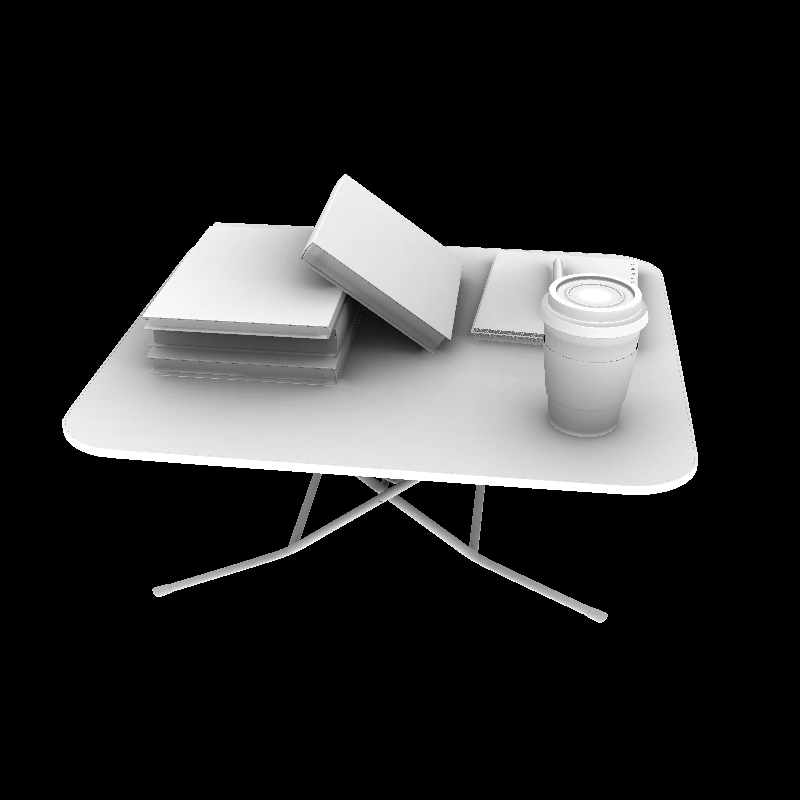} \\
\includegraphics[width=\width]{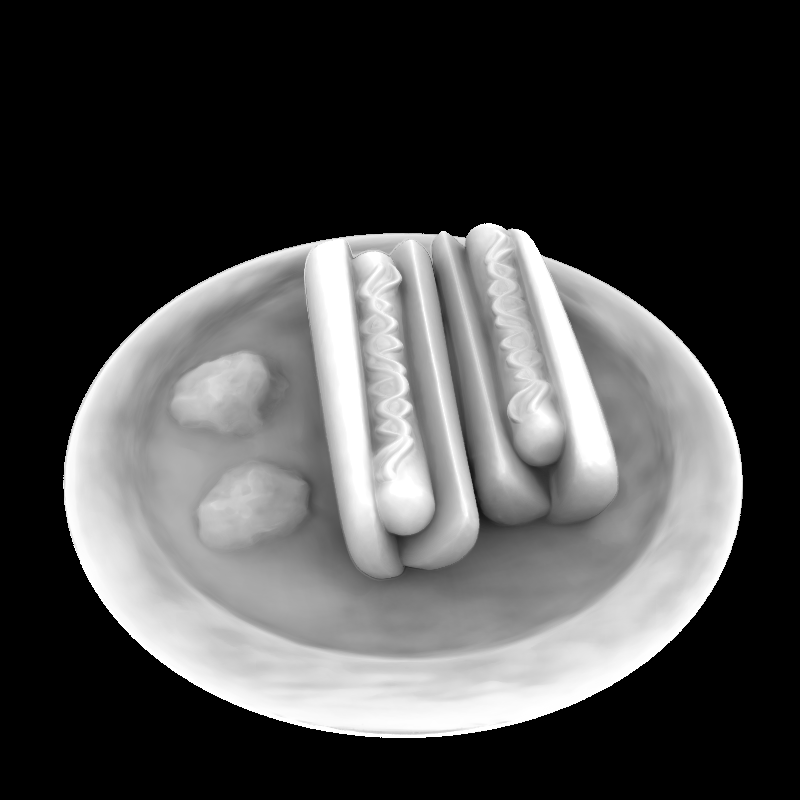} &
\includegraphics[width=\width]{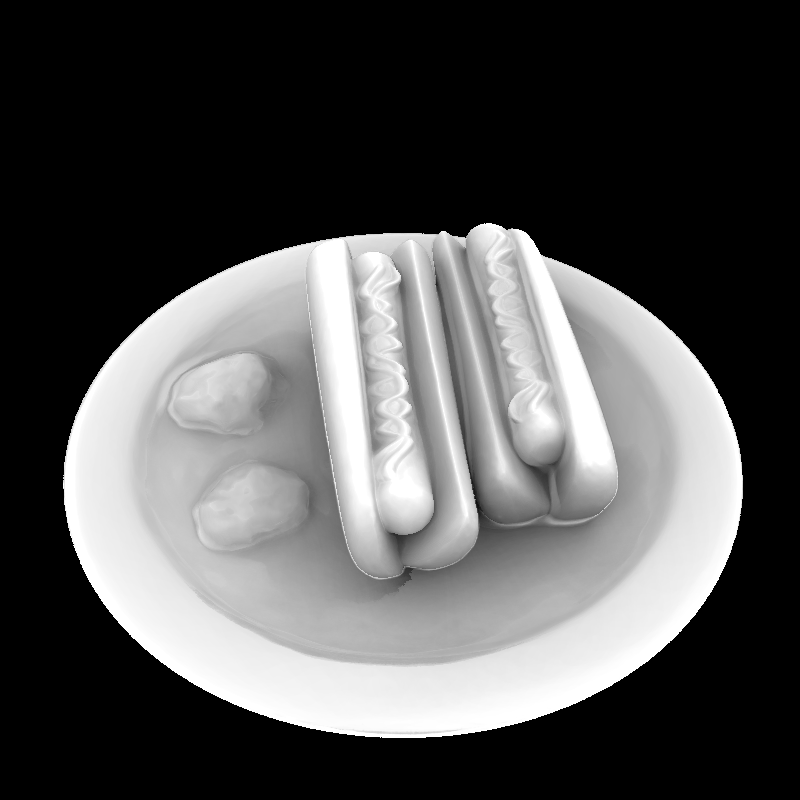} &
\includegraphics[width=\width]{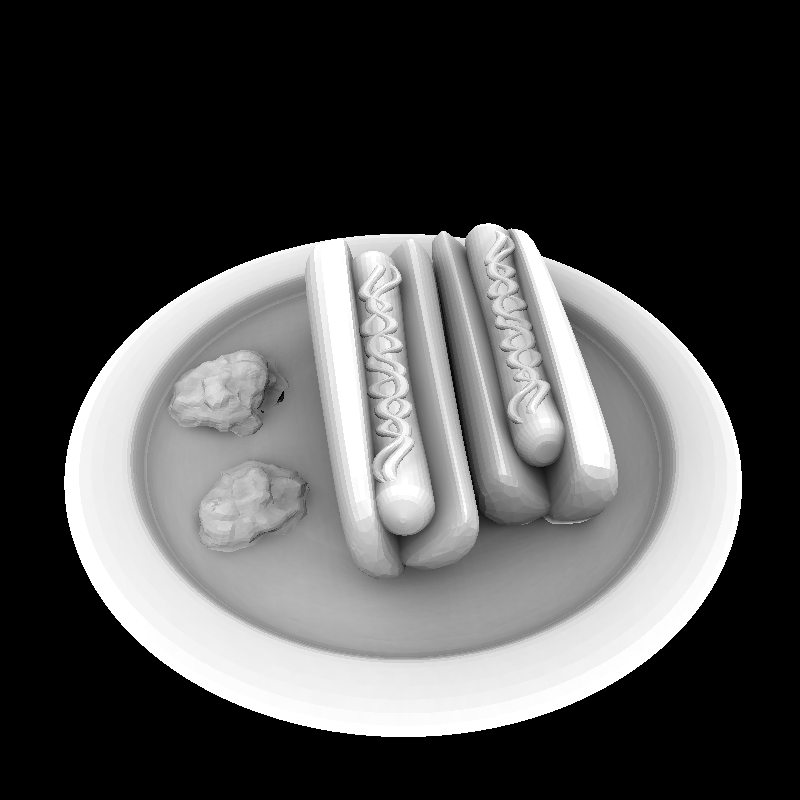} \\
\end{tabular}

\caption{Ambient occlusion rendering. With the proxy mesh, our ambient occlusion rendering is smoother and more accurate.}
\label{fig:ambient}

\end{figure}

\setlength\tabcolsep{\oldtabcolsep}

\subsection{Mesh Proxy Geometry}

We use a mesh as a proxy geometry for incident queries for its better accuracy compared to integrating 2D Gaussians. To demonstrate its effects, we show ambient occlusion maps in Figure \ref{fig:ambient}. When ray tracing 2D Gaussians, the origins must be moved along the ray by an appropriate amount, which can be challenging to determine and adapt. A small value may lead to ray intersecting surfaces at the origin, while a large value can move the origins so far that some intersections are ignored. In practice, IRGS~\cite{gu2024IRGS} uses a large factor of 0.05 which, as we show, leads to noisy and erroneous incident visibility results.

\providelength\width
\setlength\width{2.6cm}

\providelength\oldtabcolsep
\setlength{\oldtabcolsep}{\tabcolsep}
\setlength{\tabcolsep}{1pt}

\begin{figure}[t]
\centering
\footnotesize

\begin{tabular}{ccc}
0 degrees & 90 degrees & 180 degrees \\
\midrule
\includegraphics[width=\width]{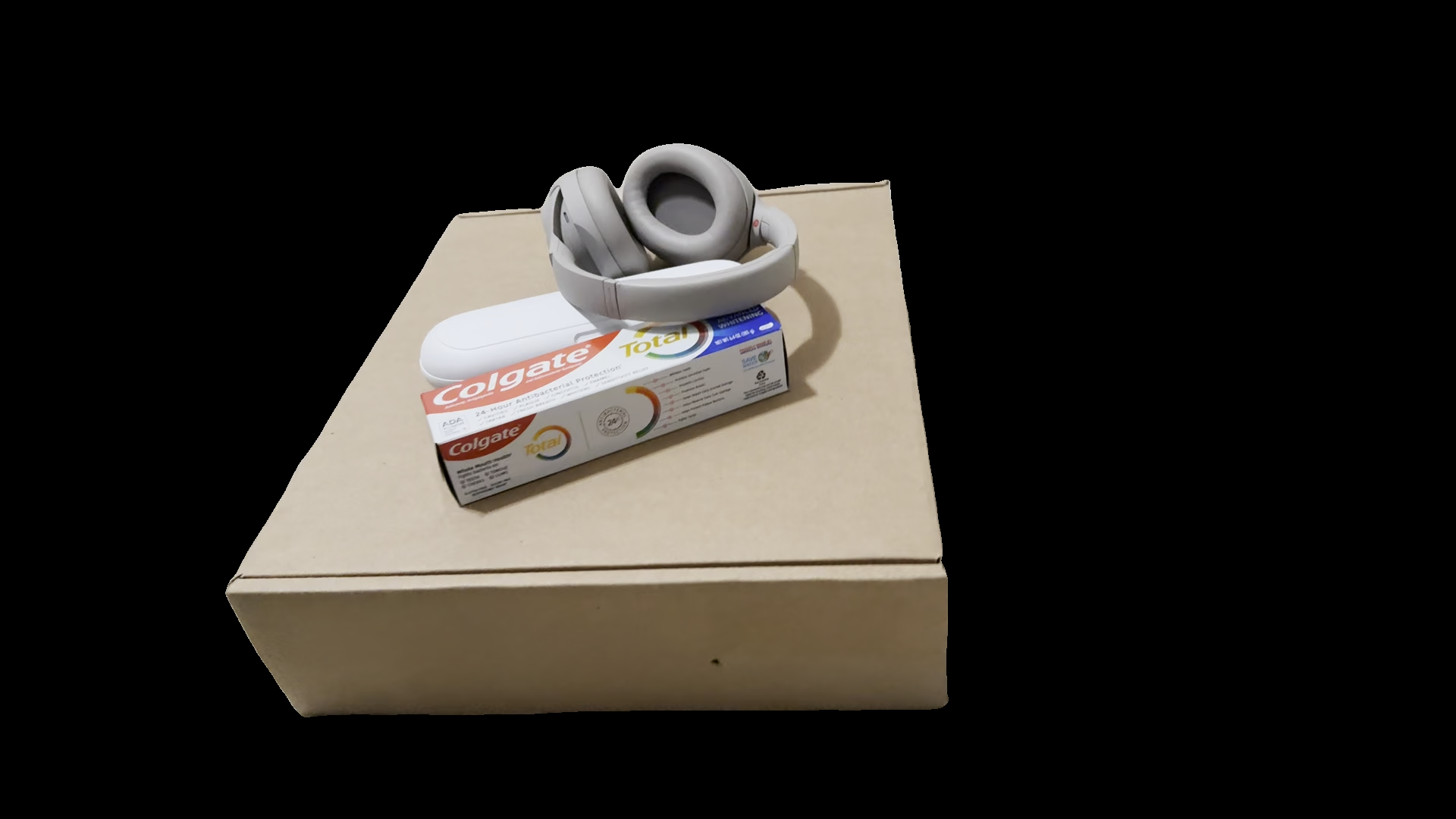} &
\includegraphics[width=\width]{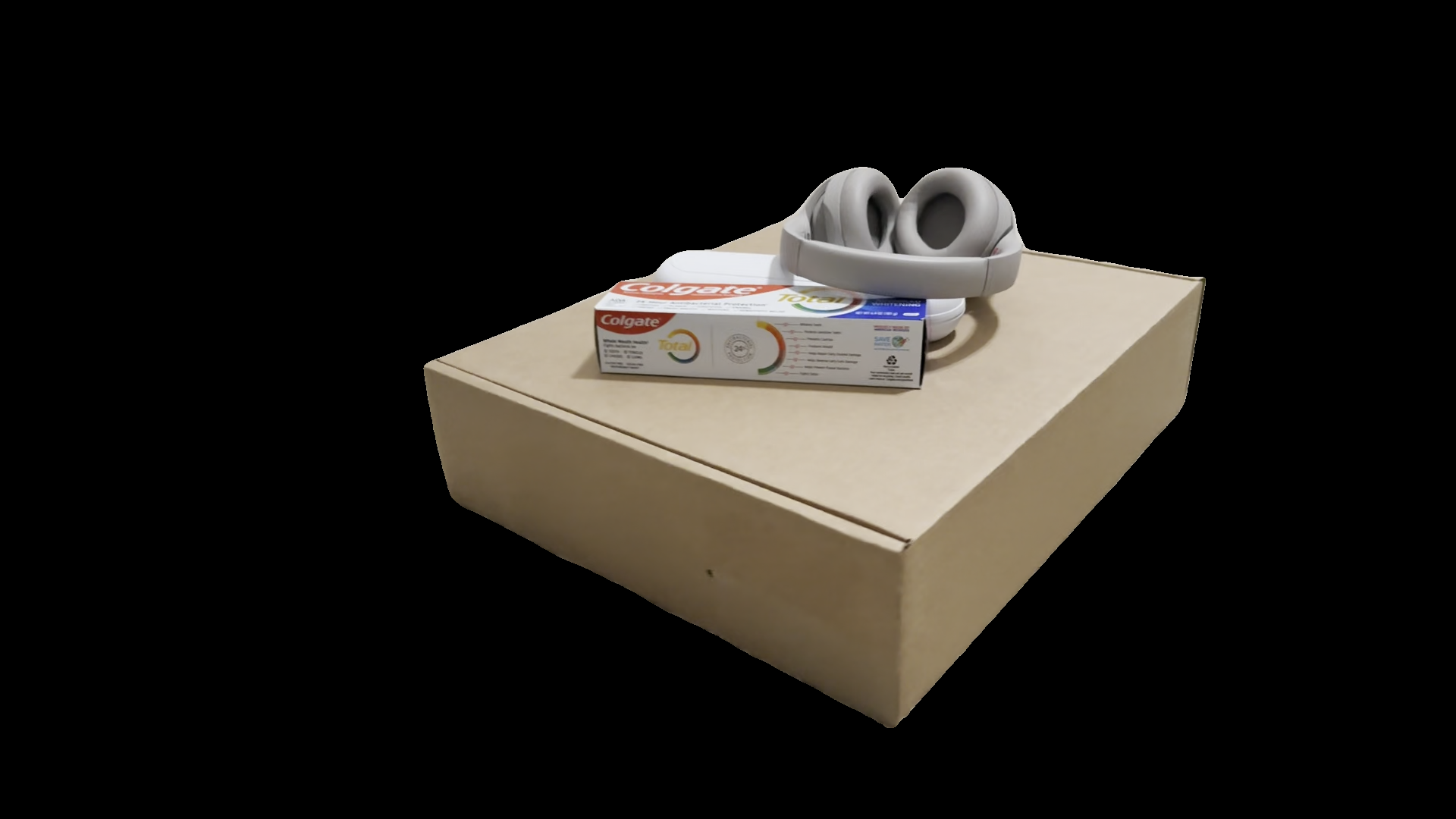} &
\includegraphics[width=\width]{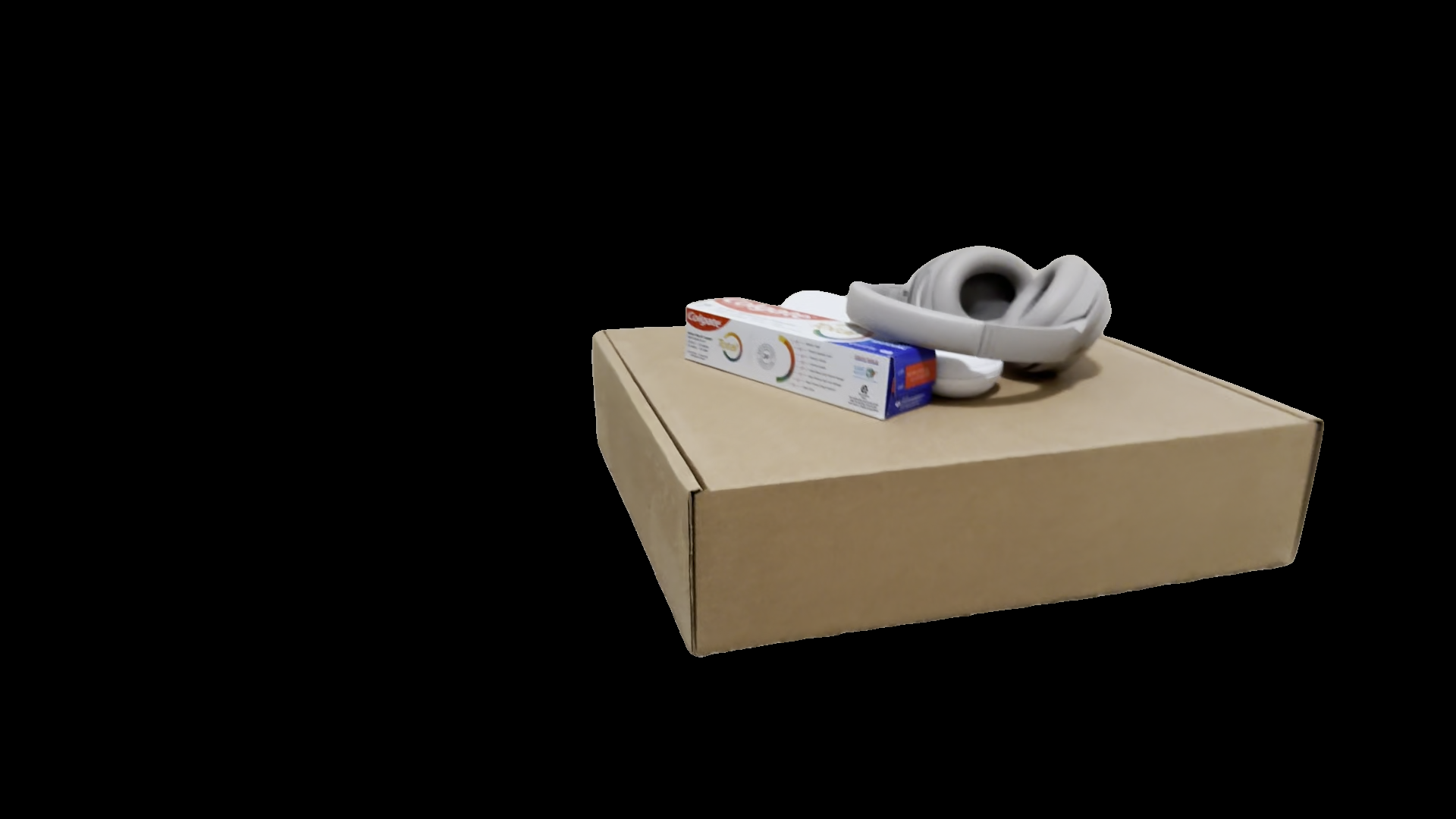} \\
\includegraphics[width=\width]{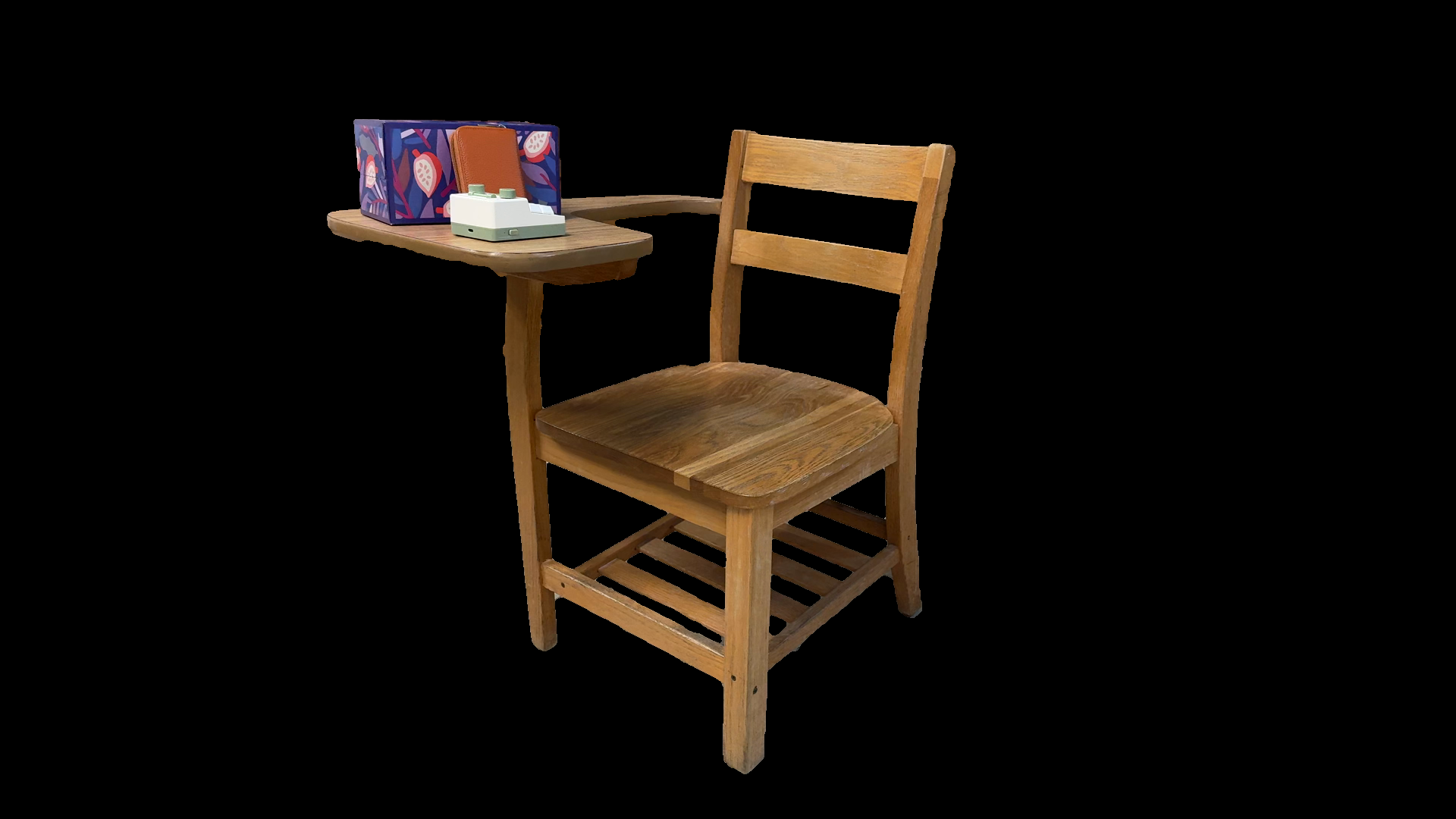} &
\includegraphics[width=\width]{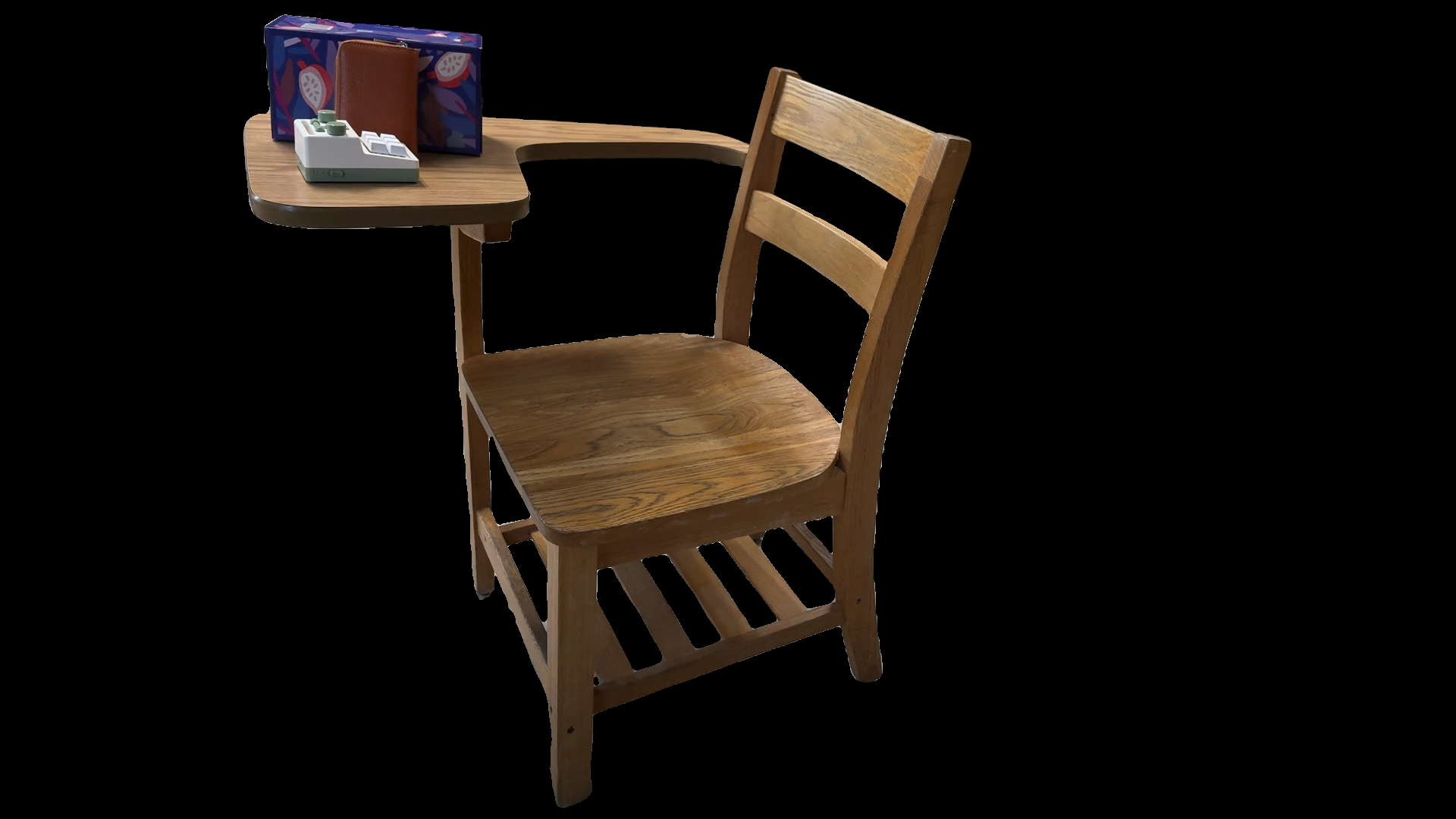} &
\includegraphics[width=\width]{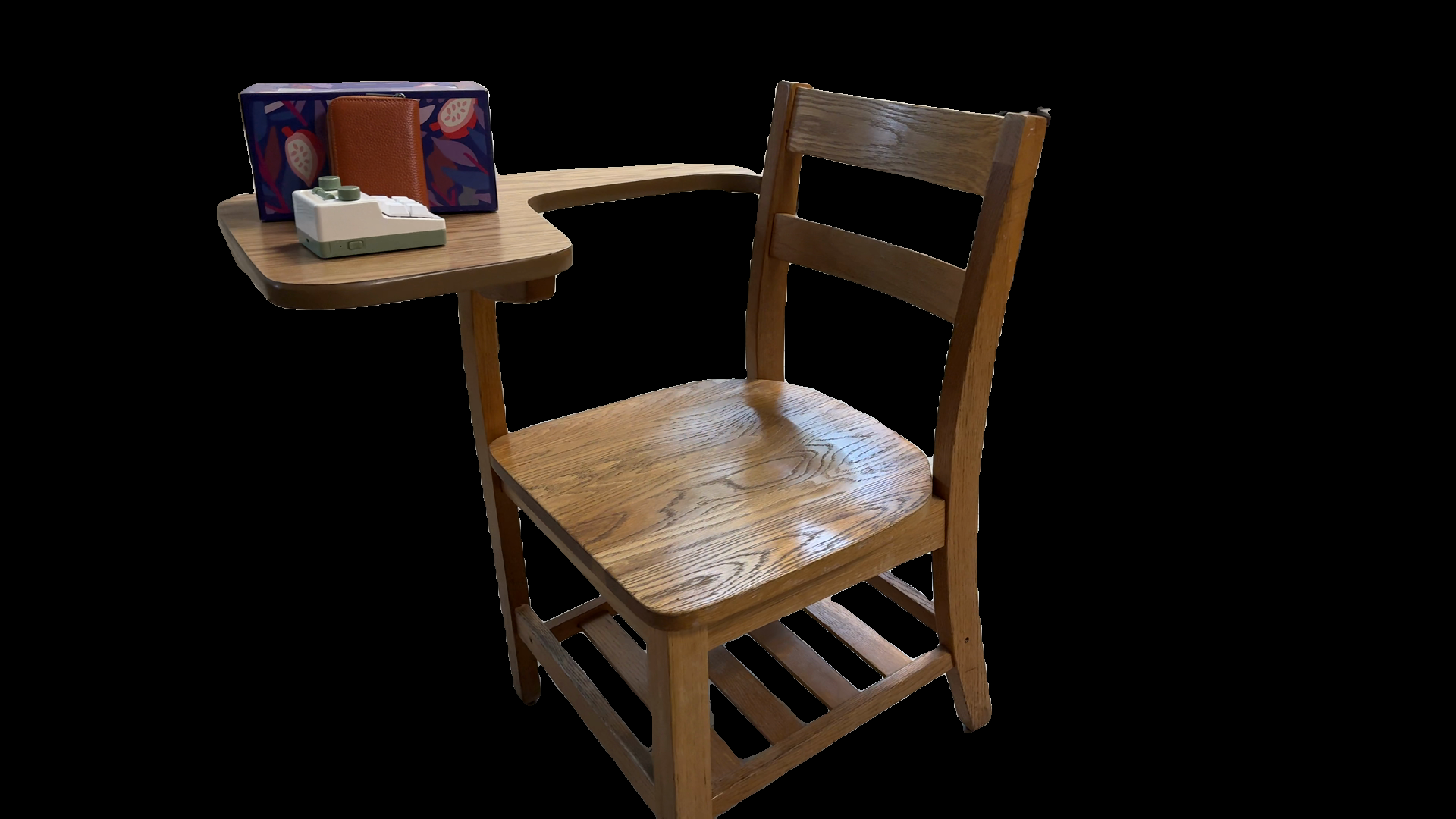} \\
\end{tabular}

\caption{Examples of the captured real-world videos, \textit{box} and \textit{desk}.}
\label{fig:inverse_real_images}

\end{figure}

\setlength\tabcolsep{\oldtabcolsep}

\providelength\width
\setlength\width{2.6cm}

\providelength\oldtabcolsep
\setlength{\oldtabcolsep}{\tabcolsep}
\setlength{\tabcolsep}{1pt}

\begin{figure}[t]
\centering
\footnotesize

\begin{tabular}{ccc}
IRGS & Ours & Input \\
\midrule
\includegraphics[width=\width]{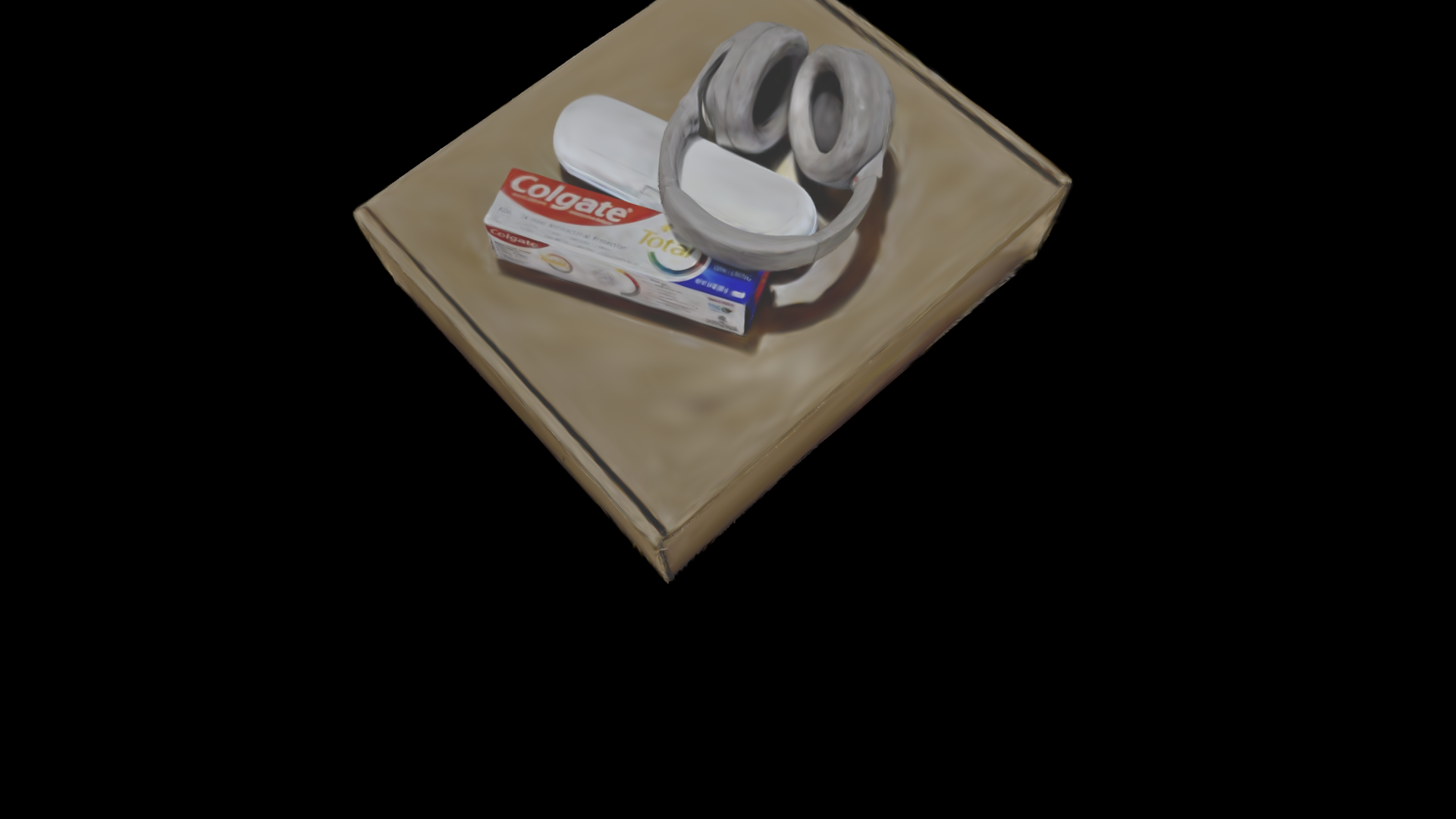} &
\includegraphics[width=\width]{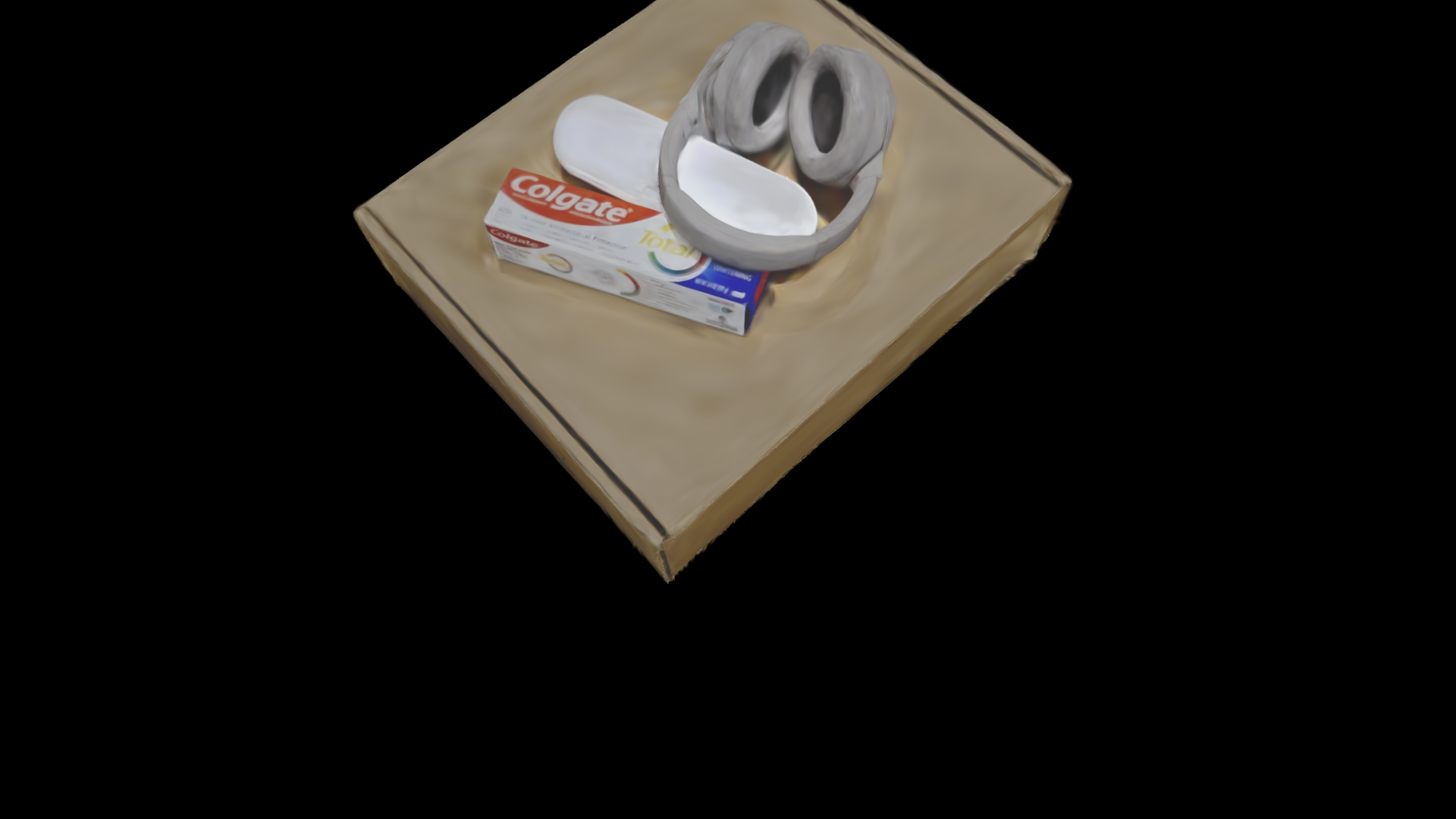} &
\includegraphics[width=\width]{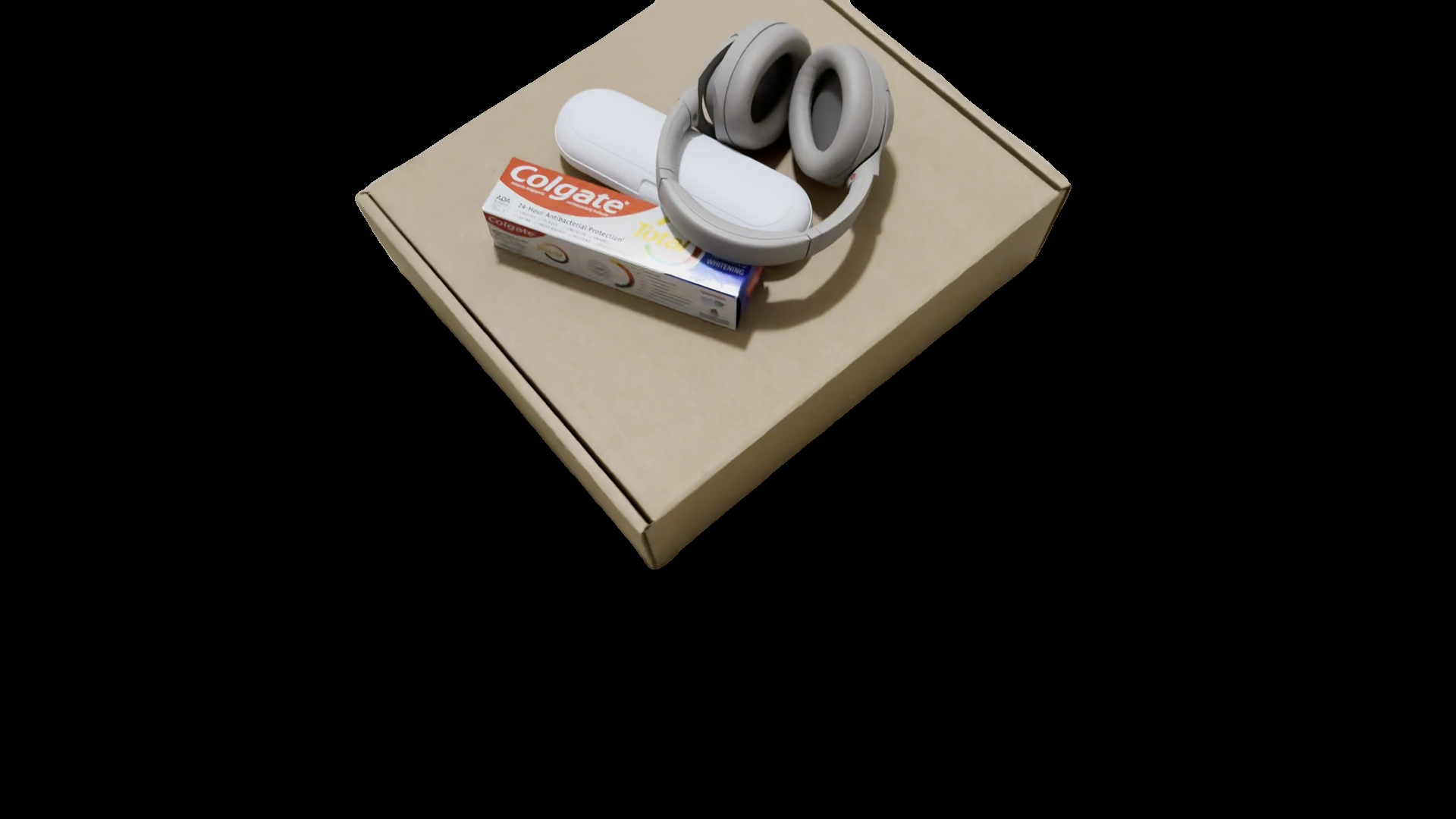} \\
\includegraphics[width=\width]{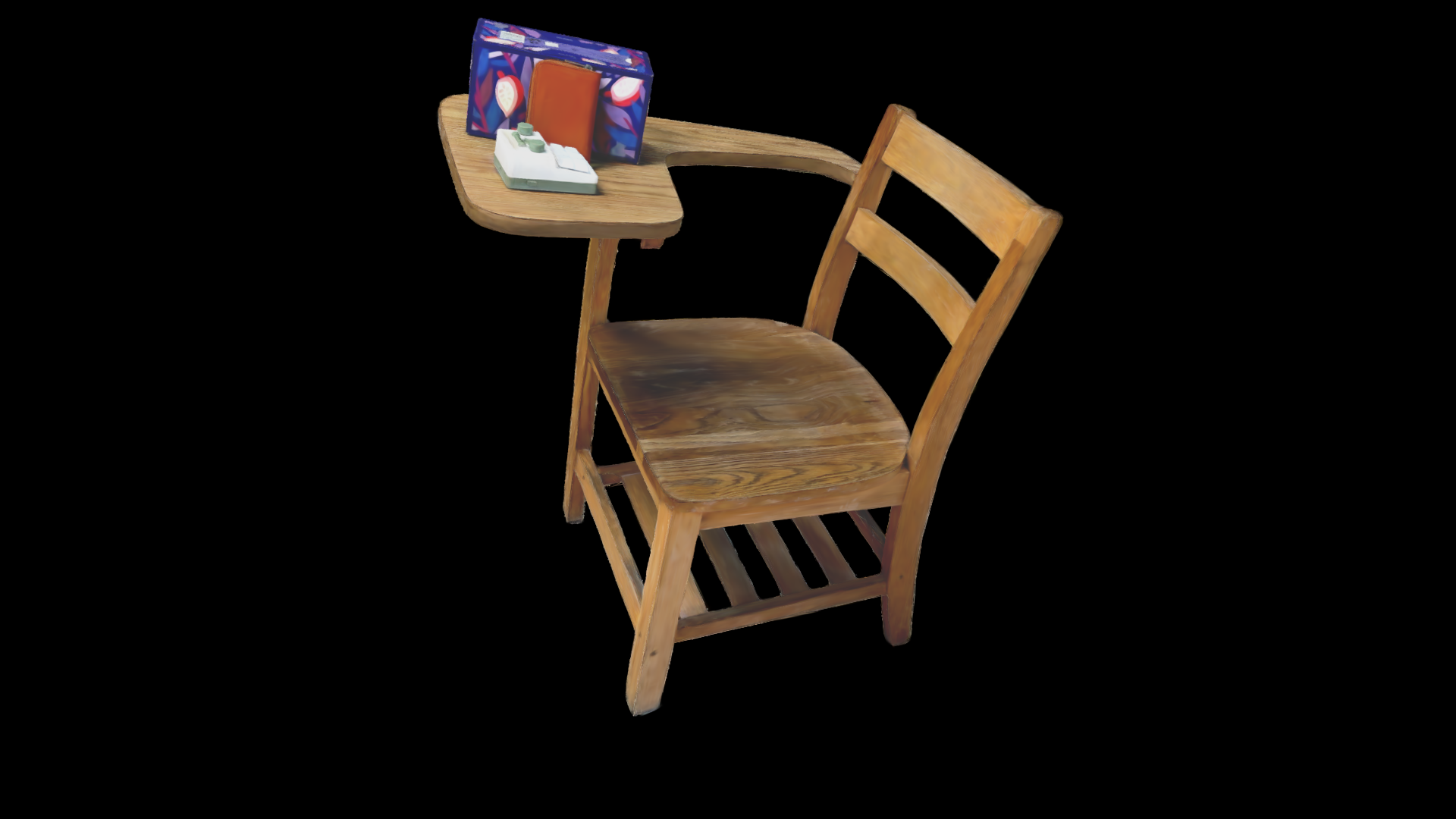} &
\includegraphics[width=\width]{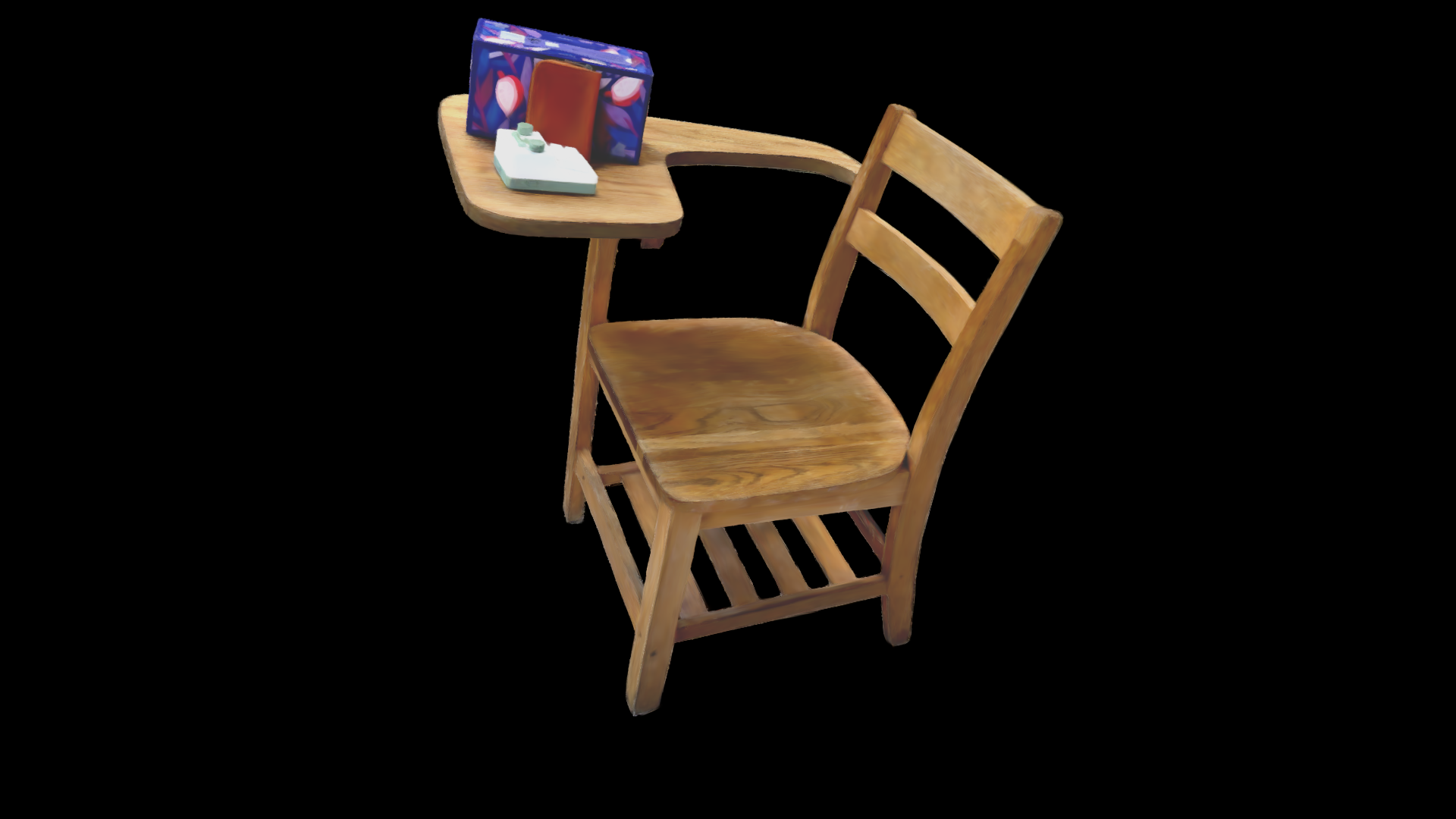} &
\includegraphics[width=\width]{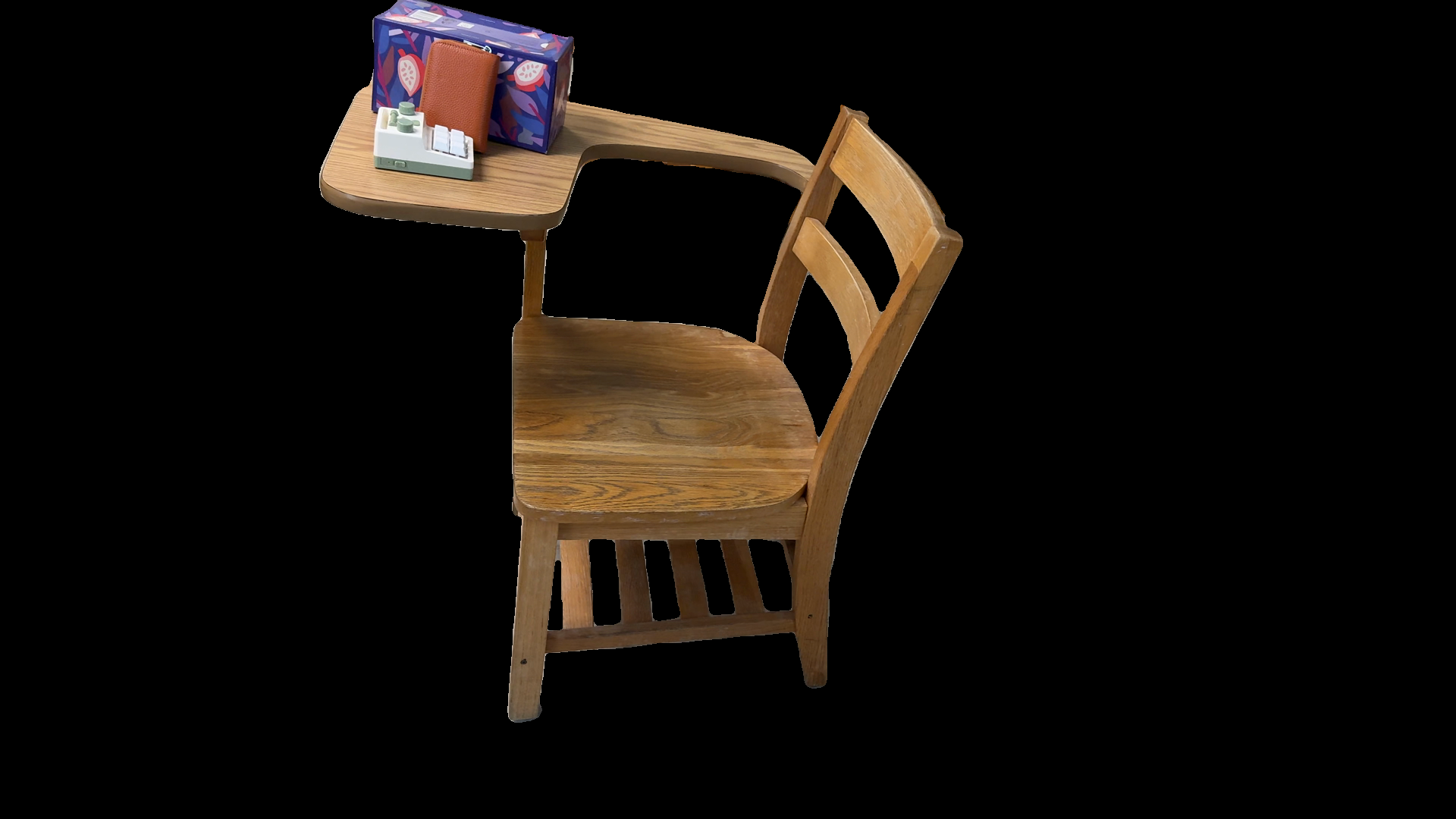} \\
\end{tabular}

\caption{Albedo recovery of real-world data. Thanks to RotLight capture, our recovered albedo maps have less baked shadows.}
\label{fig:inverse_real_albedo}

\end{figure}

\setlength\tabcolsep{\oldtabcolsep}

\subsection{Real World Examples}

We capture two real-world videos, \textit{box} and \textit{desk}, with examples shown in Figure \ref{fig:inverse_real_images}.
Each scene has three videos with the object rotated by 0, 90 and 180 degrees.
For every scene, object masks are created by rotoscoping in Adobe After Effects, and the masked images from all videos of the scene are registered with COLMAP~\cite{schoenberger2016sfm, schoenberger2016mvs}.
We show the recovered albedo maps from IRGS and our method in Figure \ref{fig:inverse_real_albedo}.
Our method makes use of multiple light sources and estimates albedo with less baked shadows.

\section{Limitations}

Our method is not free from limitations.
First, its performance is highly dependent on the quality of the mesh extracted from the initial 2D Gaussian model.
This is particularly challenging near the object intersections like the bottom of the hotdogs, where inaccurate mesh prevents decomposition of shadows from albedo.
We further demonstrate this in the appendix, where we experiment with known mesh geometry and it helps reduce baked artifacts.
Second, the details of the reconstructed albedo in the shadowed regions can be blurred, like under the books of \textit{table}.
The reason is that the material properties are encoded in the Gaussians, and these dark regions tend to be under-represented with low Gaussians density.
It may be possible to mitigate this problem with alternative material encoding methods such as MLPs.

\section{Conclusion}

We propose a novel inverse rendering method that utilizes rotated captures of an object. Using 2D Gaussians as scene representation, we incorporate a proxy mesh and residual constraints on the scene radiance cache to improve accuracy of global illumination effect modeling, resulting in state-of-the-art performance in albedo estimation.

\section*{Acknowledgement}

This research is based upon work supported by the Office of the Director of National Intelligence (ODNI), Intelligence Advanced Research Projects Activity (IARPA), via IARPA R\&D Contract No. 140D0423C0076. The views and conclusions contained herein are those of the authors and should not be interpreted as necessarily representing the official policies or endorsements, either expressed or implied, of the ODNI, IARPA, or the U.S. Government. The U.S. Government is authorized to reproduce and distribute reprints for Governmental purposes notwithstanding any copyright annotation thereon. Further support was provided by the National Science Foundation through grant IIS-2126407.

\clearpage

{
    \small
    \bibliographystyle{ieeenat_fullname}
    \bibliography{references}
}

\clearpage
\appendix
\section*{Appendix}

We include additional experimental results and details of our method below. The supplementary material also includes videos of our synthetic and captured data.

\section{Additional Qualitative Results}

Qualitative examples of the ablation studies are shown in Figure \ref{fig:ablation} and Figure \ref{fig:inverse_ablation_2}.

Roughness examples are shown in Figure \ref{fig:inverse_roughness}.

Environment estimations are shown in Figure \ref{fig:inverse_env}.

\providelength\width
\setlength\width{3cm}

\providelength\oldtabcolsep
\setlength{\oldtabcolsep}{\tabcolsep}
\setlength{\tabcolsep}{1pt}

\begin{figure*}[t]
\centering
\footnotesize

\begin{tabular}{ccccc}
w/o \textit{RotLight} & w/o Proxy Mesh & w/o Residual & Ours & GT \\
\midrule
\includegraphics[width=\width]{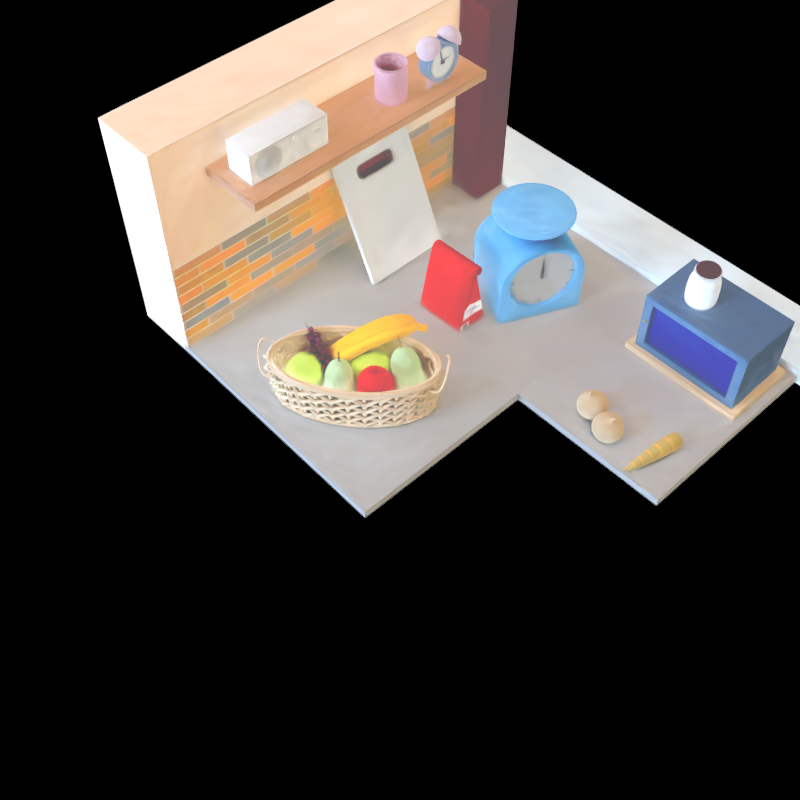} &
\includegraphics[width=\width]{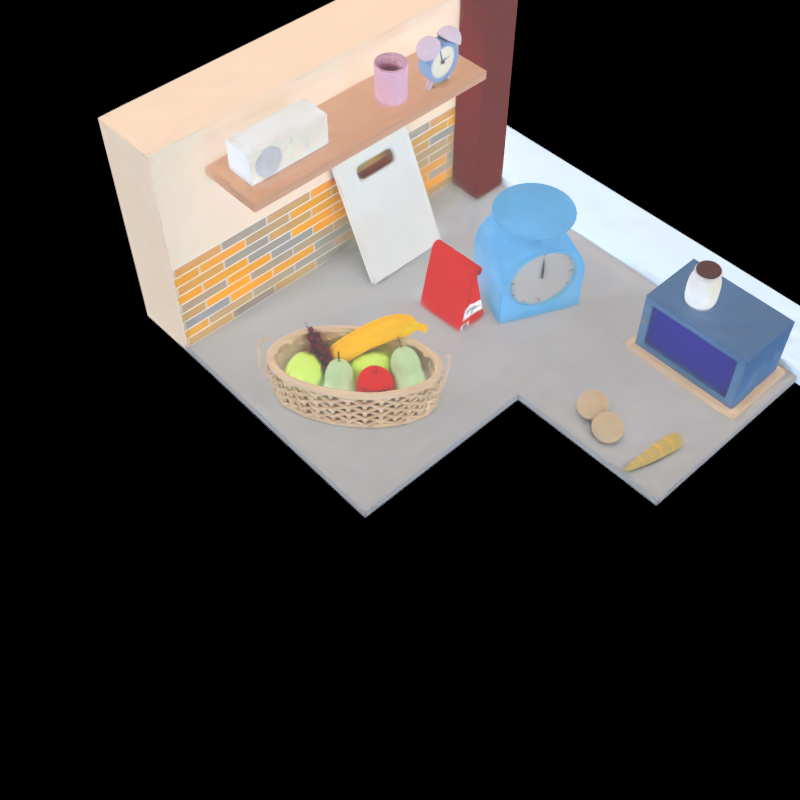} &
\includegraphics[width=\width]{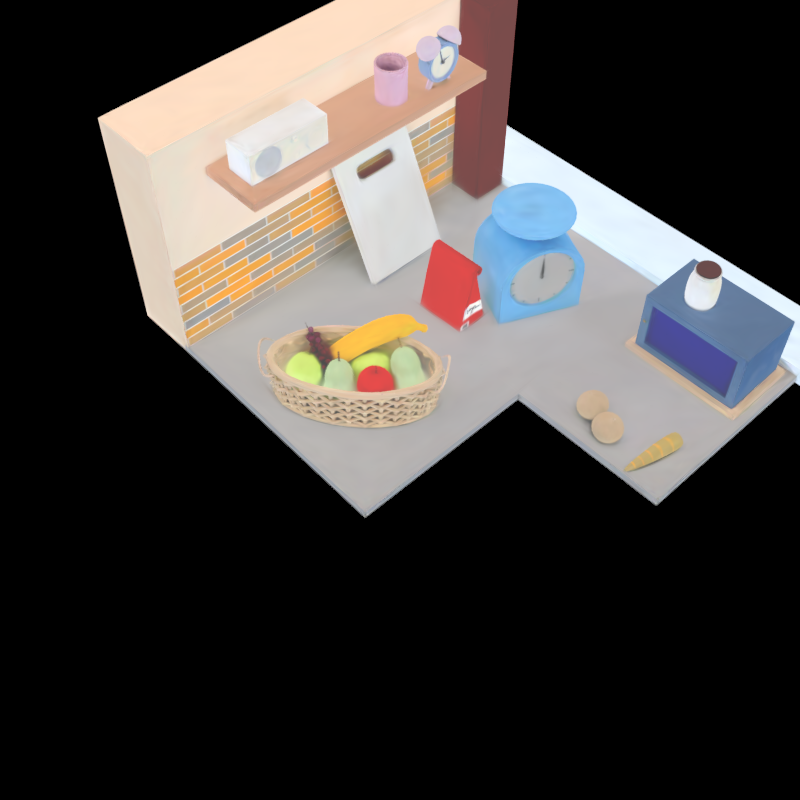} &
\includegraphics[width=\width]{image/albedo/counter/108_ours.png} &
\includegraphics[width=\width]{image/albedo/counter/108_gt.png} \\
\includegraphics[width=\width]{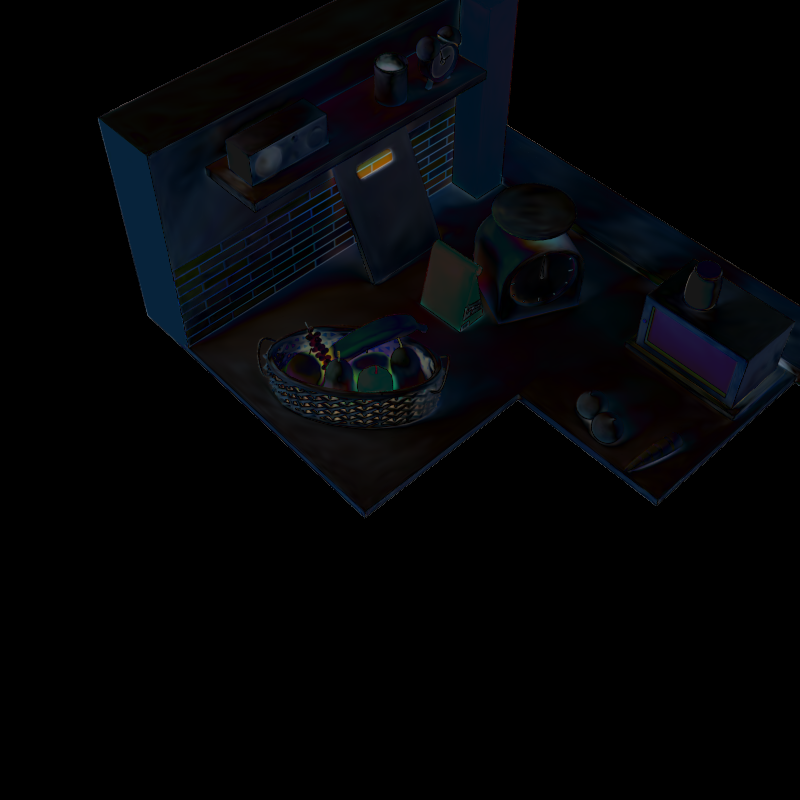} &
\includegraphics[width=\width]{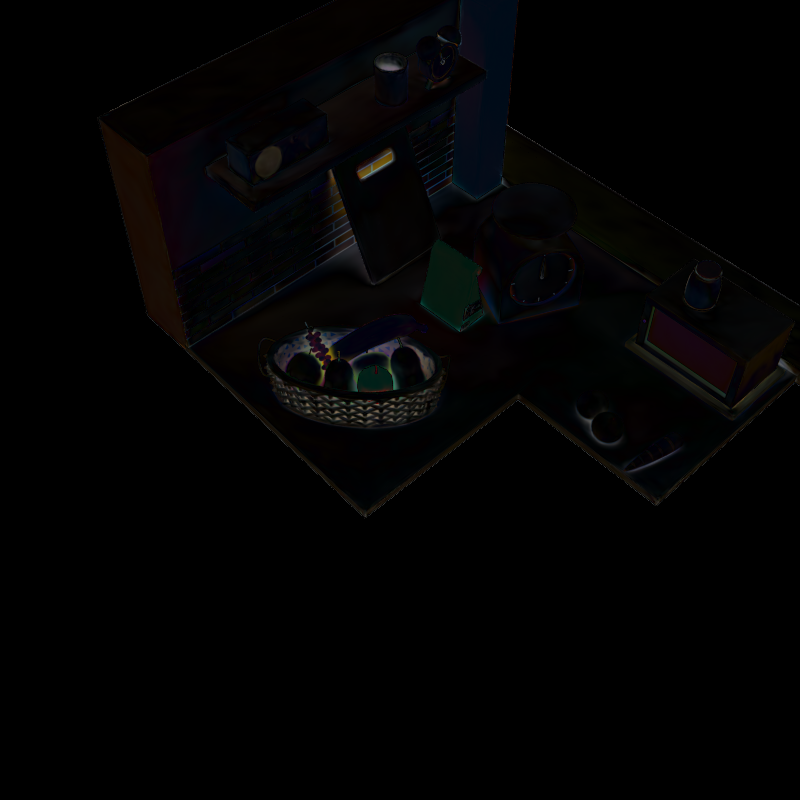} &
\includegraphics[width=\width]{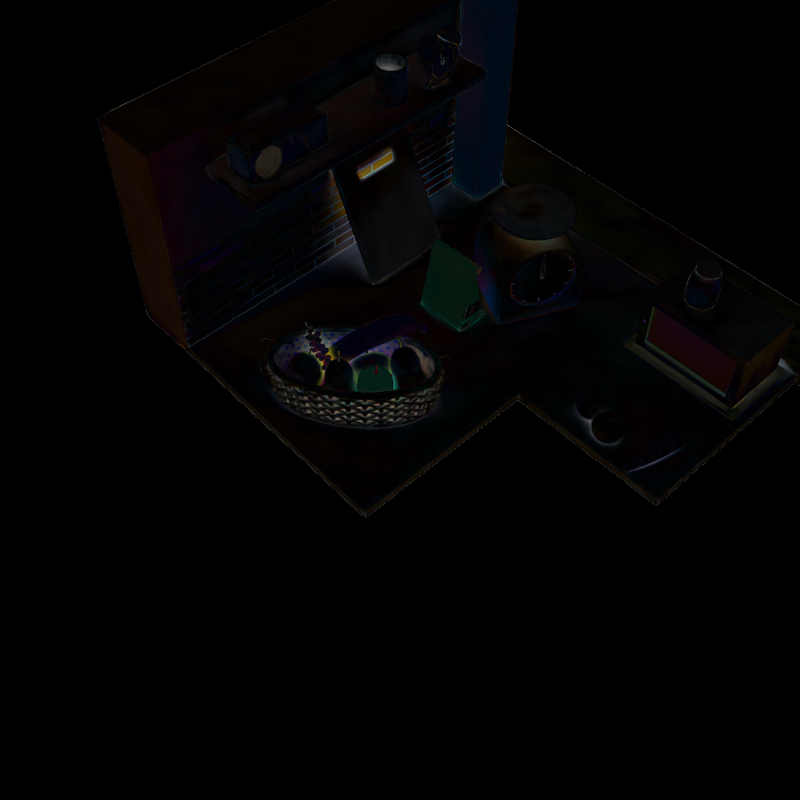} &
\includegraphics[width=\width]{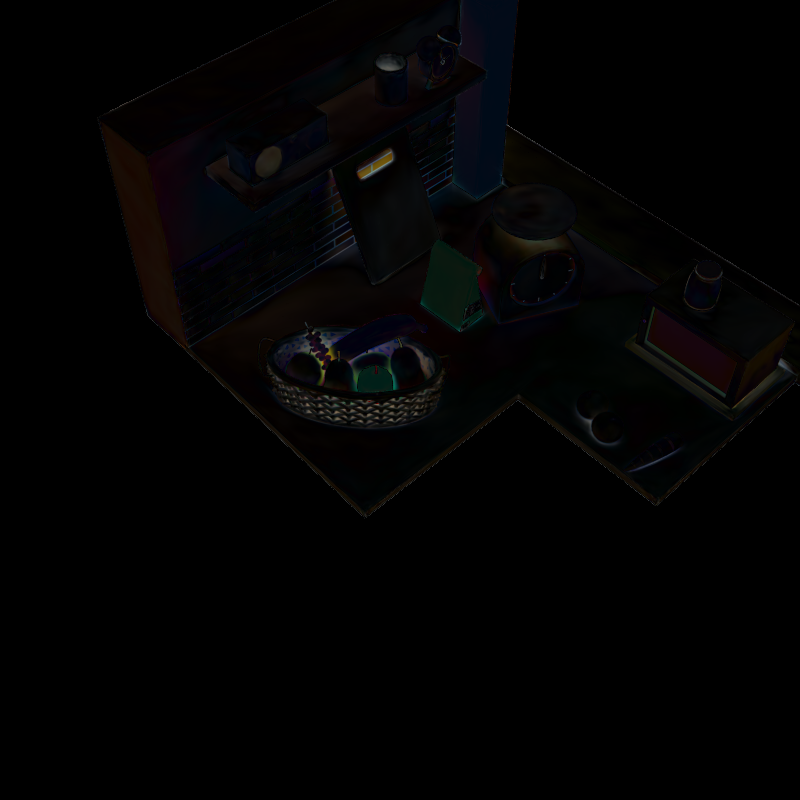} &
 \\
22.204 / 0.918 & 27.279 / 0.955 & 27.481 / 0.955 & 27.722 / 0.955 &  \\
\includegraphics[width=\width]{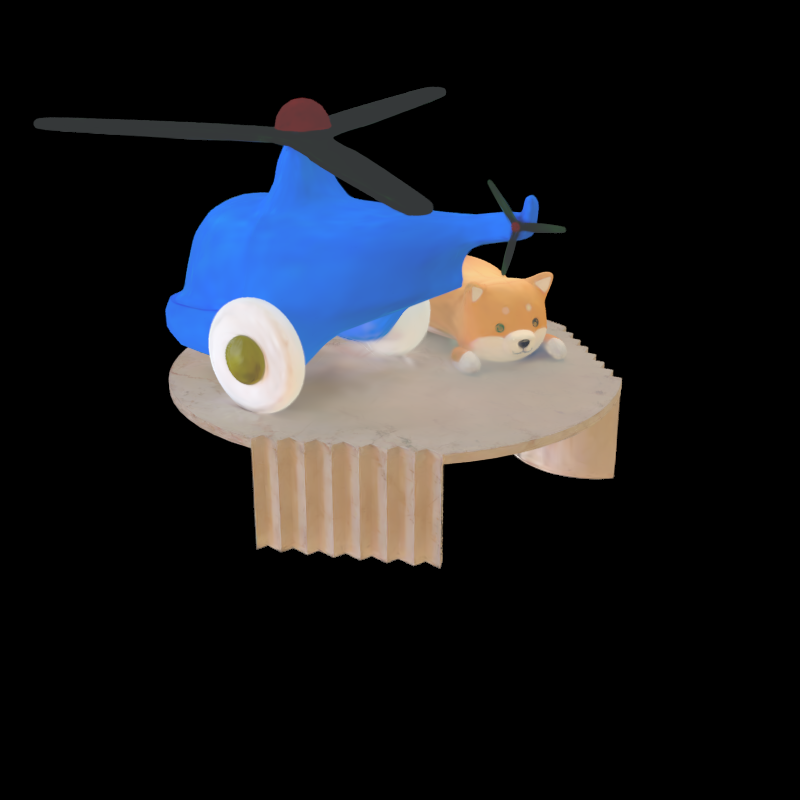} &
\includegraphics[width=\width]{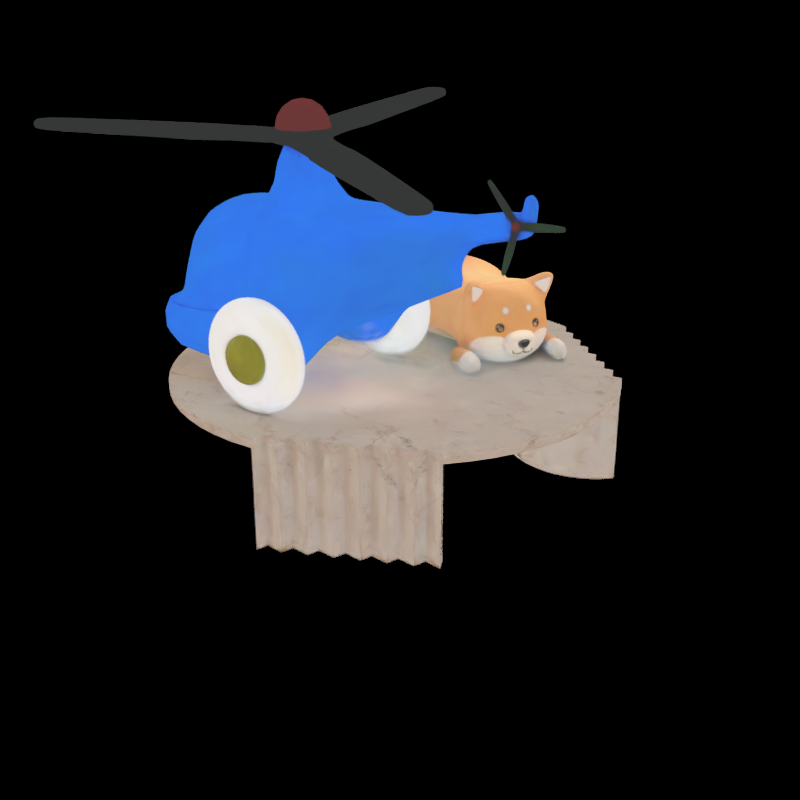} &
\includegraphics[width=\width]{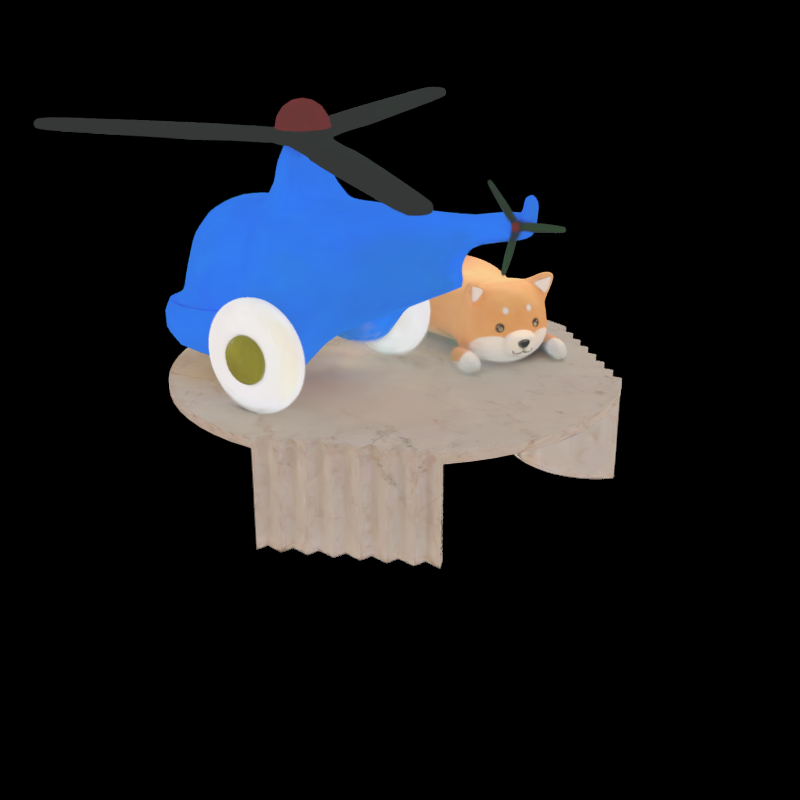} &
\includegraphics[width=\width]{image/albedo/toy/036_ours.png} &
\includegraphics[width=\width]{image/albedo/toy/036_gt.png} \\
\includegraphics[width=\width]{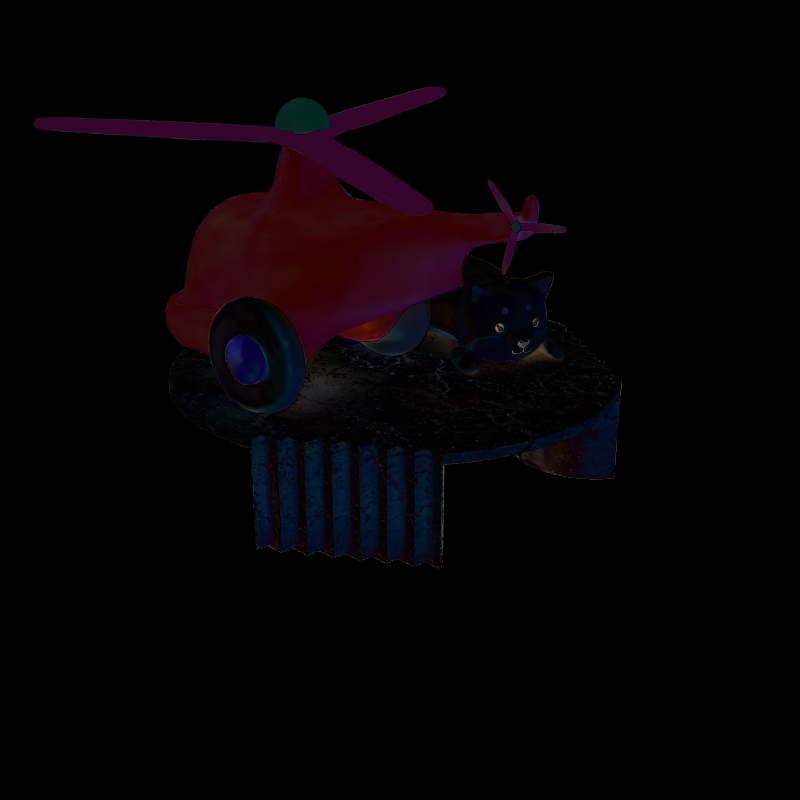} &
\includegraphics[width=\width]{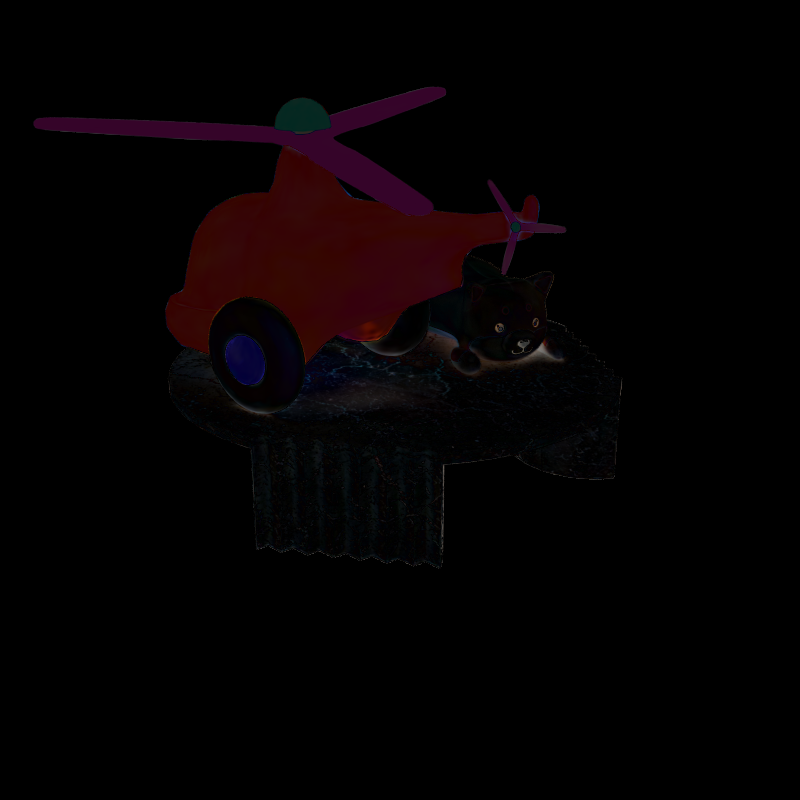} &
\includegraphics[width=\width]{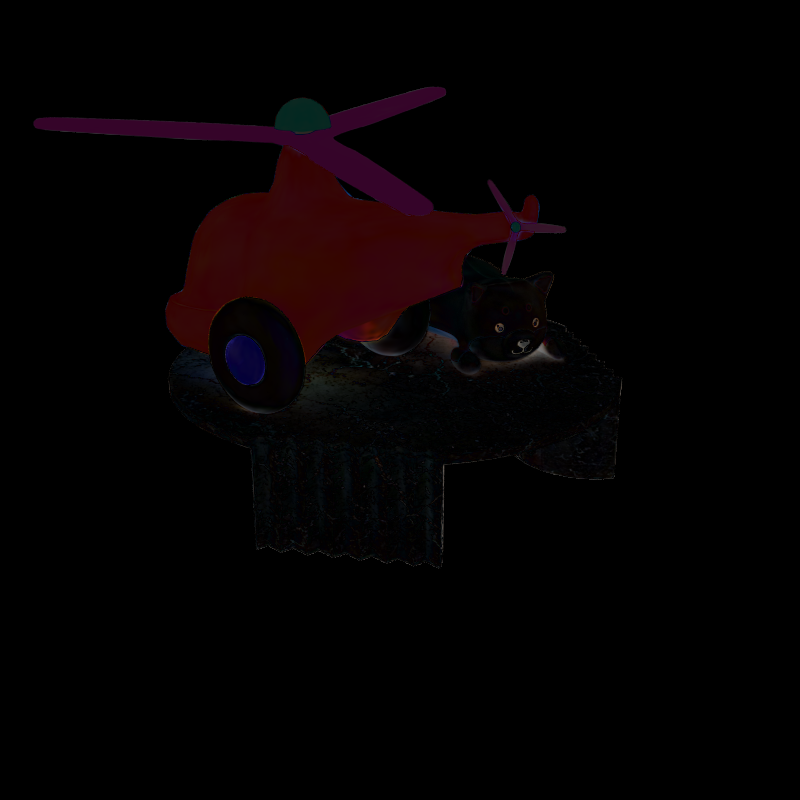} &
\includegraphics[width=\width]{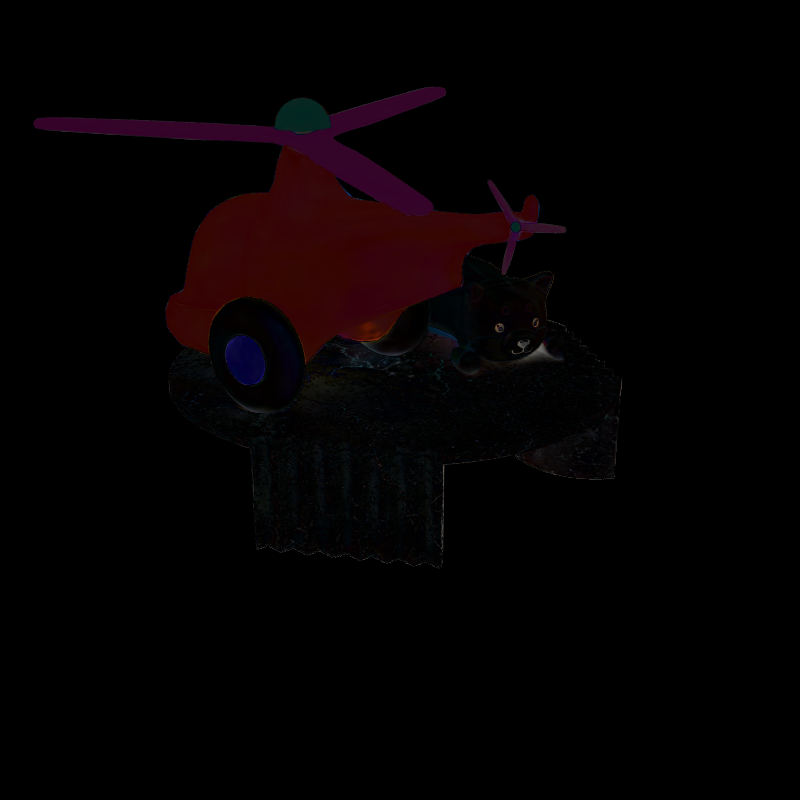} &
 \\
29.880 / 0.929 & 32.982 / 0.940 & 33.347 / 0.941 & 34.206 / 0.941 &  \\
\includegraphics[width=\width]{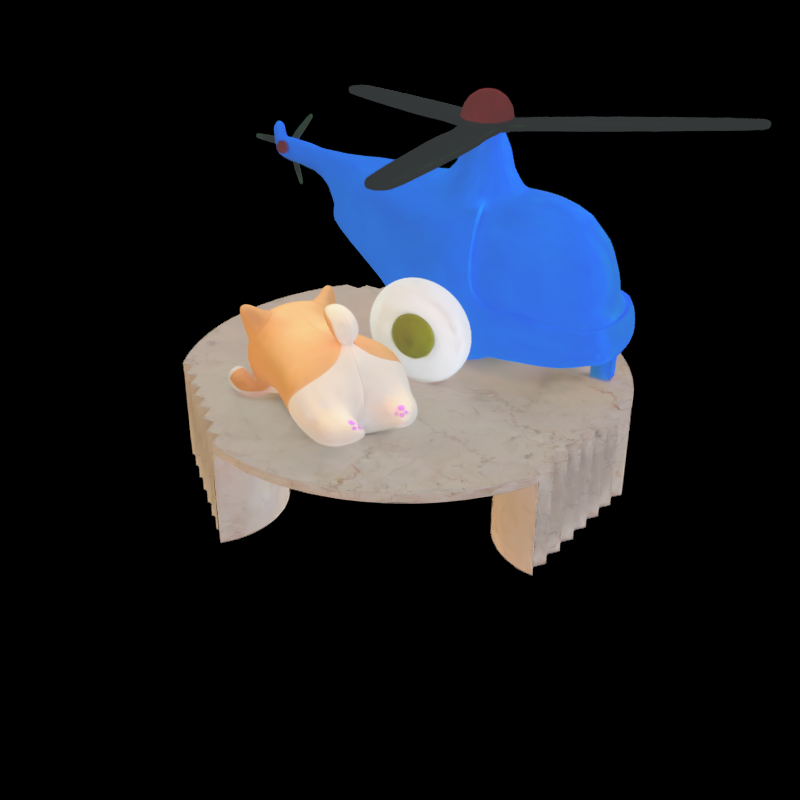} &
\includegraphics[width=\width]{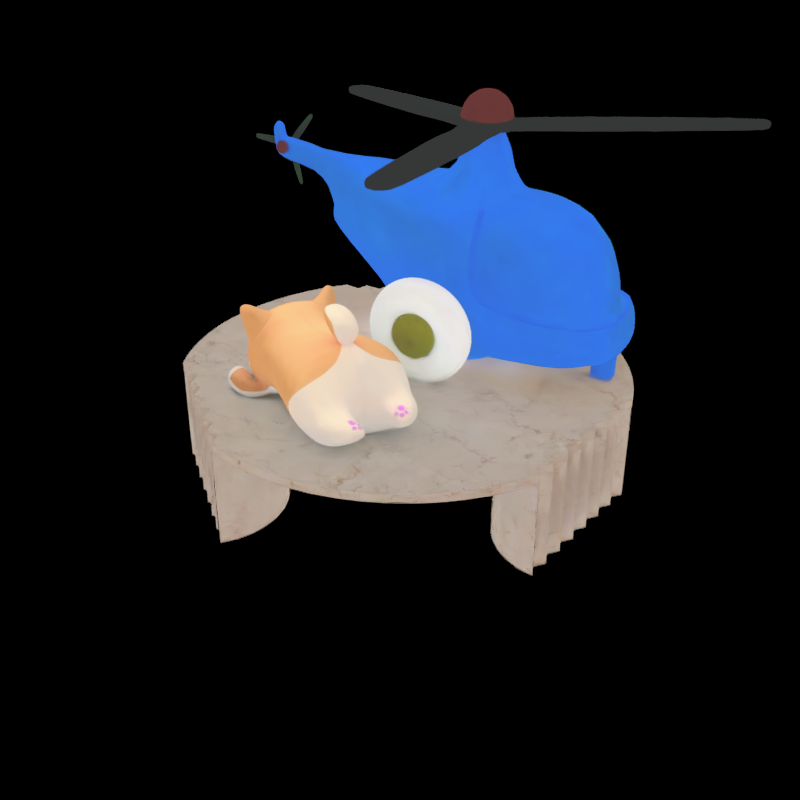} &
\includegraphics[width=\width]{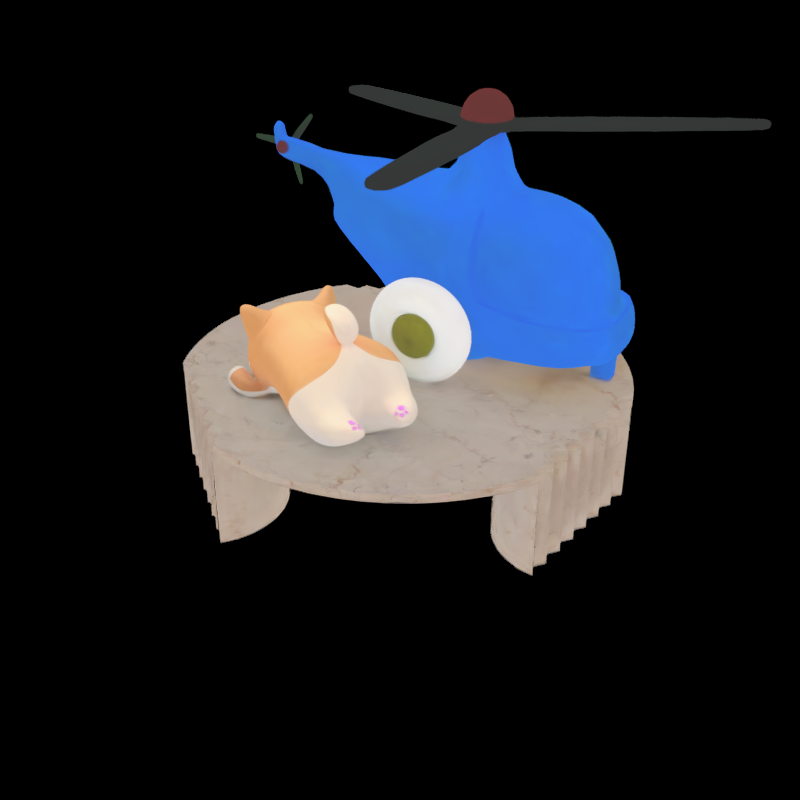} &
\includegraphics[width=\width]{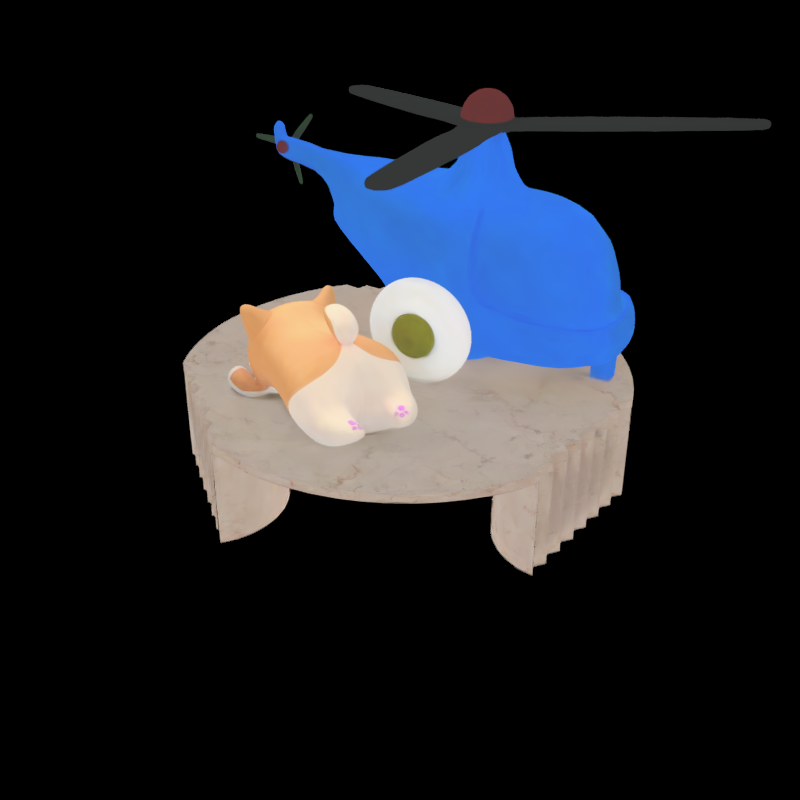} &
\includegraphics[width=\width]{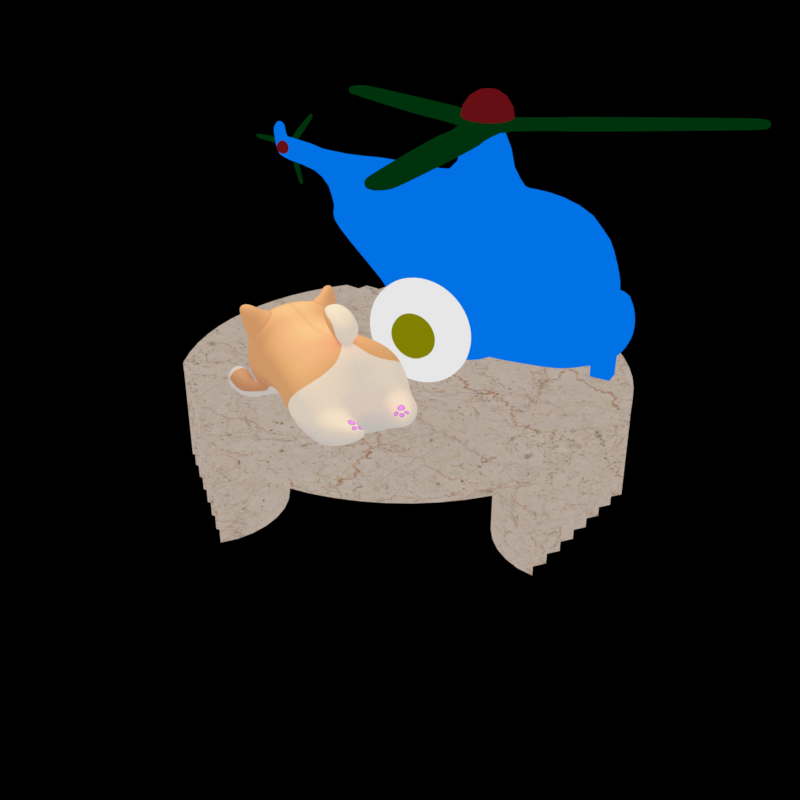} \\
\includegraphics[width=\width]{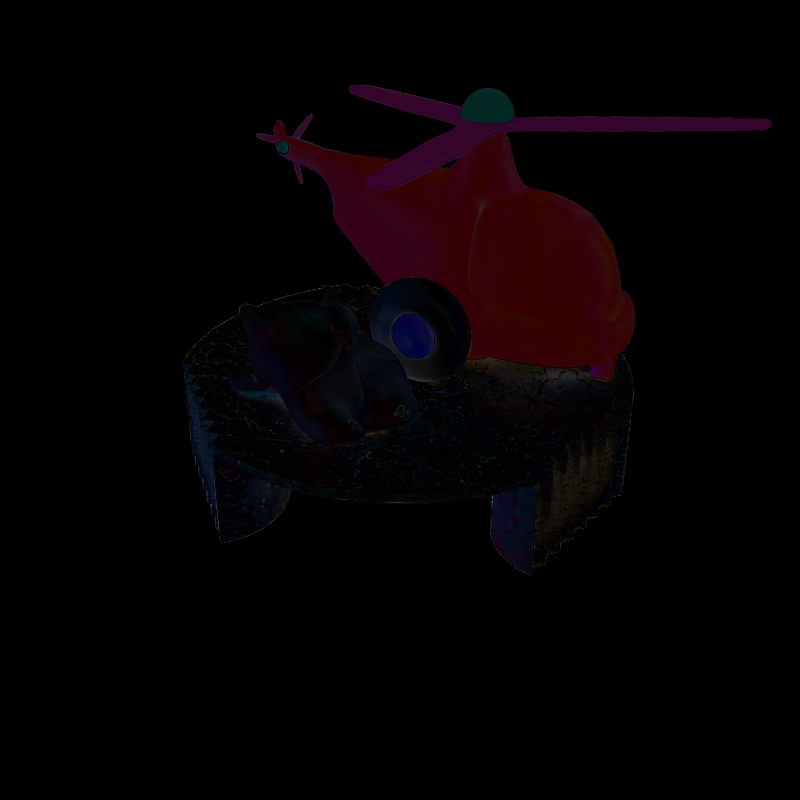} &
\includegraphics[width=\width]{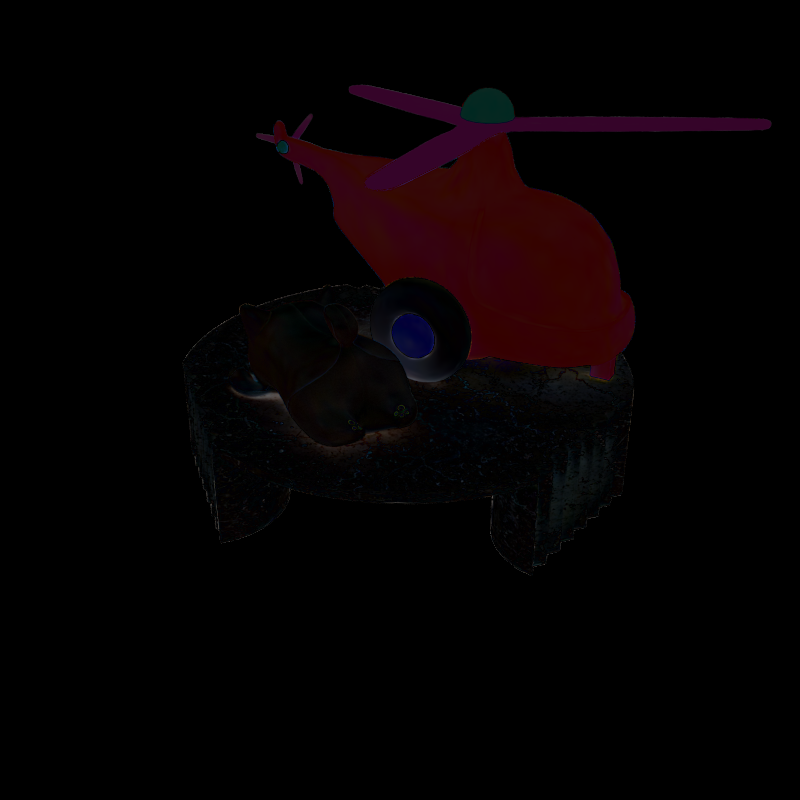} &
\includegraphics[width=\width]{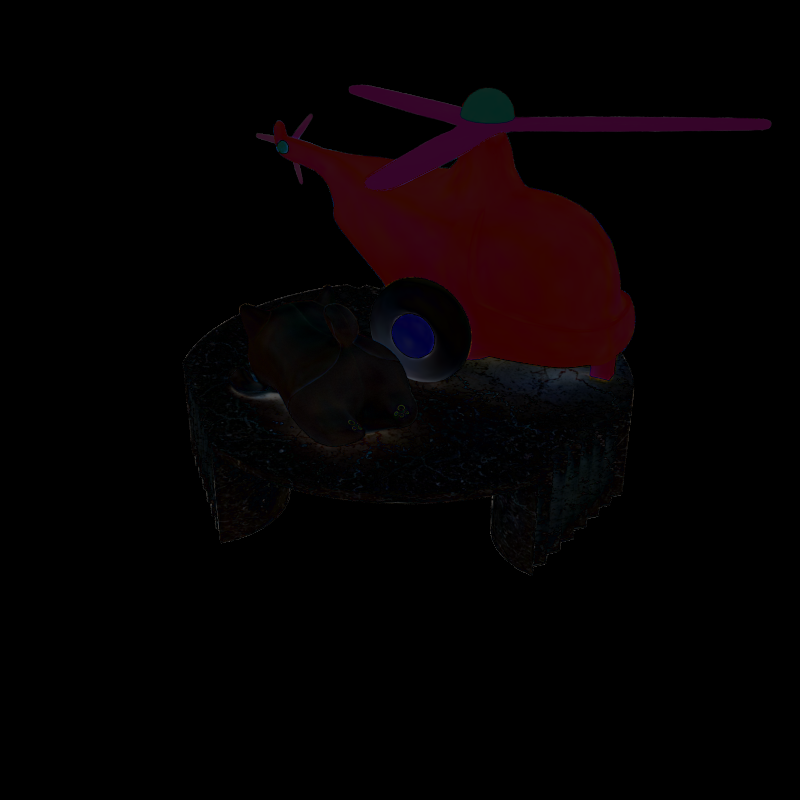} &
\includegraphics[width=\width]{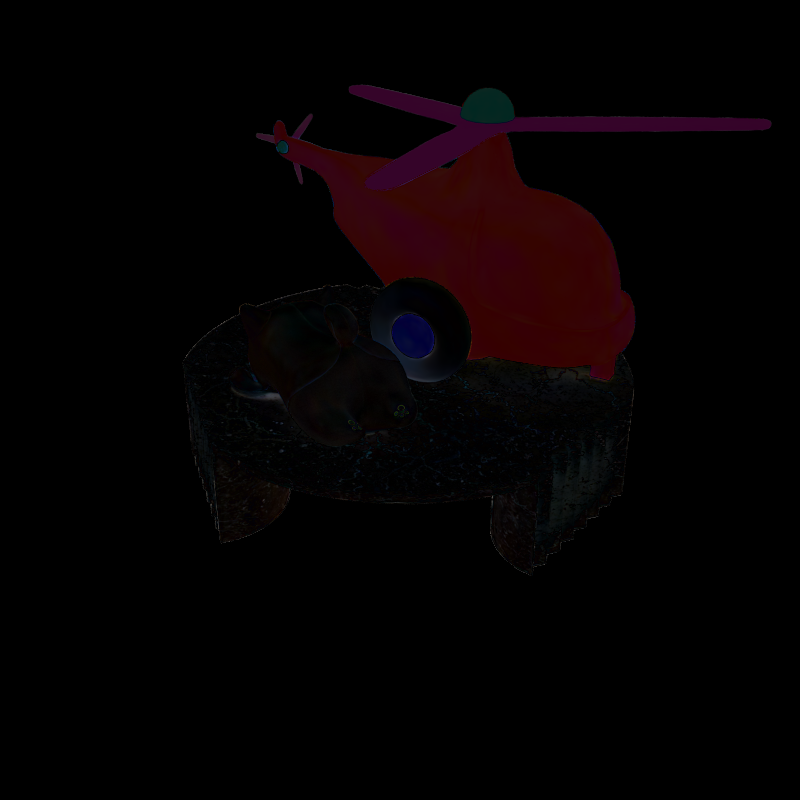} &
 \\
31.984 / 0.939 & 32.943 / 0.942 & 32.620 / 0.942 & 33.540 / 0.942 & \\
\end{tabular}

\caption{Ablation study. Even rows show error maps. Additional examples shown in Figure \ref{fig:inverse_ablation_2}. Without residual constraint, the estimated albedo has more baked shadow where incident illumination presents from locations not directly visible from the camera. \textit{counter}: under the chopping board, between the red bag and blue scale. }
\label{fig:ablation}

\end{figure*}

\setlength\tabcolsep{\oldtabcolsep}

\providelength\width
\setlength\width{3cm}

\providelength\oldtabcolsep
\setlength{\oldtabcolsep}{\tabcolsep}
\setlength{\tabcolsep}{1pt}

\begin{figure*}[t]
\centering
\footnotesize

\begin{tabular}{ccccc}
w/o \textit{RotLight} & w/o Proxy Mesh & w/o Residual & Ours & GT \\
\midrule
\includegraphics[width=\width]{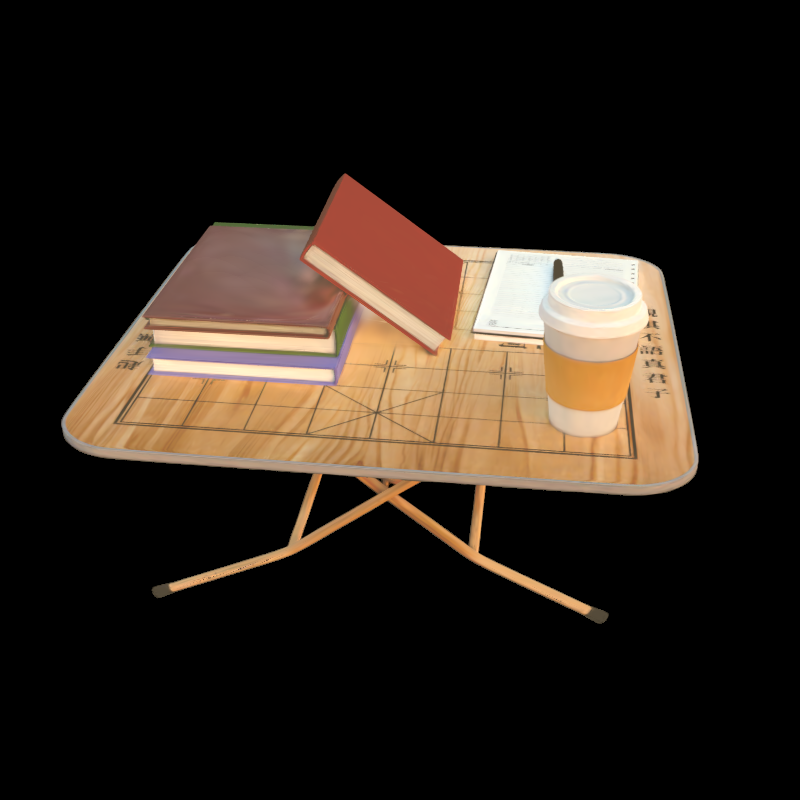} &
\includegraphics[width=\width]{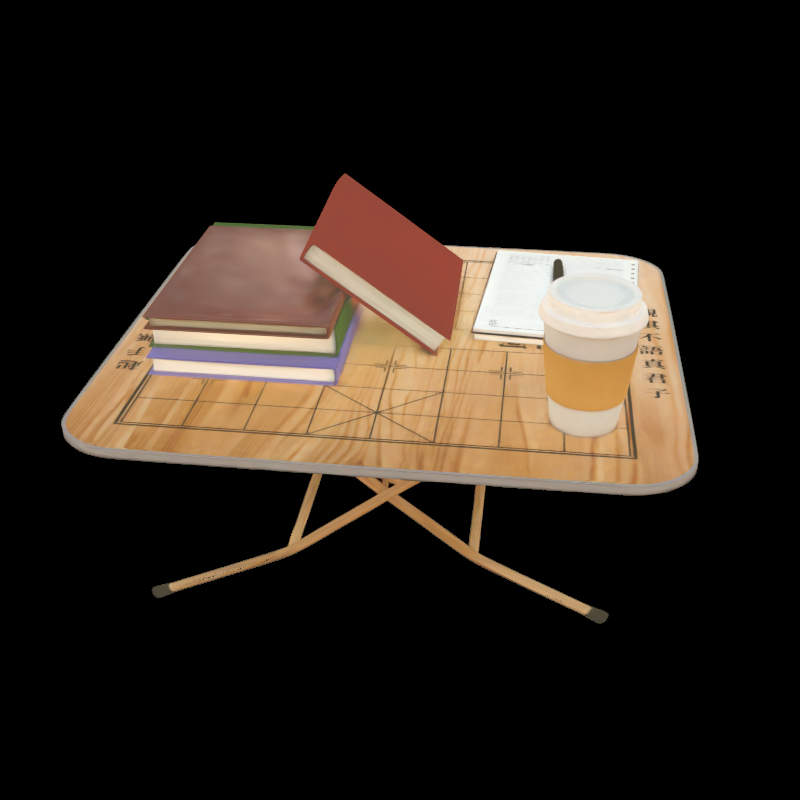} &
\includegraphics[width=\width]{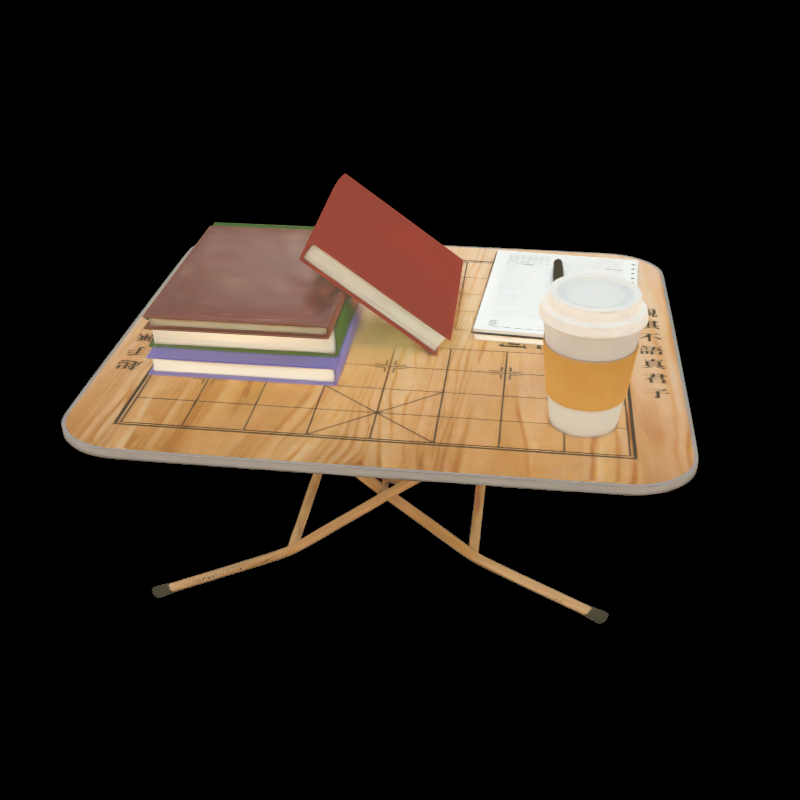} &
\includegraphics[width=\width]{image/albedo/table/057_ours.png} &
\includegraphics[width=\width]{image/albedo/table/057_gt.png} \\
\includegraphics[width=\width]{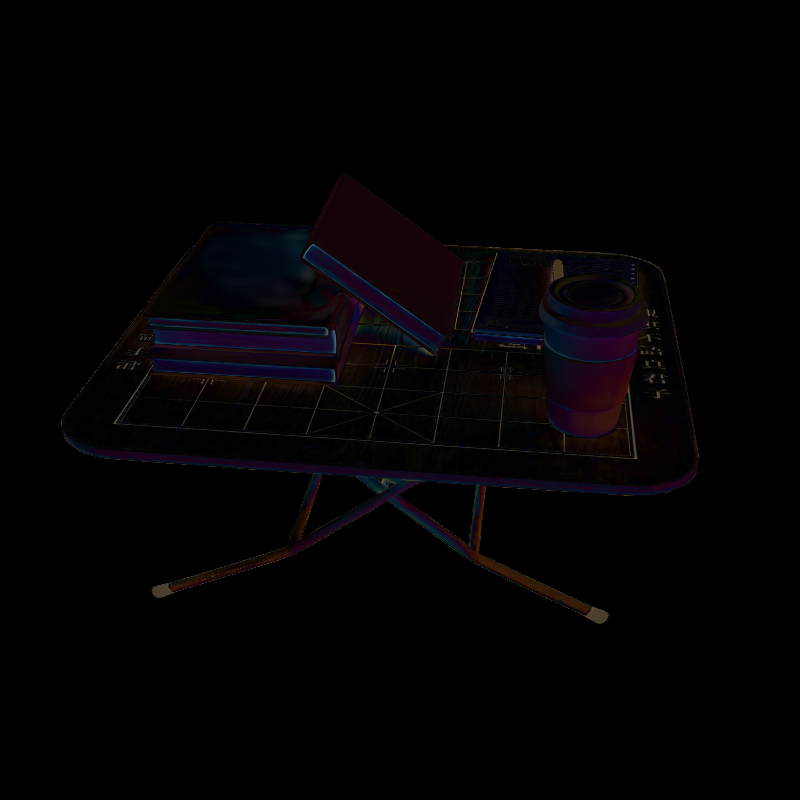} &
\includegraphics[width=\width]{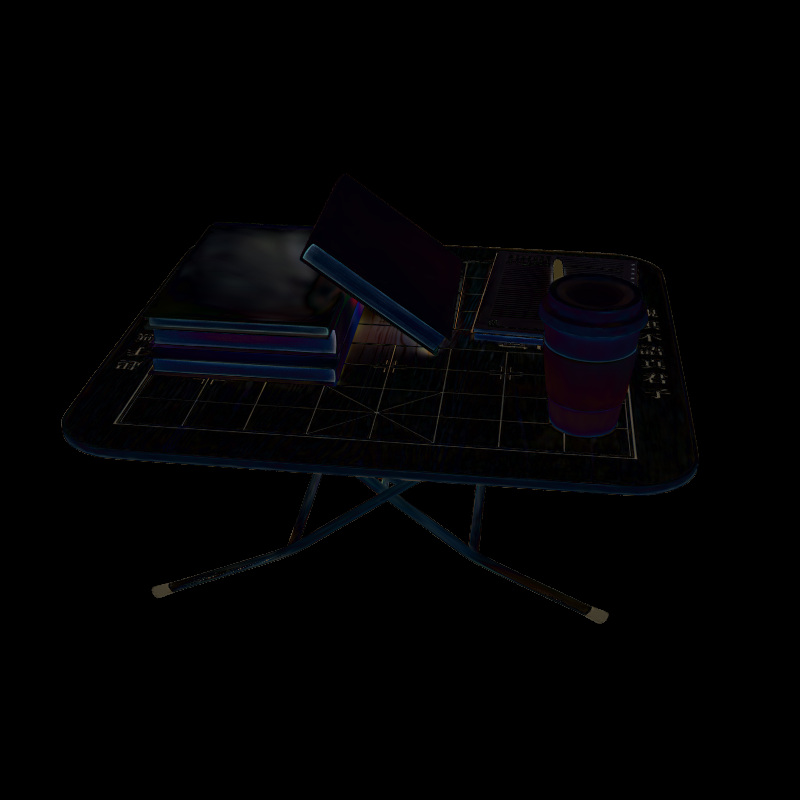} &
\includegraphics[width=\width]{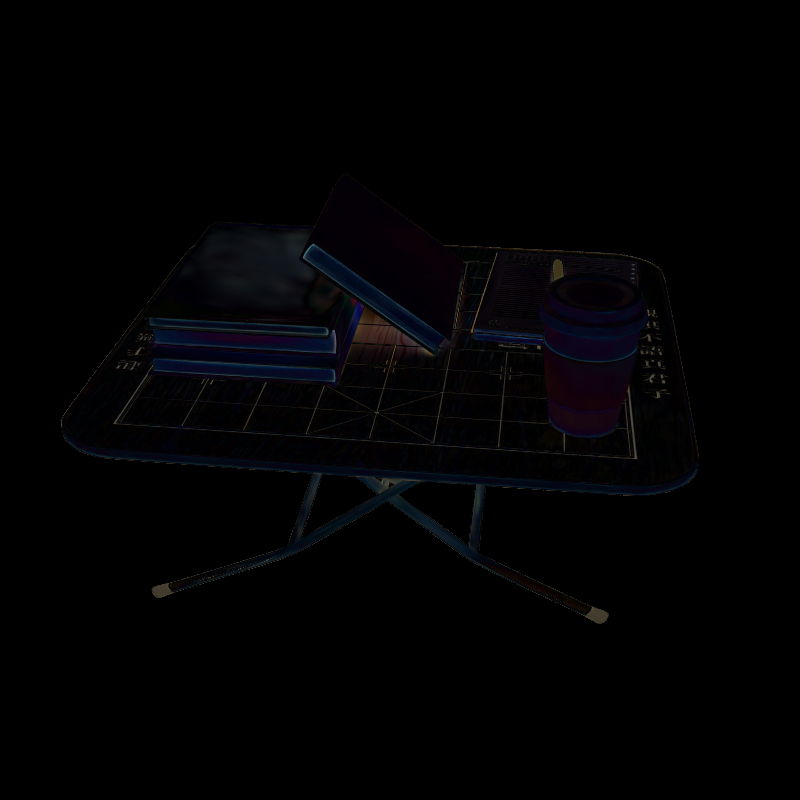} &
\includegraphics[width=\width]{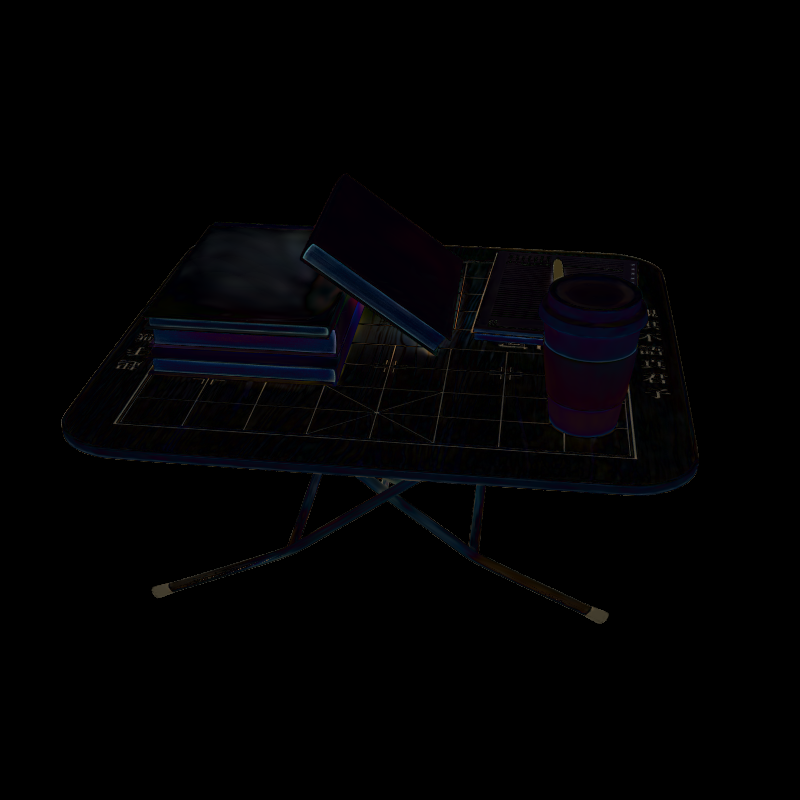} & \\
\includegraphics[width=\width]{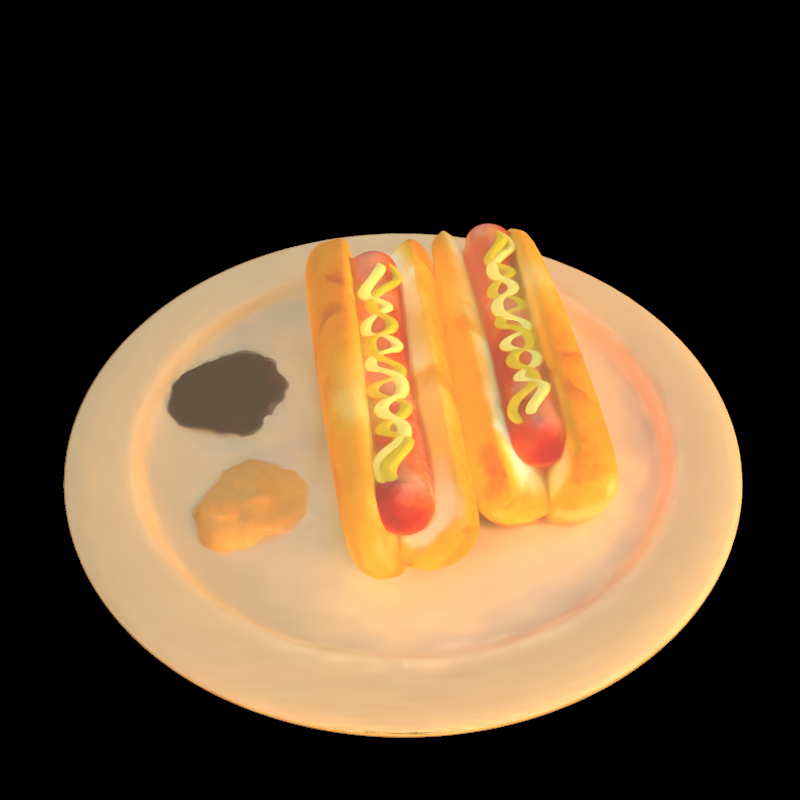} &
\includegraphics[width=\width]{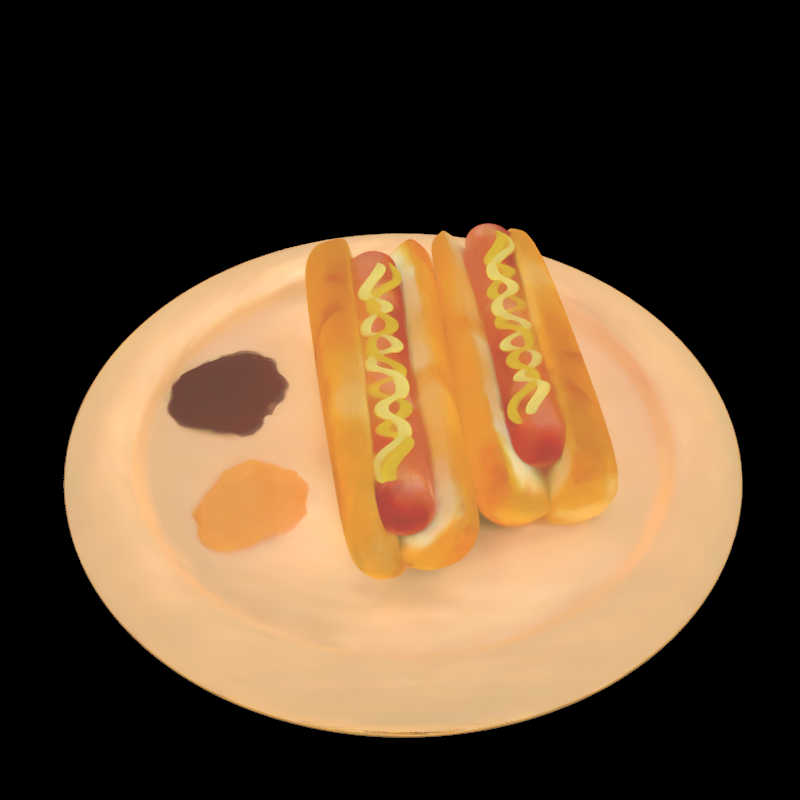} &
\includegraphics[width=\width]{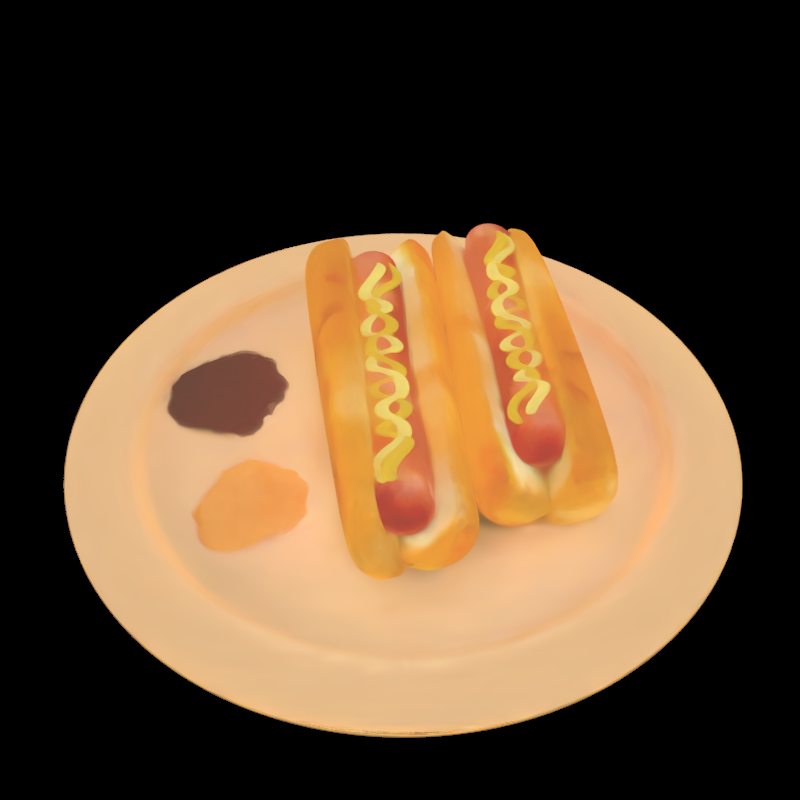} &
\includegraphics[width=\width]{image/albedo/hotdog/093_ours.png} &
\includegraphics[width=\width]{image/albedo/hotdog/093_gt.png} \\
\includegraphics[width=\width]{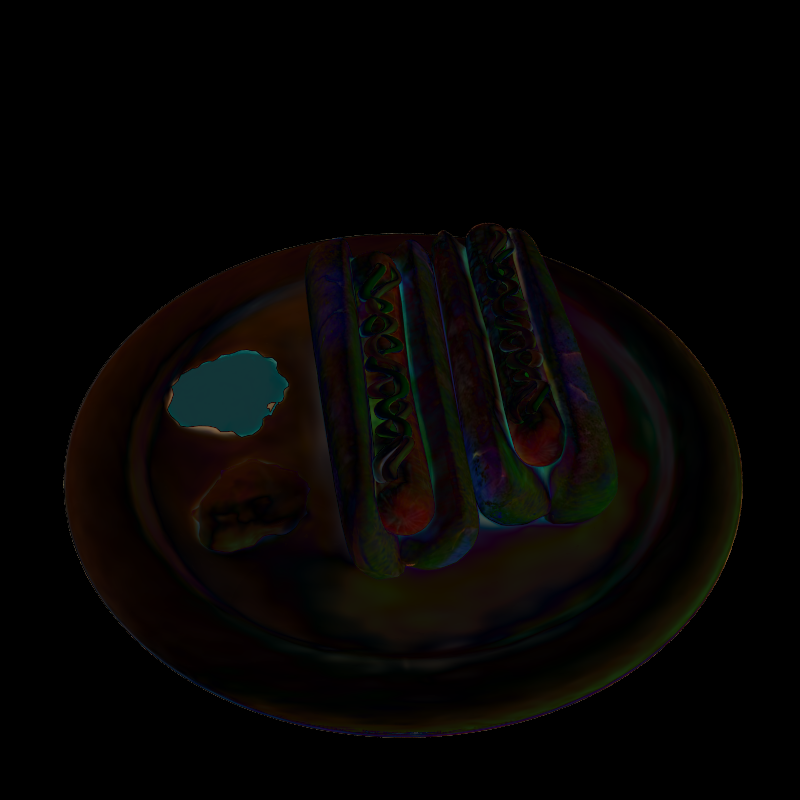} &
\includegraphics[width=\width]{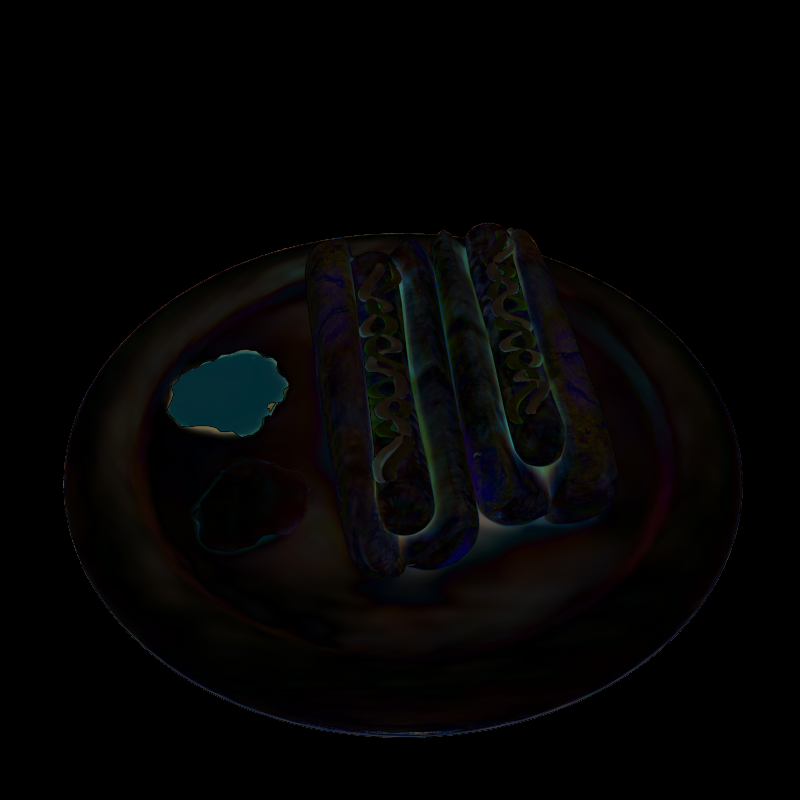} &
\includegraphics[width=\width]{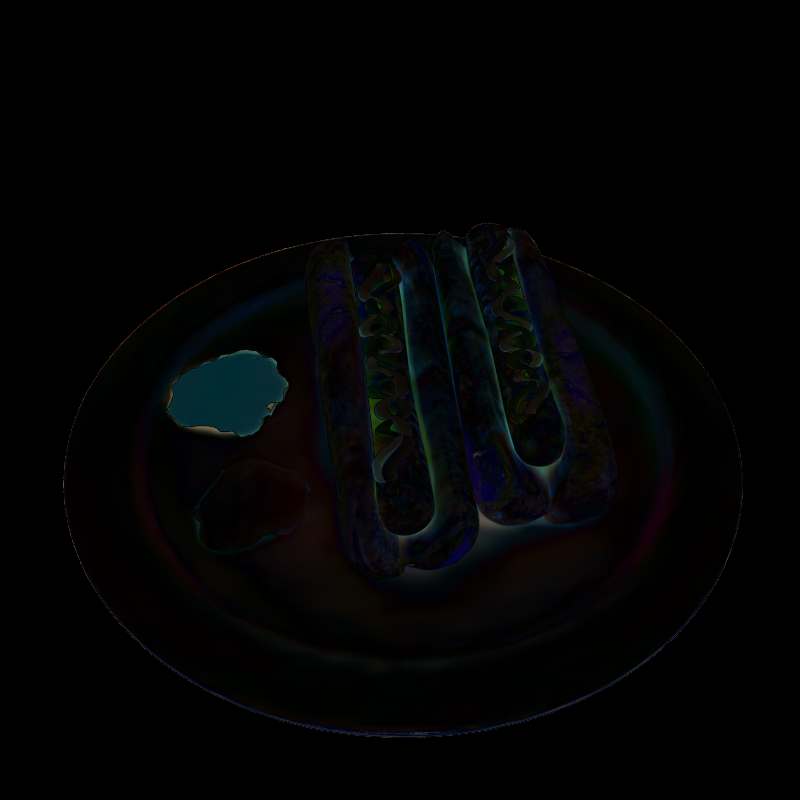} &
\includegraphics[width=\width]{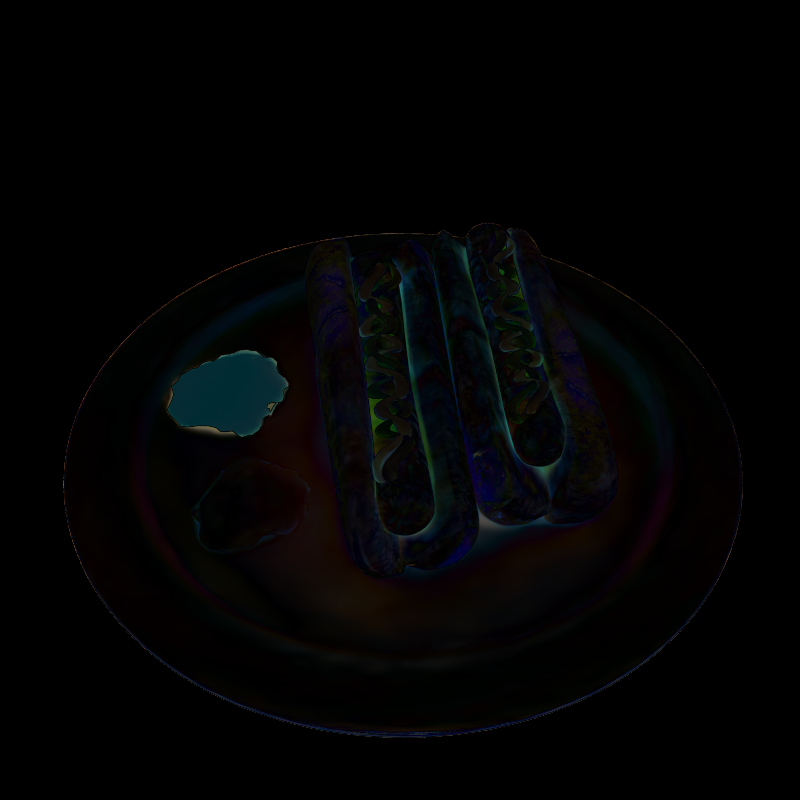} & \\
\includegraphics[width=\width]{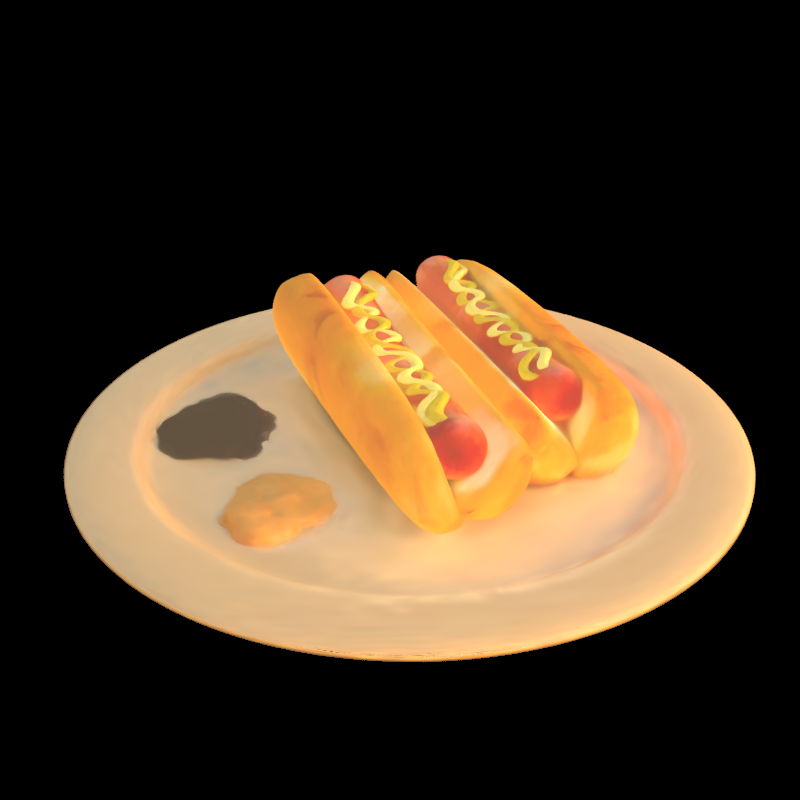} &
\includegraphics[width=\width]{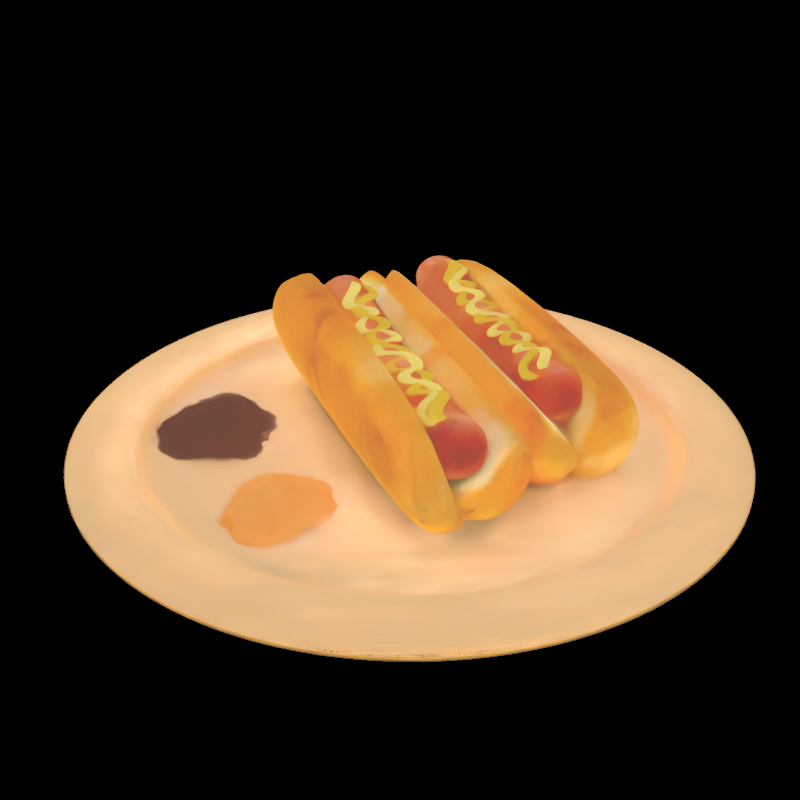} &
\includegraphics[width=\width]{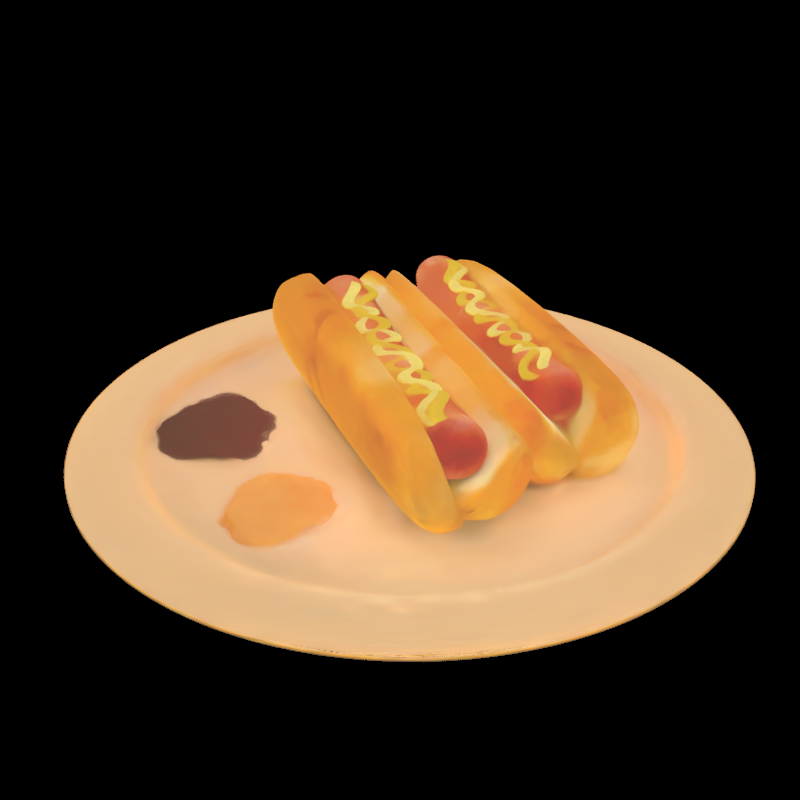} &
\includegraphics[width=\width]{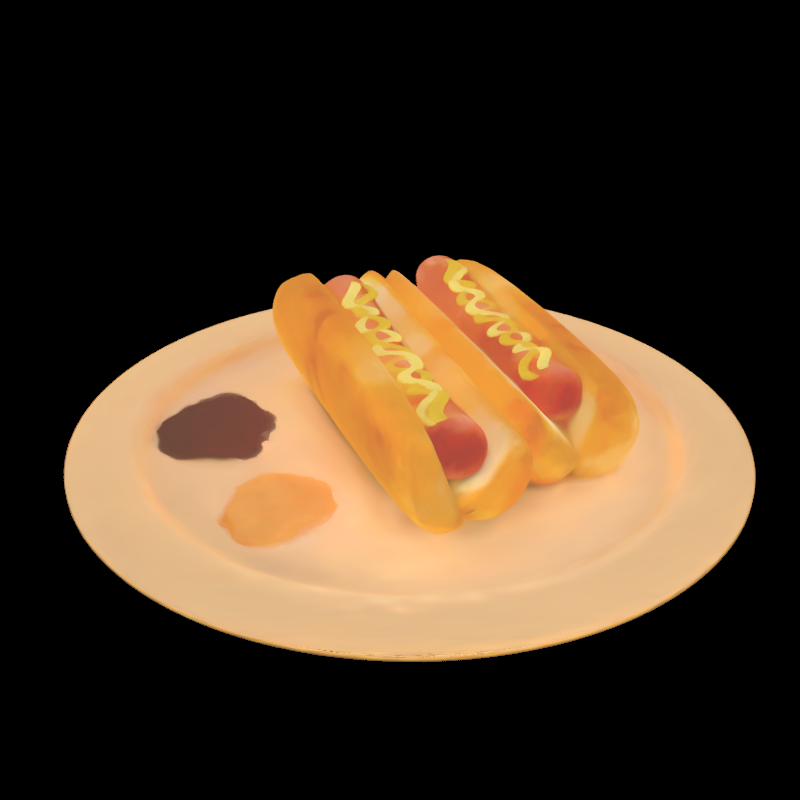} &
\includegraphics[width=\width]{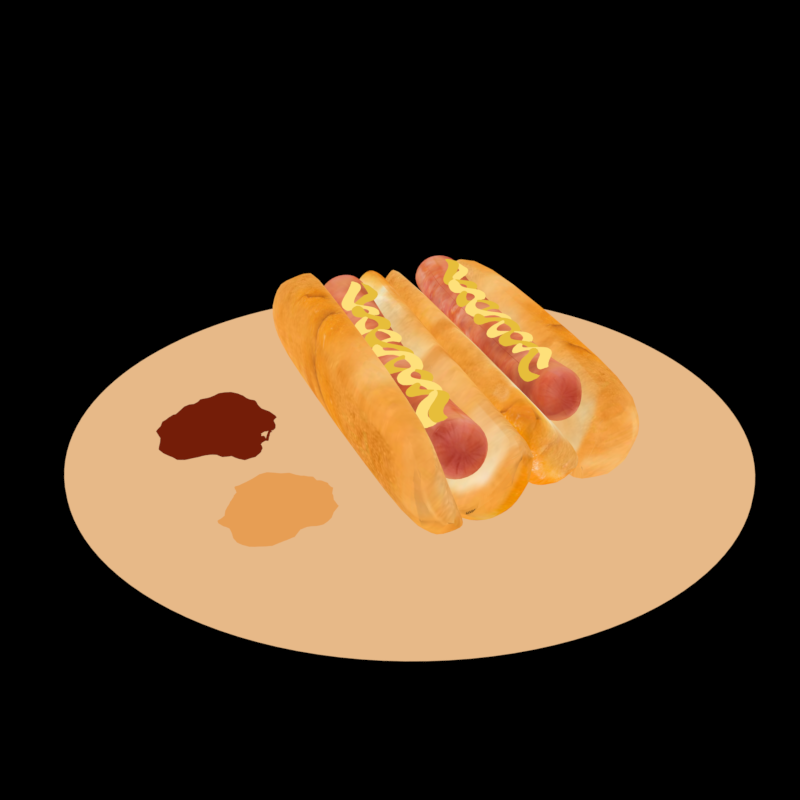} \\
\includegraphics[width=\width]{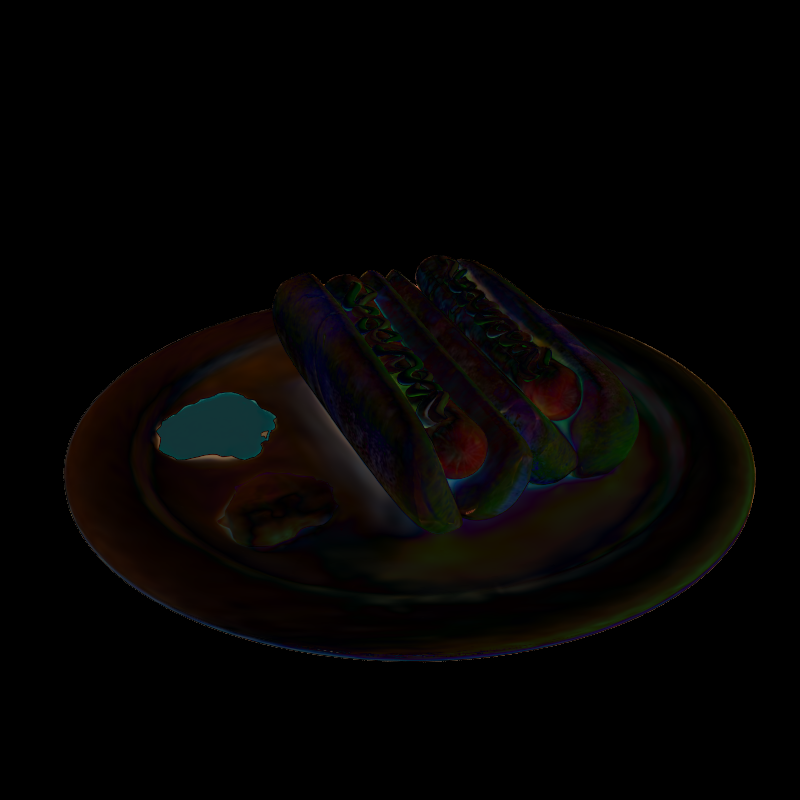} &
\includegraphics[width=\width]{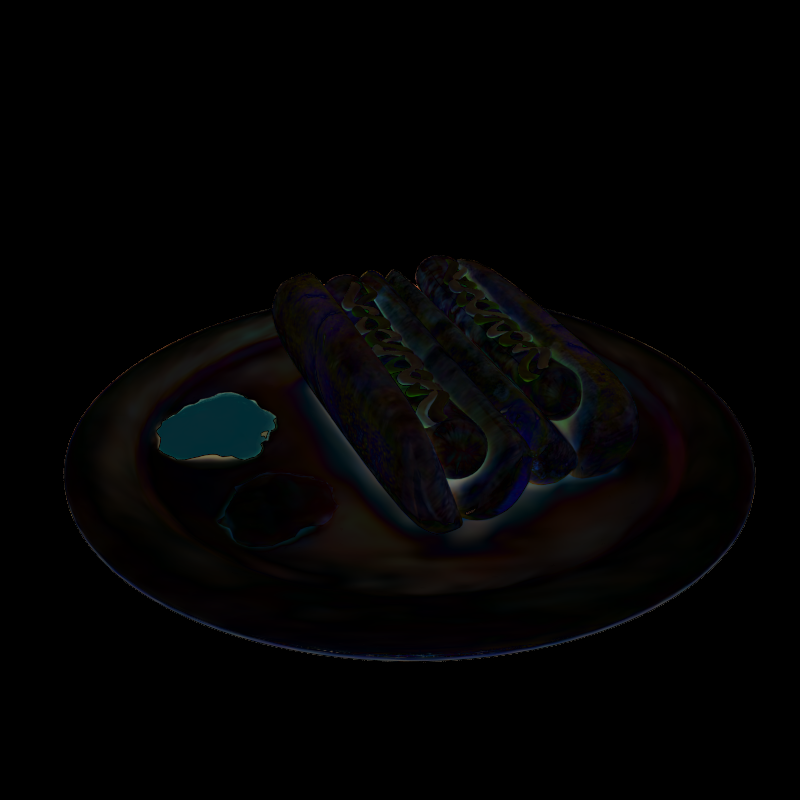} &
\includegraphics[width=\width]{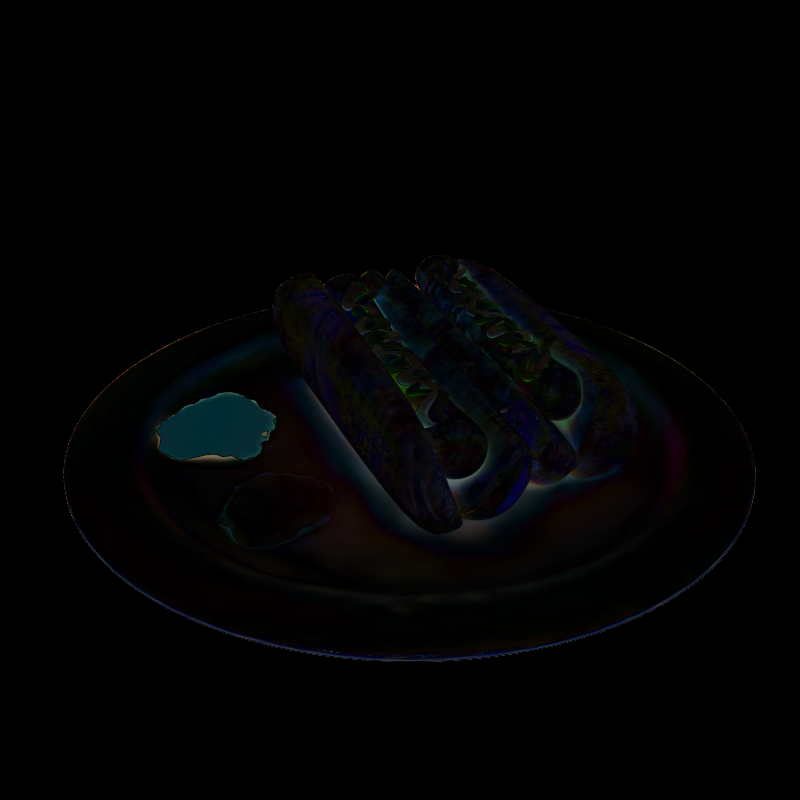} &
\includegraphics[width=\width]{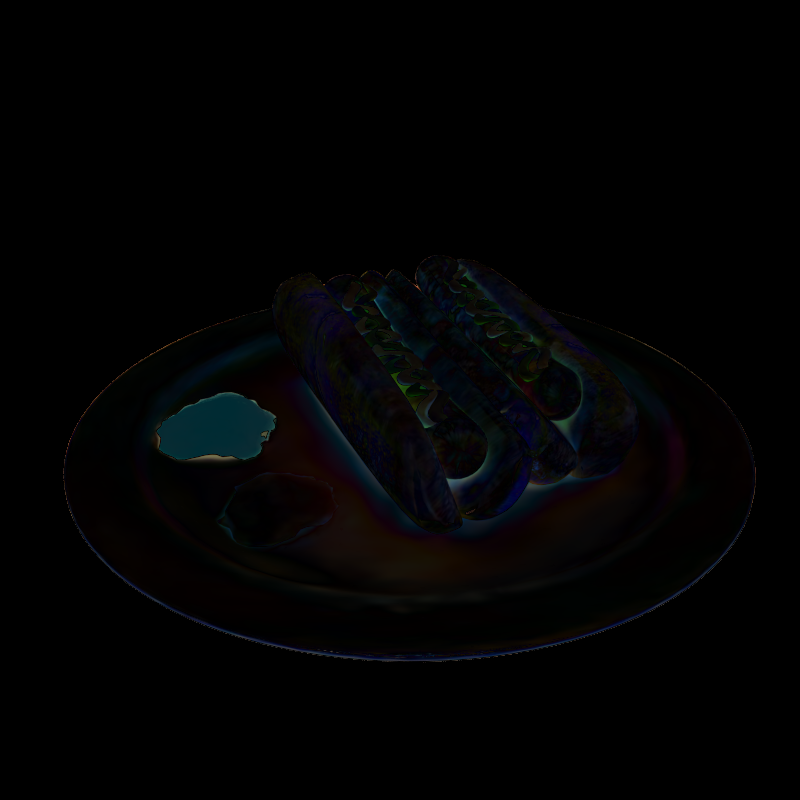} &
\end{tabular}

\caption{Ablation study of more scenes. Baked shadows in albedo can be found without residual constraint. \textit{toy}: under the helicopter and dog. \textit{table}: between the books. \textit{hotdog}: under the front end of the hotdog.}
\label{fig:inverse_ablation_2}

\end{figure*}

\setlength\tabcolsep{\oldtabcolsep}

\providelength\width
\setlength\width{2.6cm}

\providelength\oldtabcolsep
\setlength{\oldtabcolsep}{\tabcolsep}
\setlength{\tabcolsep}{1pt}

\begin{figure}[t]
\centering
\footnotesize

\begin{tabular}{ccc}
IRGS & Ours & GT \\
\midrule
\includegraphics[width=\width]{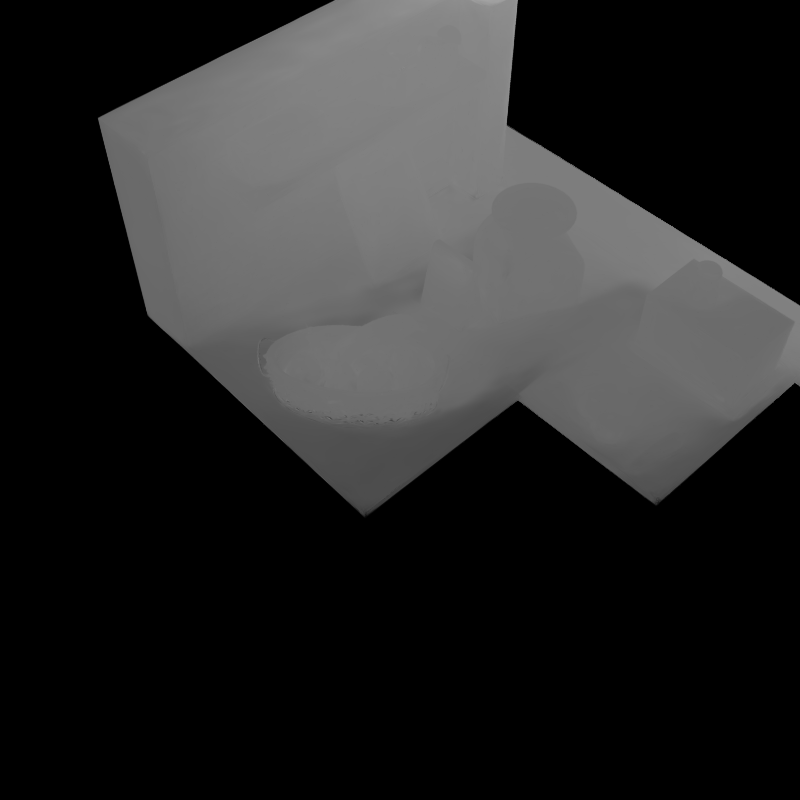} &
\includegraphics[width=\width]{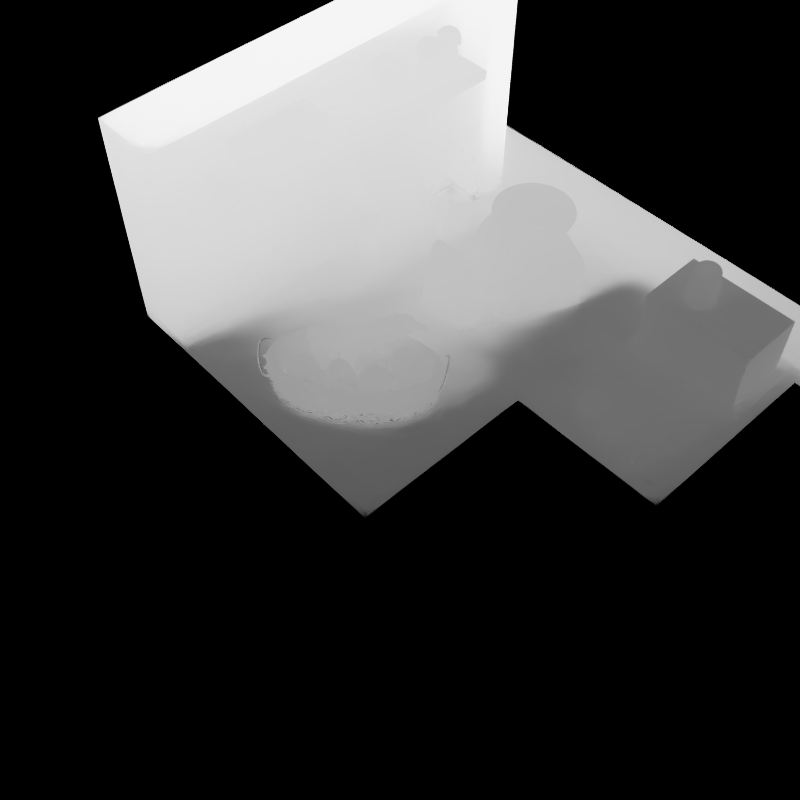} &
\includegraphics[width=\width]{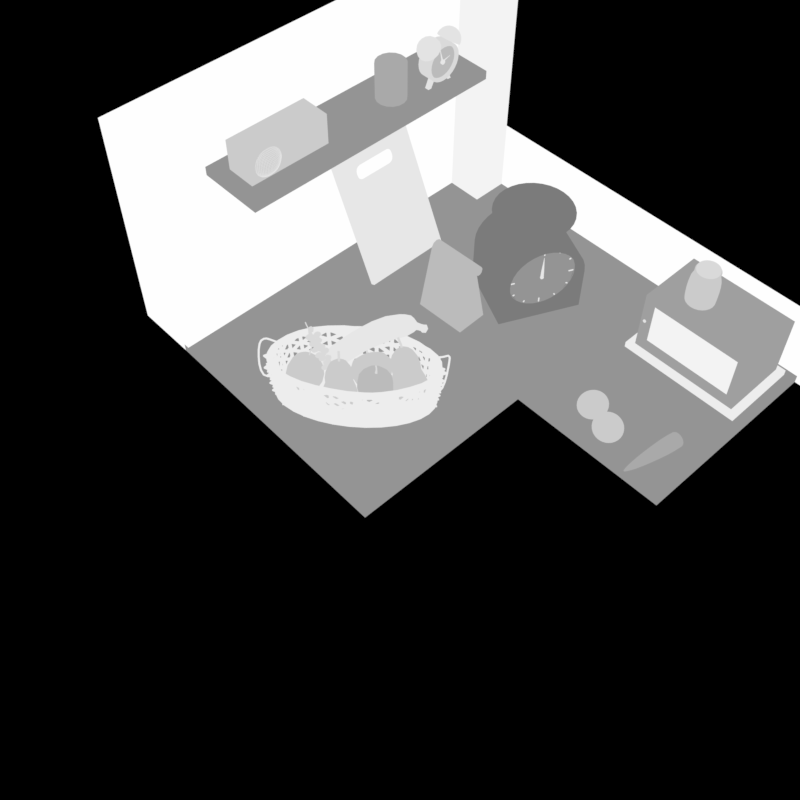} \\
\includegraphics[width=\width]{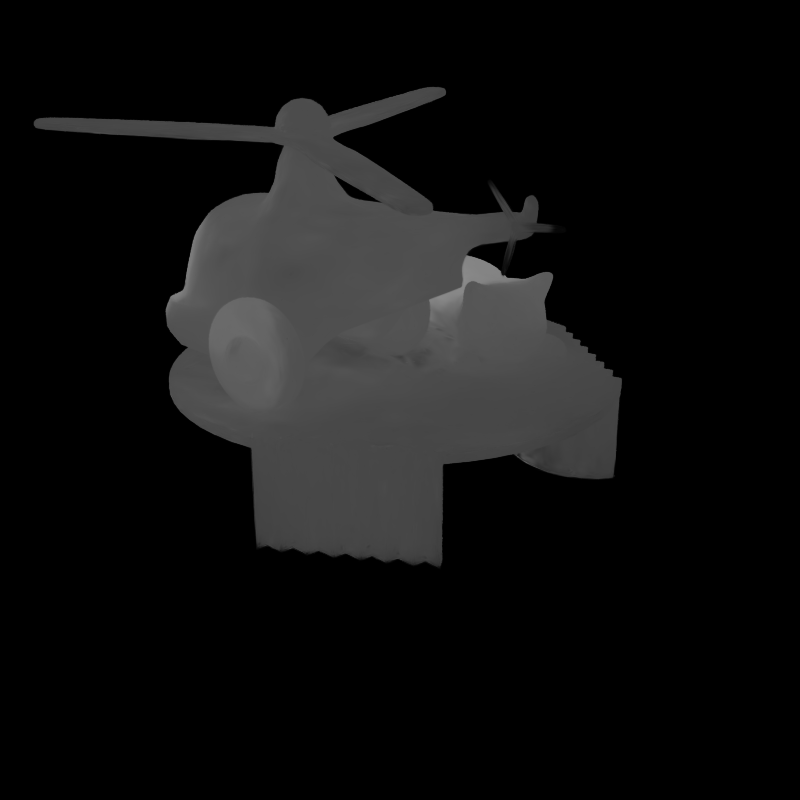} &
\includegraphics[width=\width]{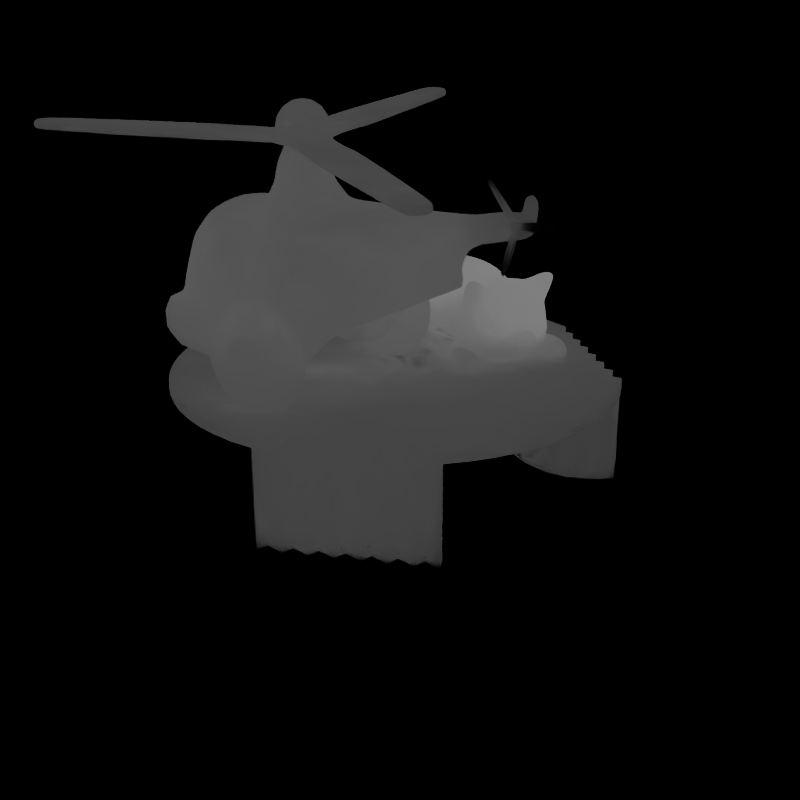} &
\includegraphics[width=\width]{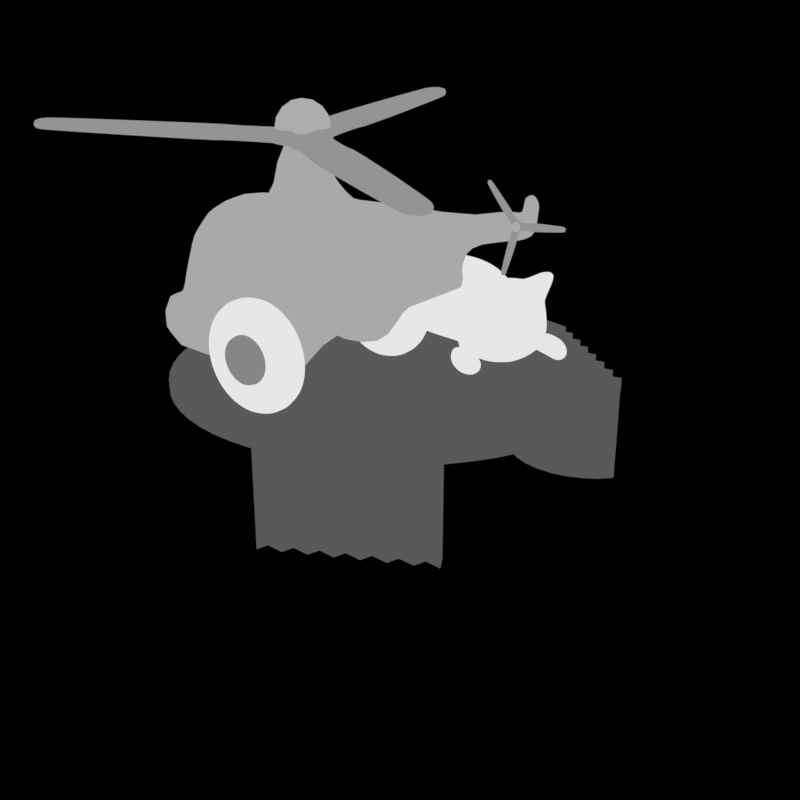}  \\
\includegraphics[width=\width]{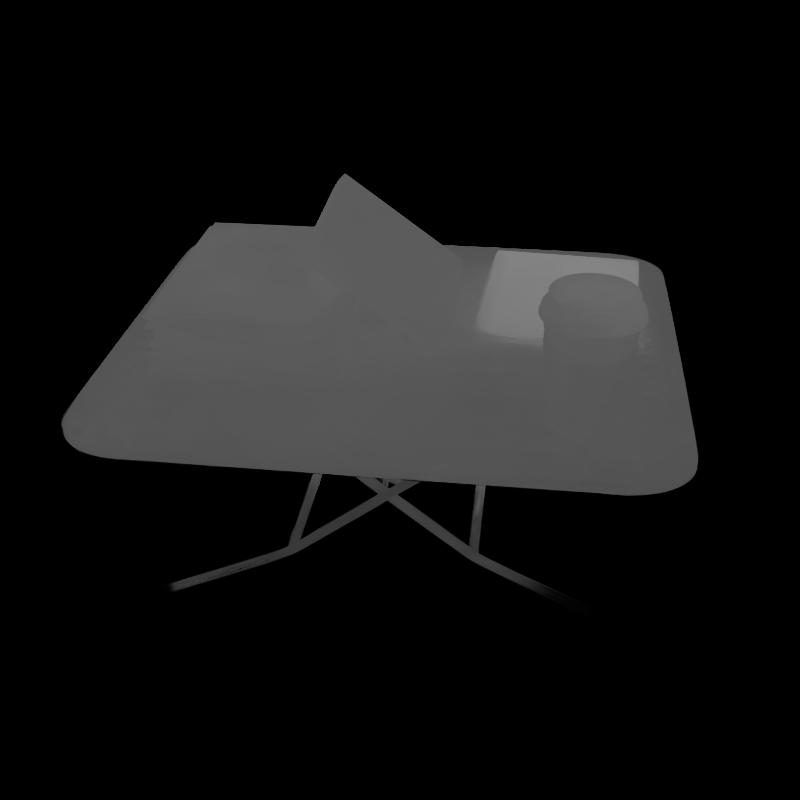} &
\includegraphics[width=\width]{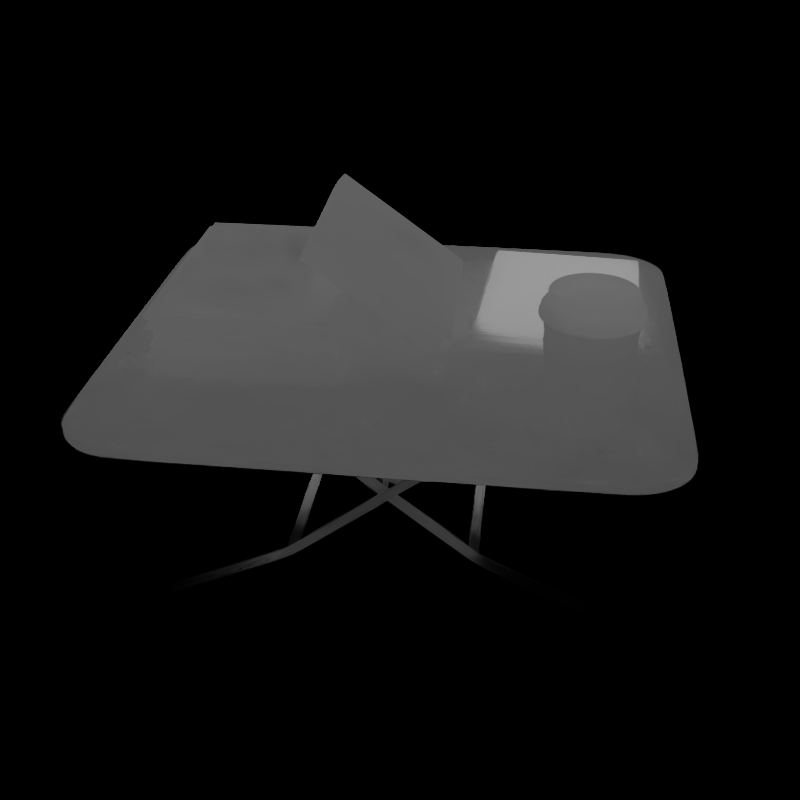} &
\includegraphics[width=\width]{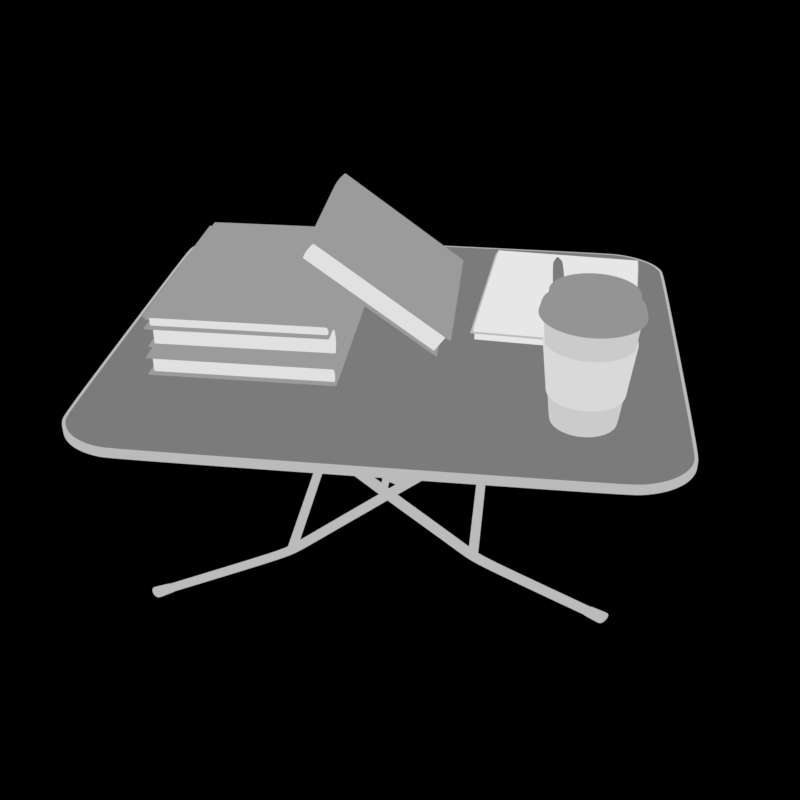} \\
\includegraphics[width=\width]{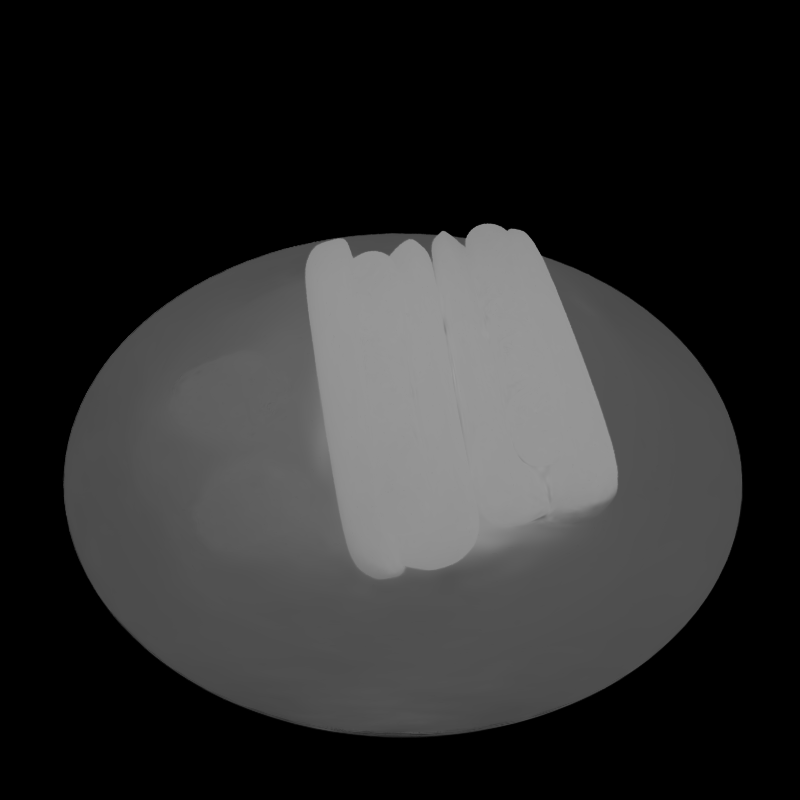} &
\includegraphics[width=\width]{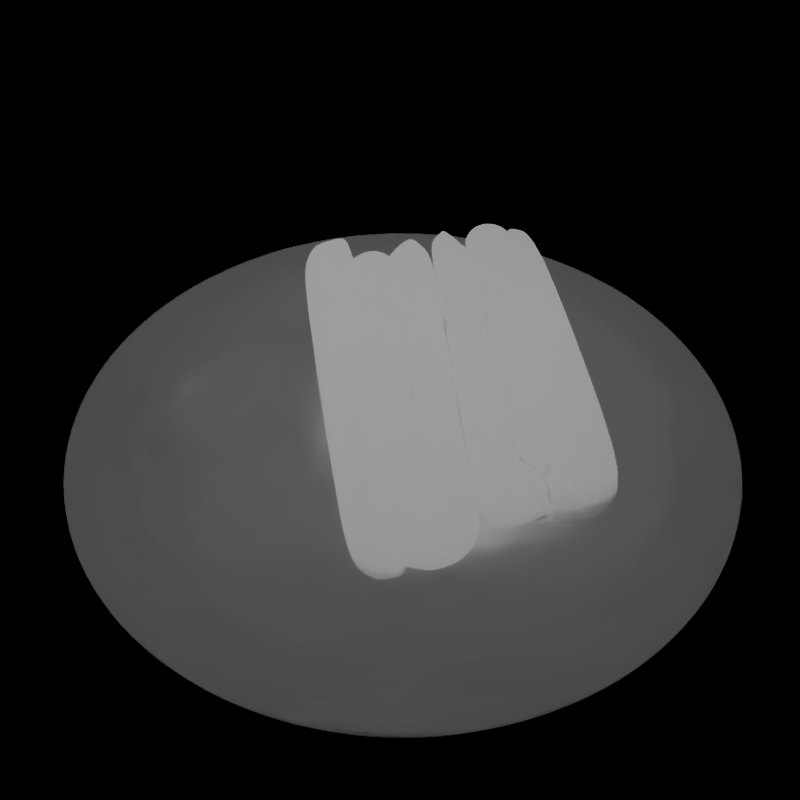} &
\includegraphics[width=\width]{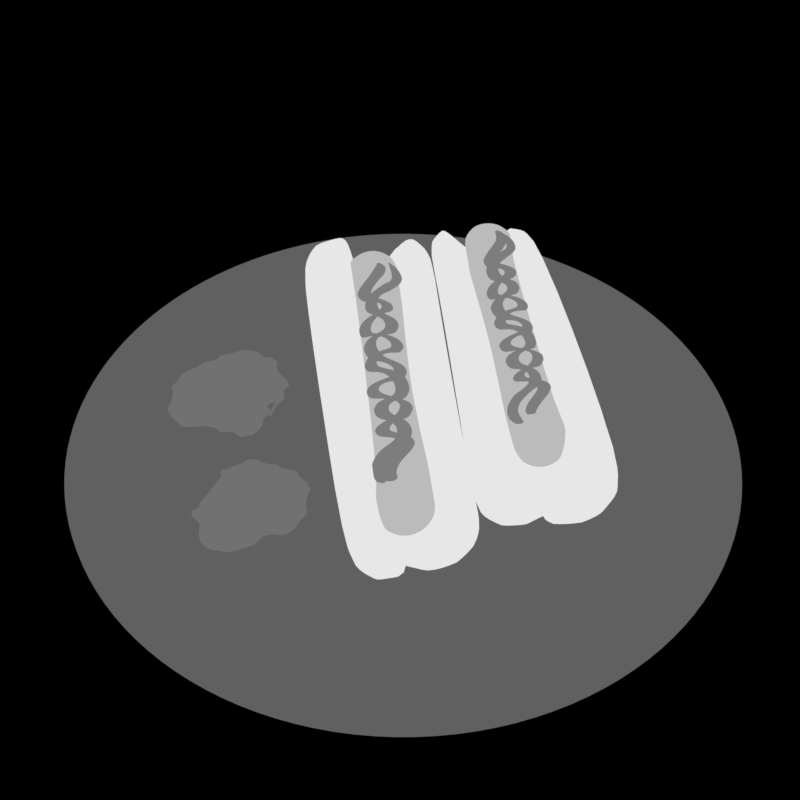}
\end{tabular}

\caption{Qualitative results of roughness estimation. Our method is better in \textit{counter} and similar to IRGS in other scenes. However, roughness estimation still remains challenging overall.}
\label{fig:inverse_roughness}

\end{figure}

\setlength\tabcolsep{\oldtabcolsep}

\providelength\width
\setlength\width{5cm}

\providelength\oldtabcolsep
\setlength{\oldtabcolsep}{\tabcolsep}
\setlength{\tabcolsep}{1pt}

\begin{figure*}[t]
\centering
\footnotesize

\begin{tabular}{ccc}
IRGS & Ours & GT \\
\midrule
\includegraphics[width=\width]{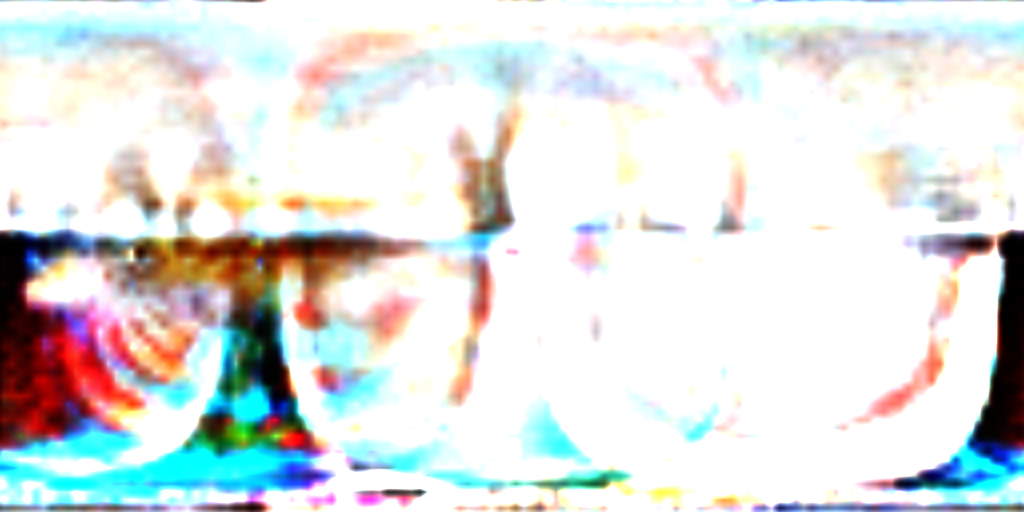} &
\includegraphics[width=\width]{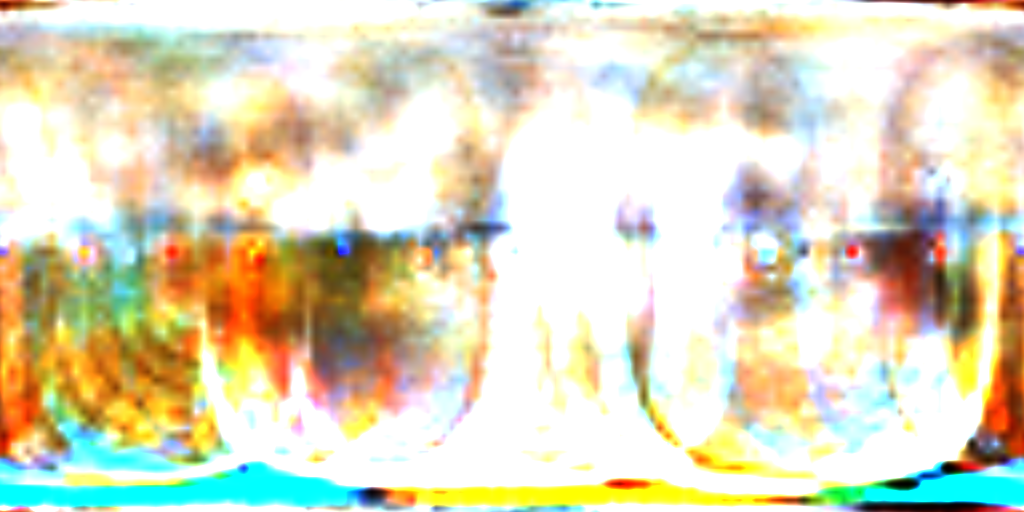} &
\includegraphics[width=\width]{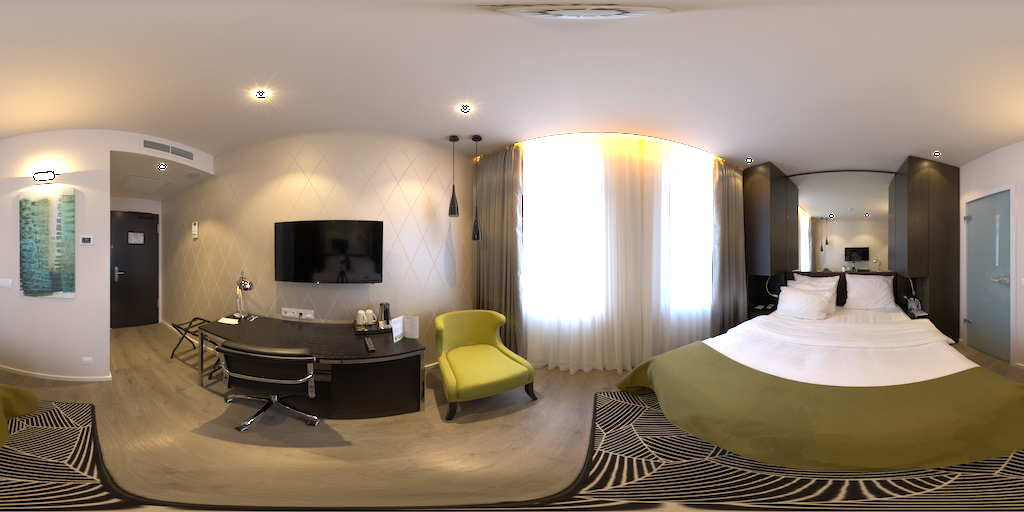} \\
\includegraphics[width=\width]{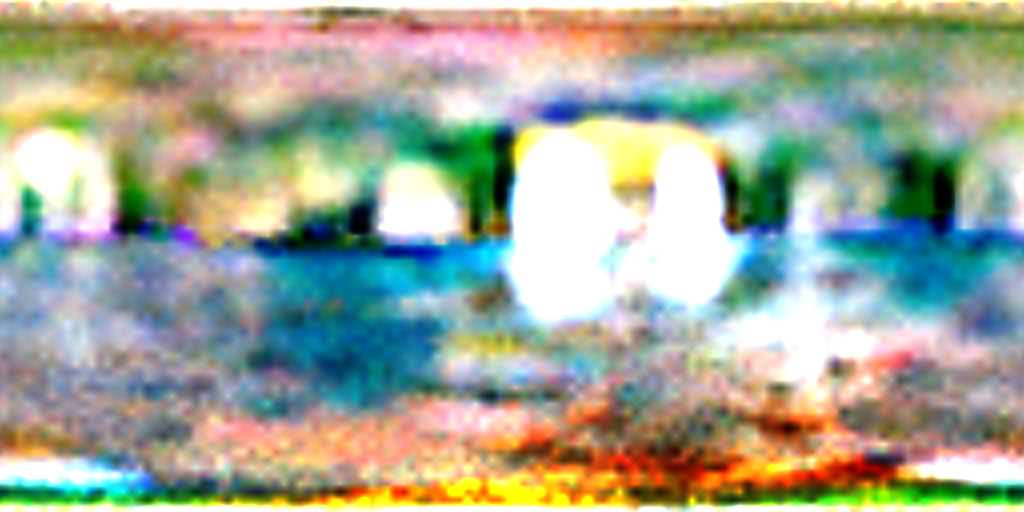} &
\includegraphics[width=\width]{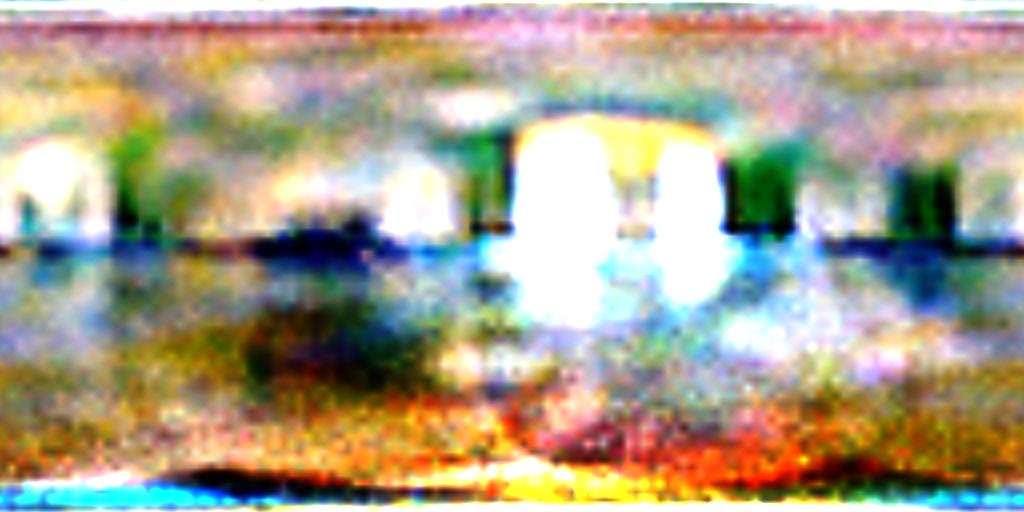} &
\includegraphics[width=\width]{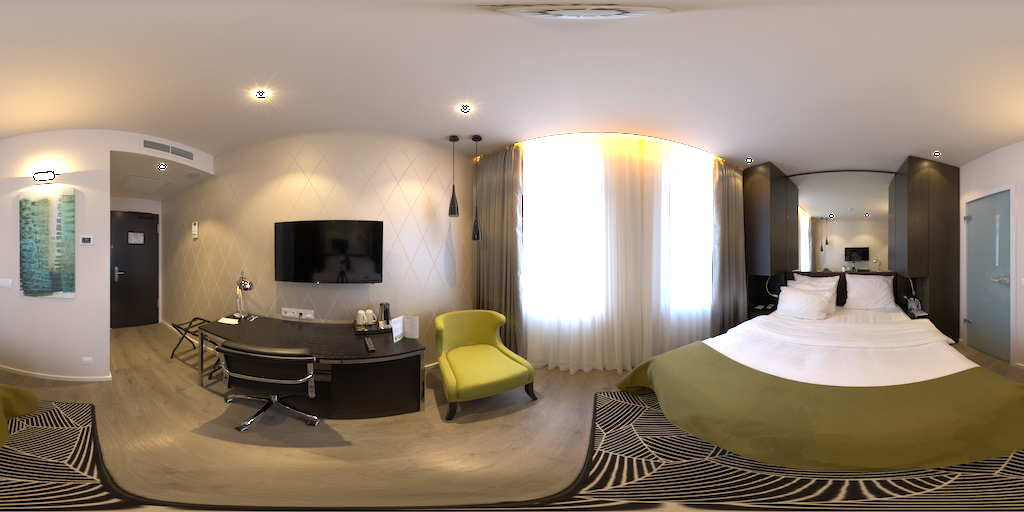}  \\
\includegraphics[width=\width]{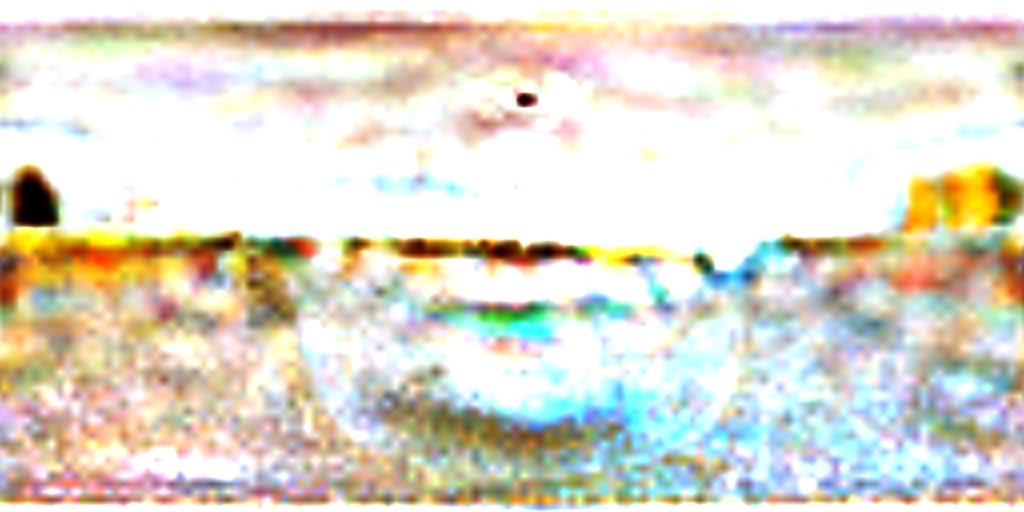} &
\includegraphics[width=\width]{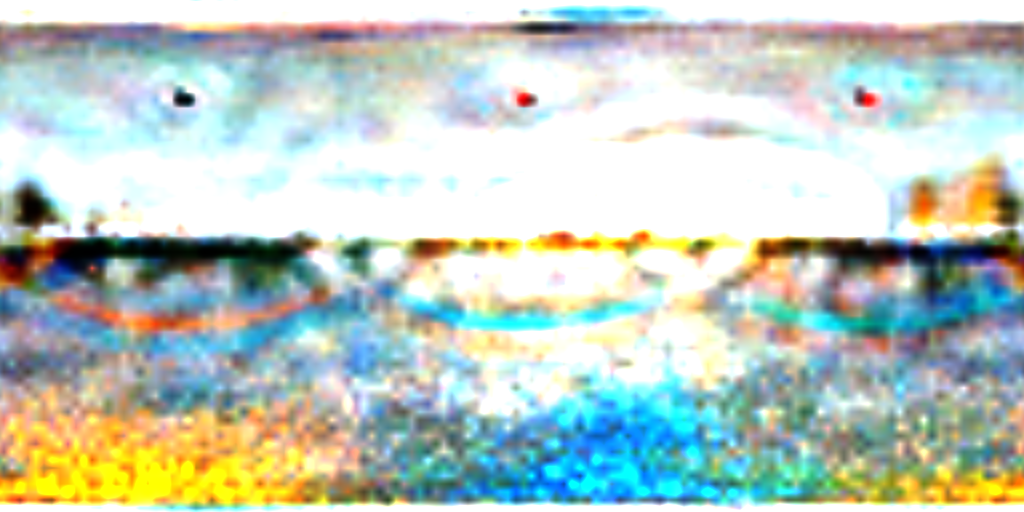} &
\includegraphics[width=\width]{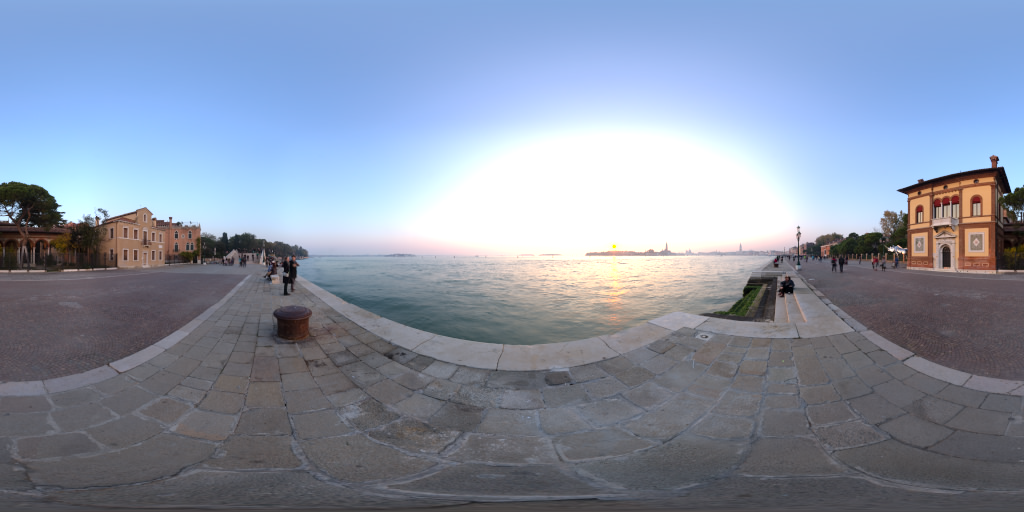} \\
\includegraphics[width=\width]{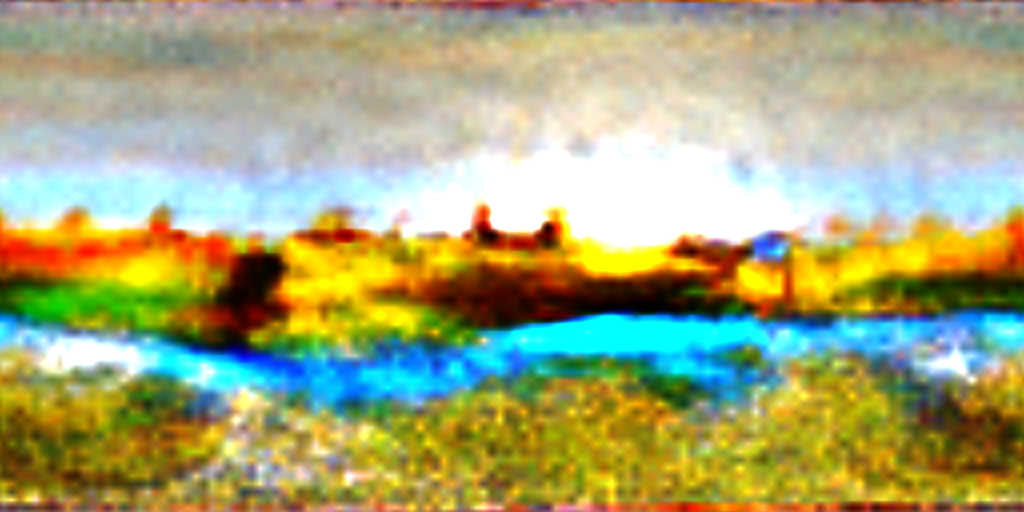} &
\includegraphics[width=\width]{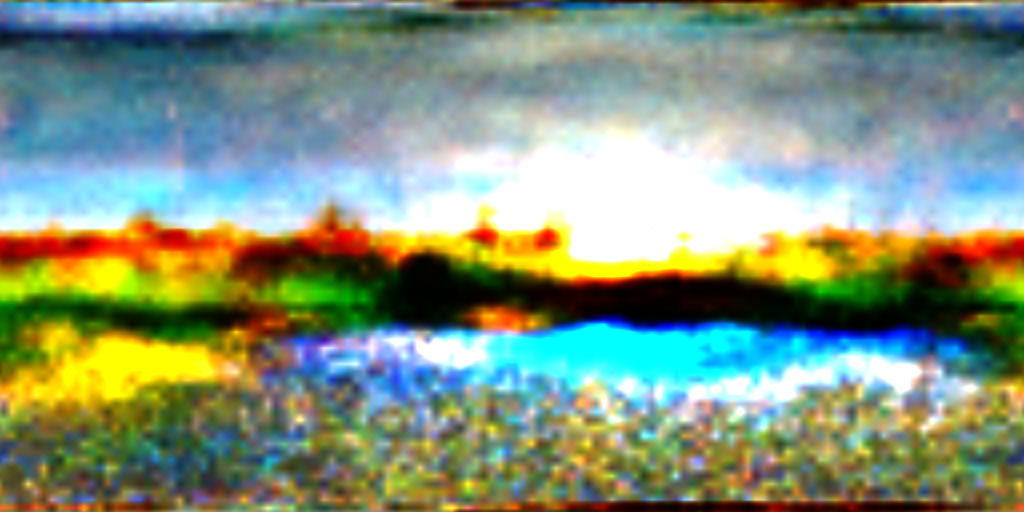} &
\includegraphics[width=\width]{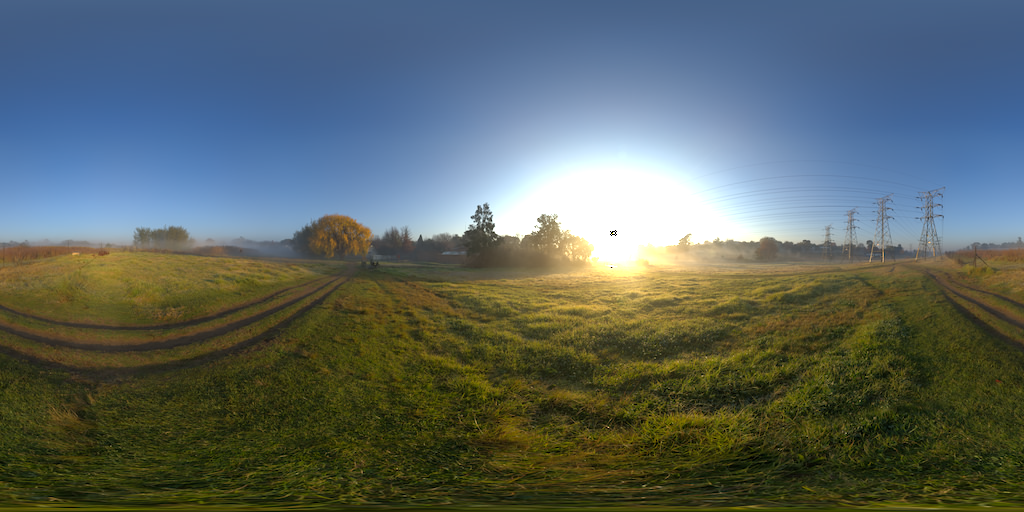}
\end{tabular}

\caption{Qualitative results of environment light estimation. Top to bottom: \textit{counter}, \textit{toy}, \textit{table}, \textit{hotdog}. Note that the problem is still extremely ill-posed without  direct observations from very specular surfaces. Our method give overall closer results, such as the color of the sky.}
\label{fig:inverse_env}

\end{figure*}

\setlength\tabcolsep{\oldtabcolsep}

\section{Relighting}

\providelength\width
\setlength\width{2.5cm}

\providelength\oldtabcolsep
\setlength{\oldtabcolsep}{\tabcolsep}
\setlength{\tabcolsep}{1pt}

\begin{figure*}[t]
\centering
\footnotesize

\begin{tabular}{ccccccc}
IRGS & Ours & GT & & IRGS & Ours & GT \\
\midrule
\includegraphics[width=\width]{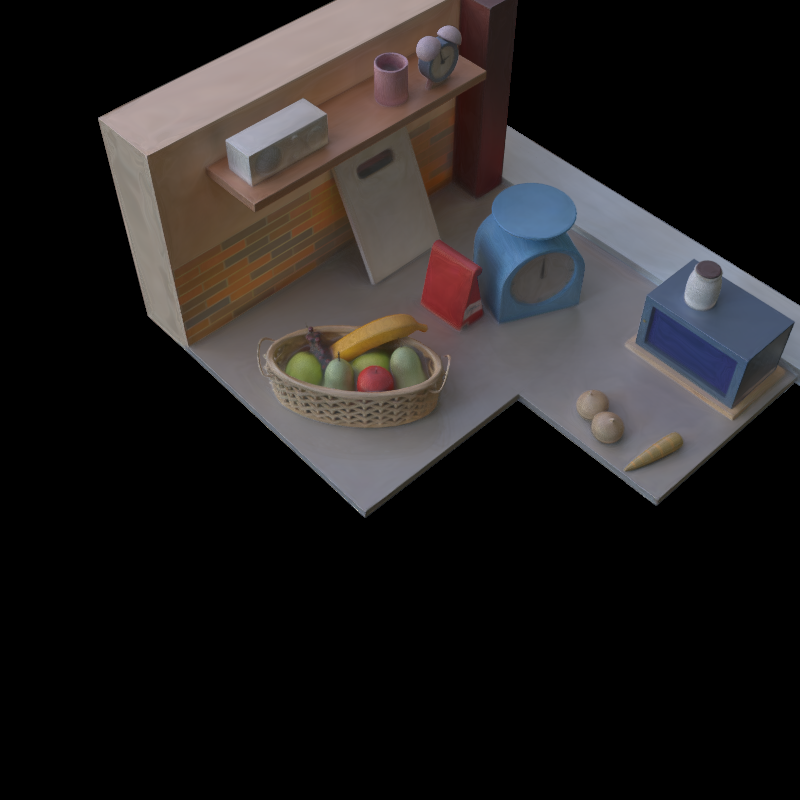} &
\includegraphics[width=\width]{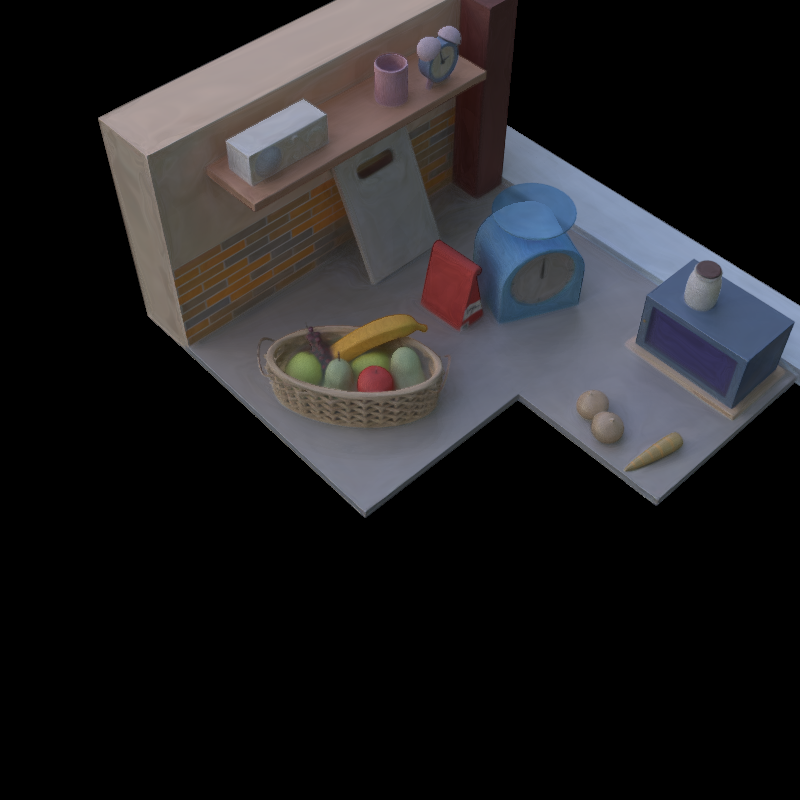} &
\includegraphics[width=\width]{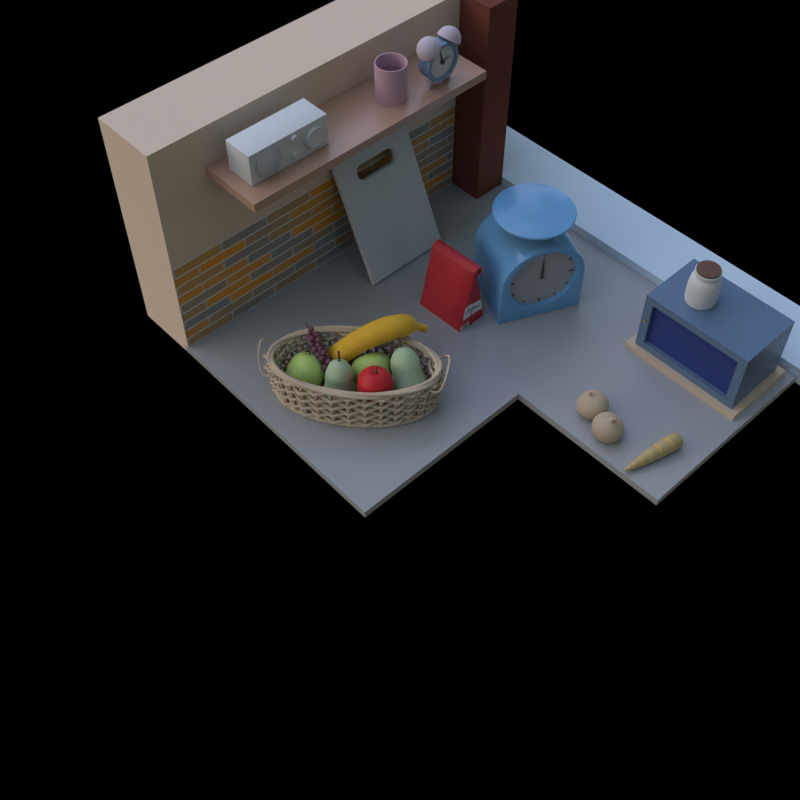} &
&
\includegraphics[width=\width]{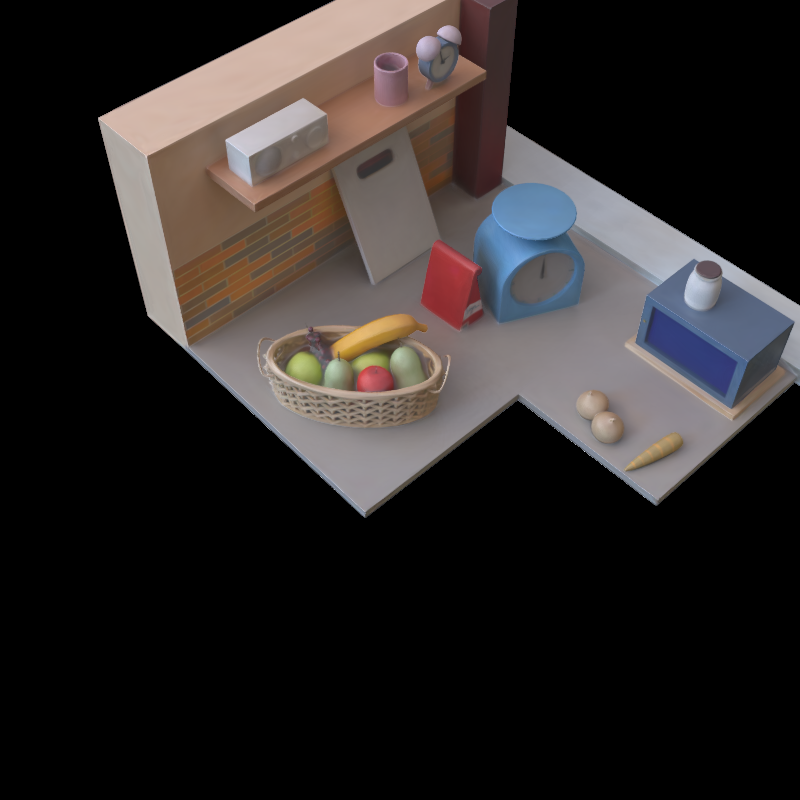} &
\includegraphics[width=\width]{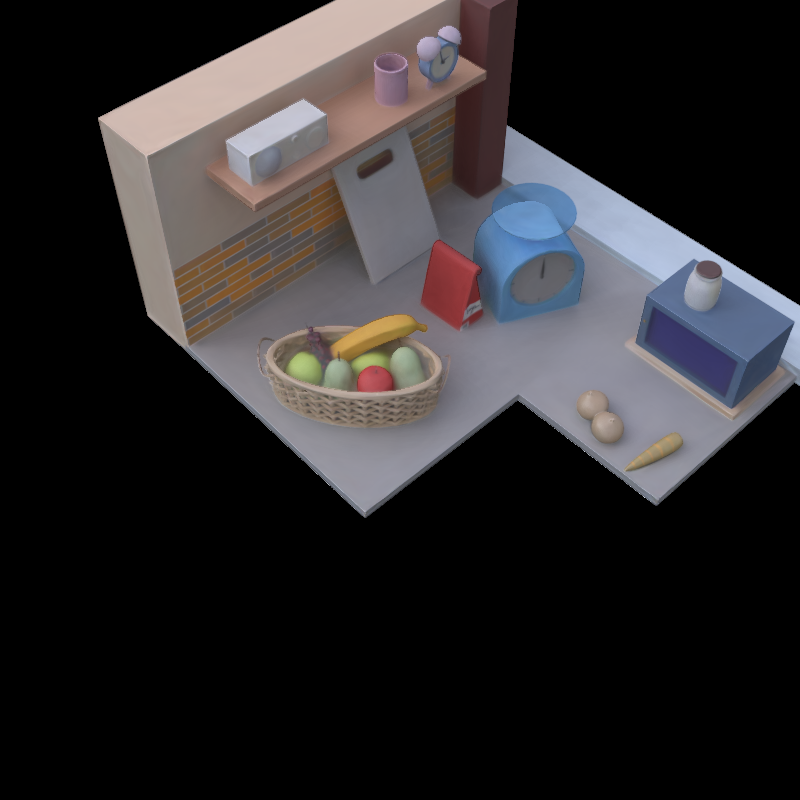} &
\includegraphics[width=\width]{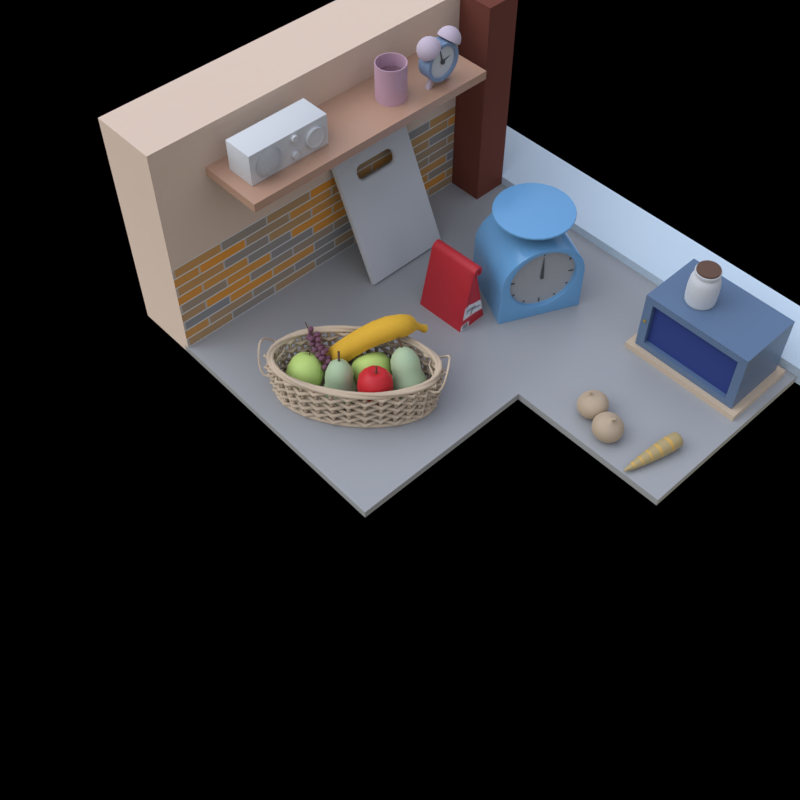} \\
\includegraphics[width=\width]{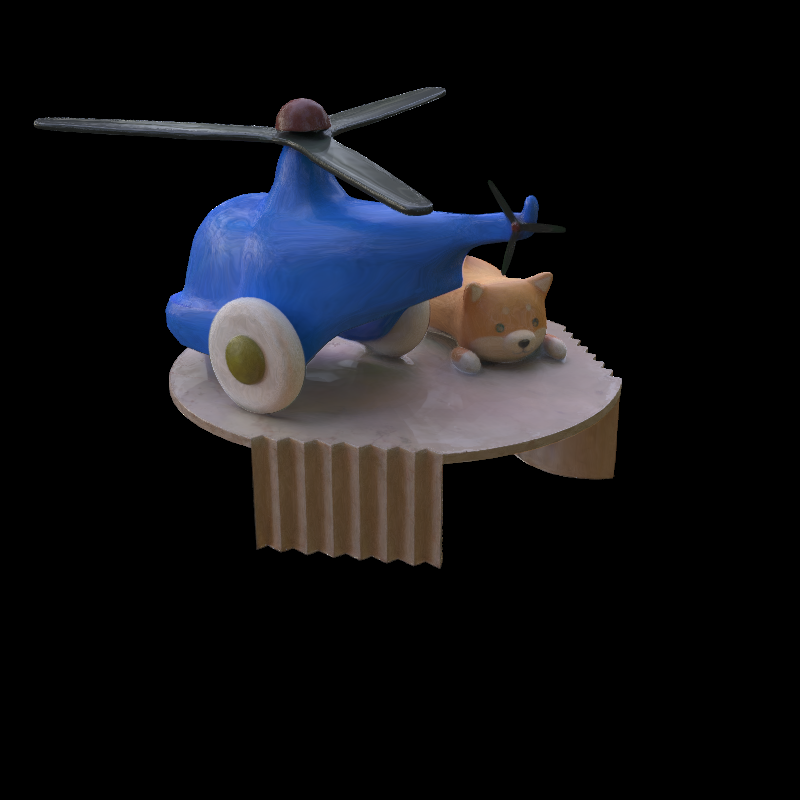} &
\includegraphics[width=\width]{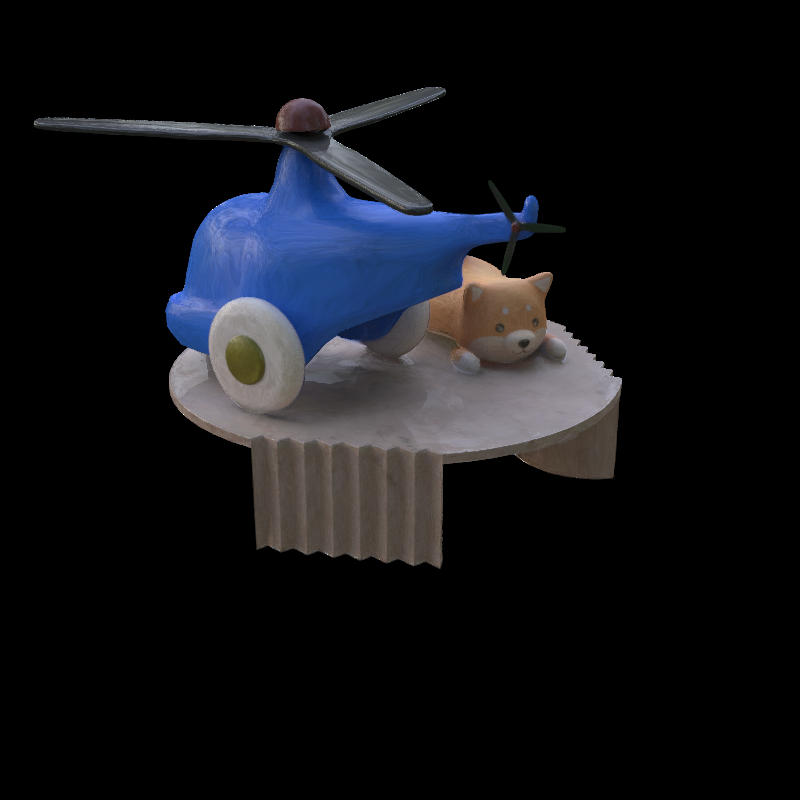} &
\includegraphics[width=\width]{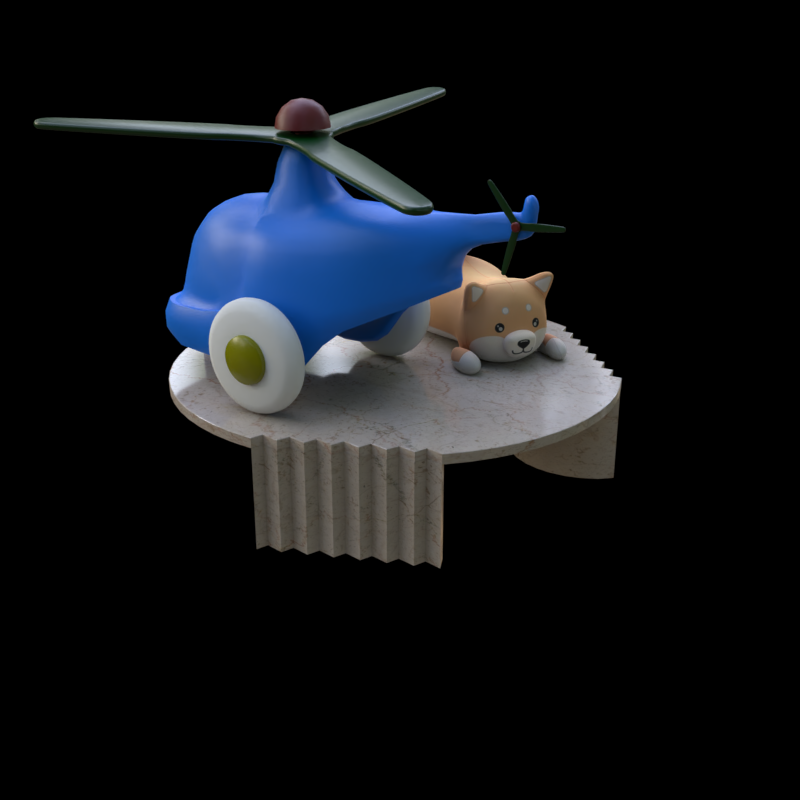} &
&
\includegraphics[width=\width]{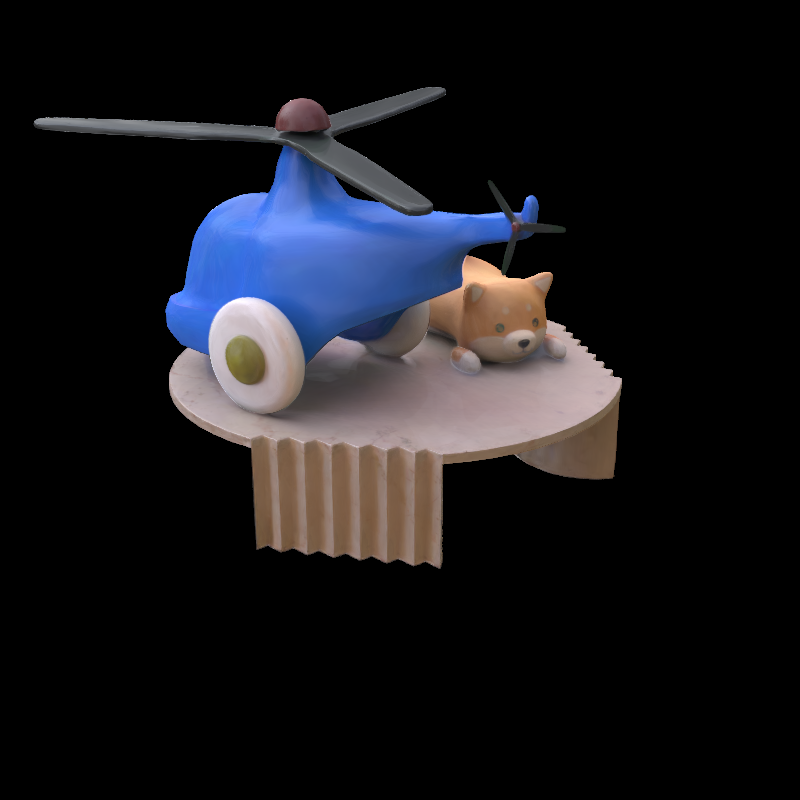} &
\includegraphics[width=\width]{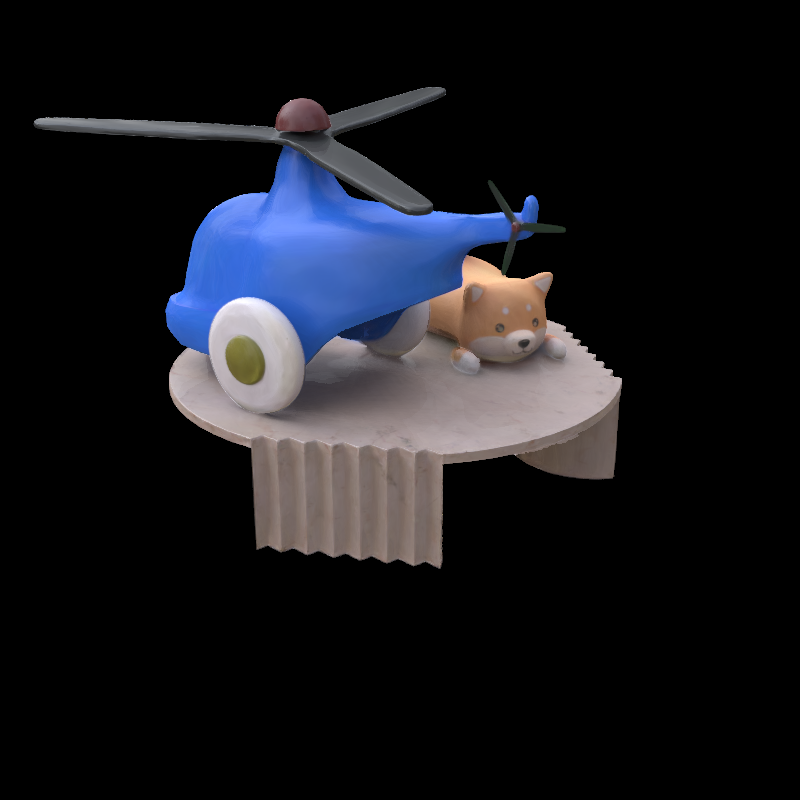} &
\includegraphics[width=\width]{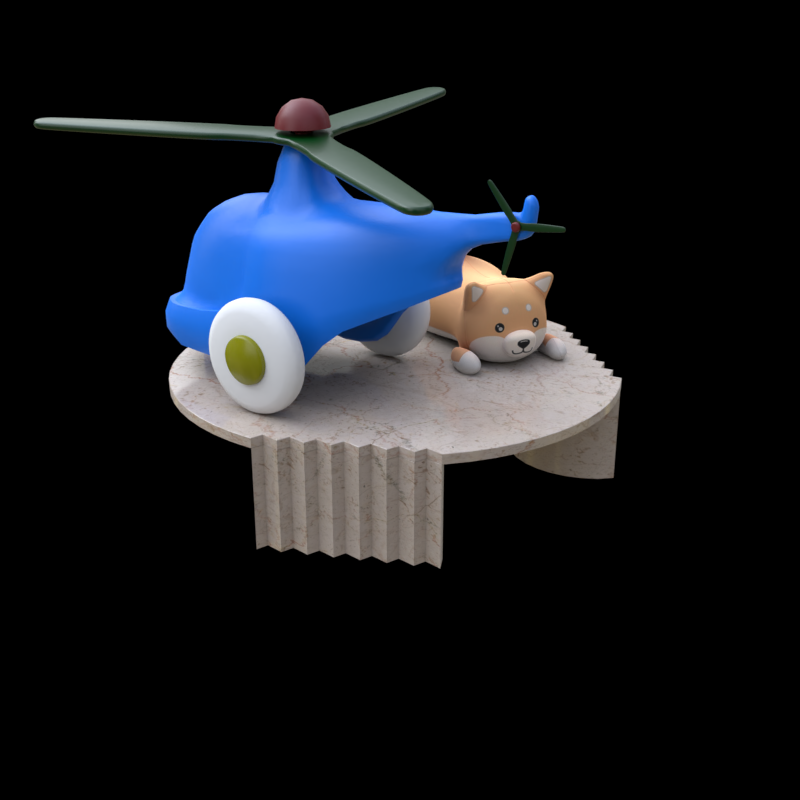} \\
\includegraphics[width=\width]{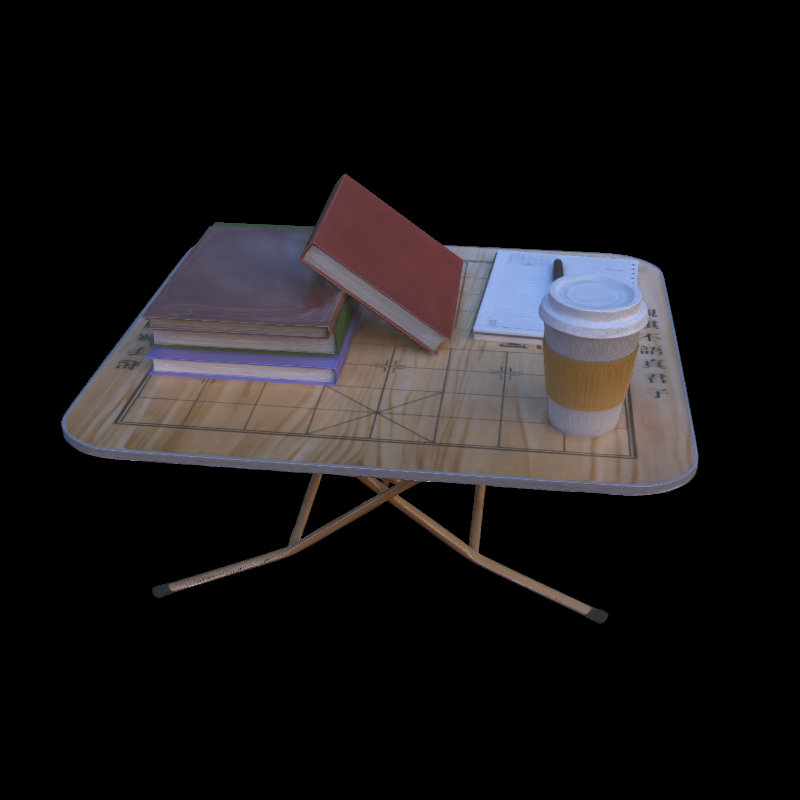} &
\includegraphics[width=\width]{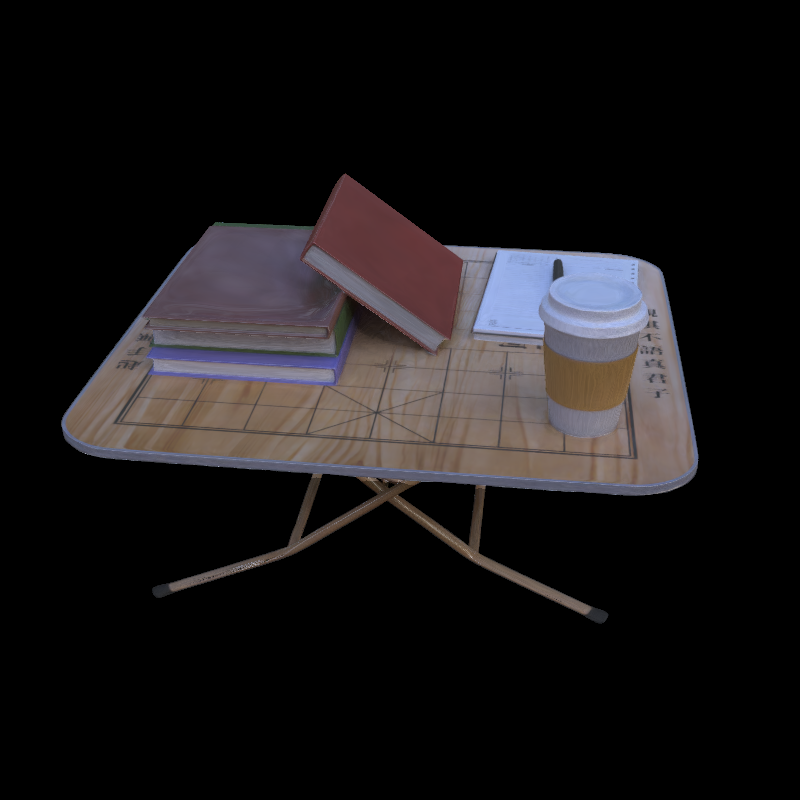} &
\includegraphics[width=\width]{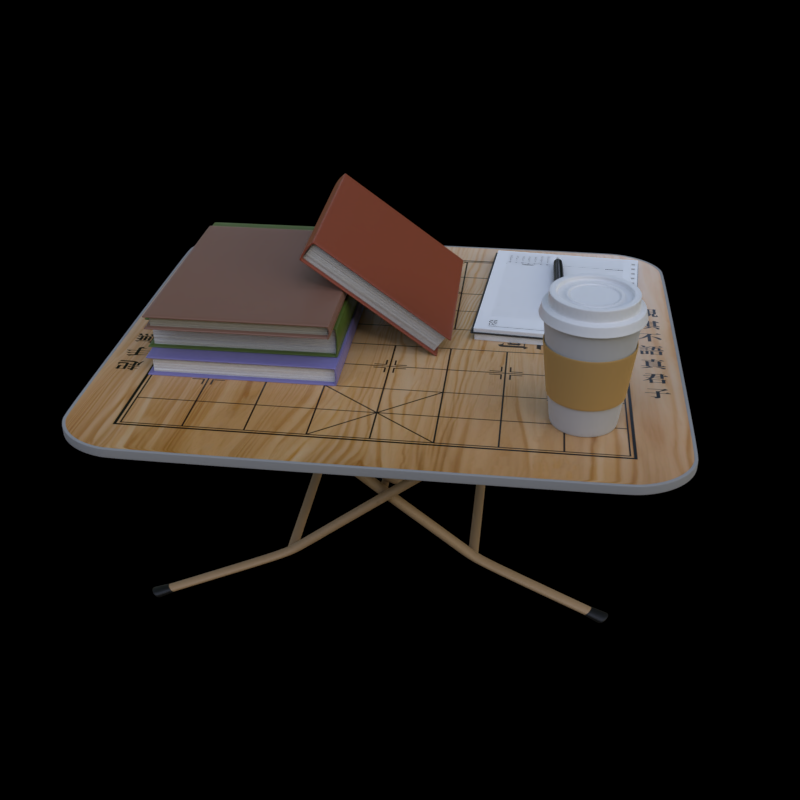} &
&
\includegraphics[width=\width]{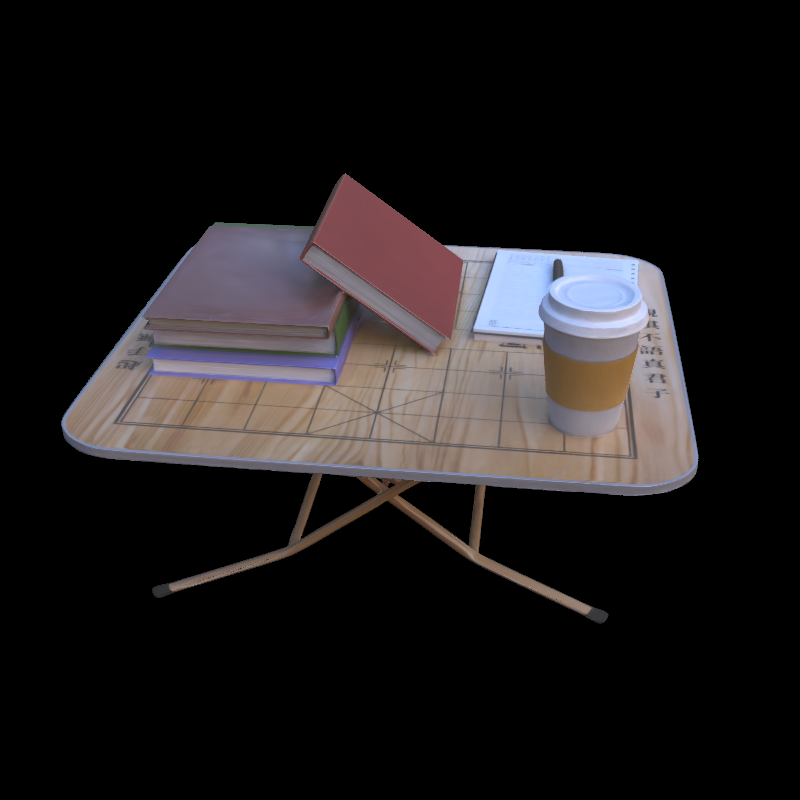} &
\includegraphics[width=\width]{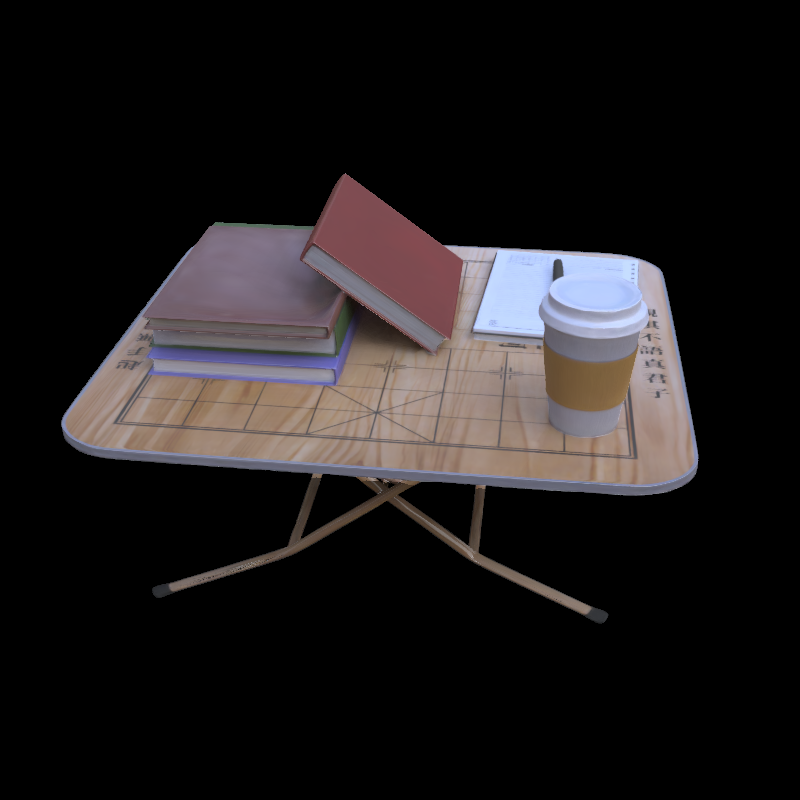} &
\includegraphics[width=\width]{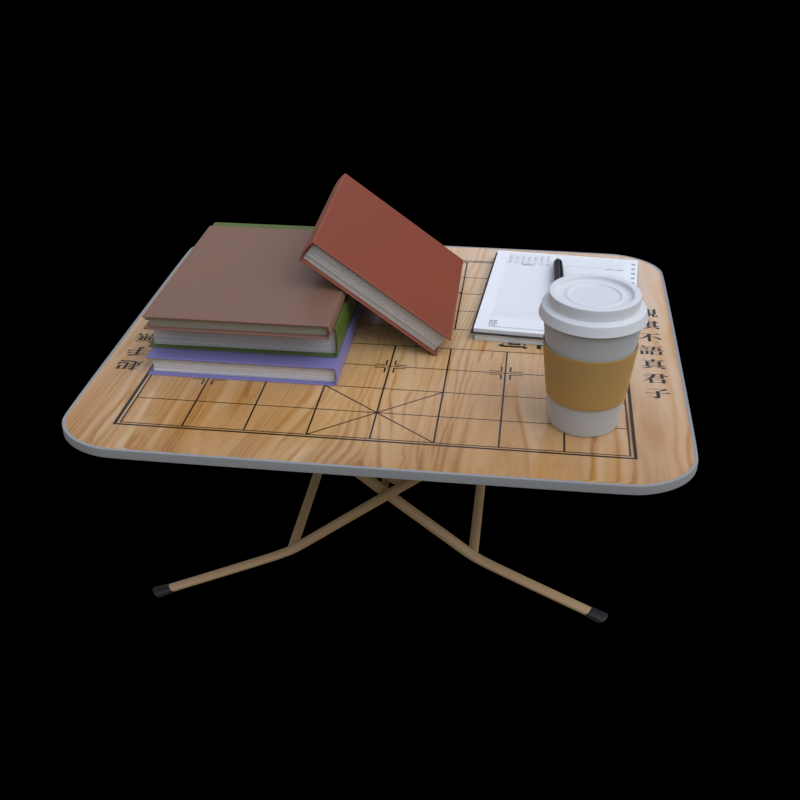} \\
\includegraphics[width=\width]{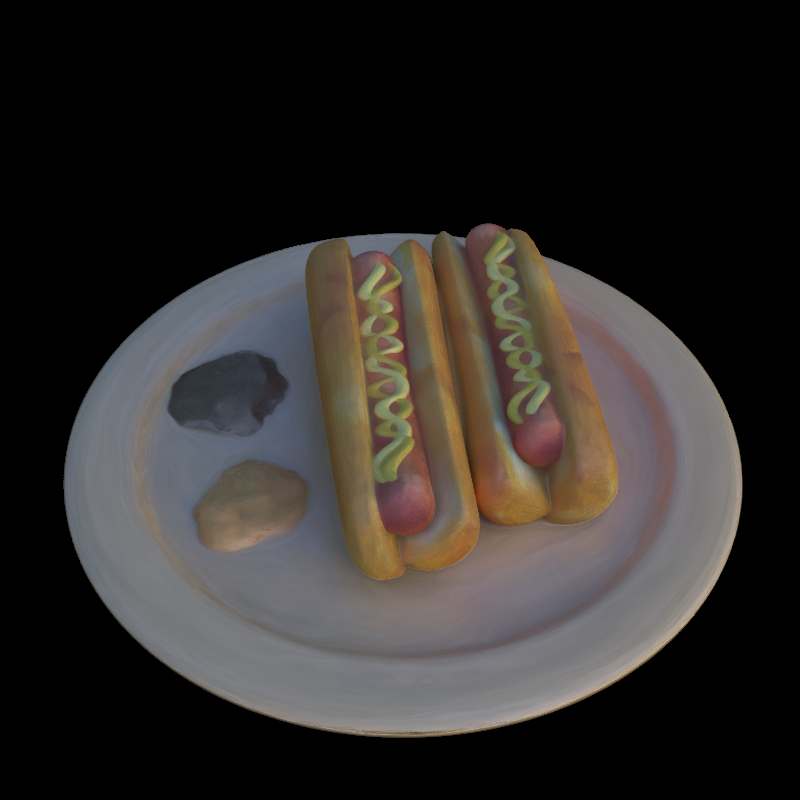} &
\includegraphics[width=\width]{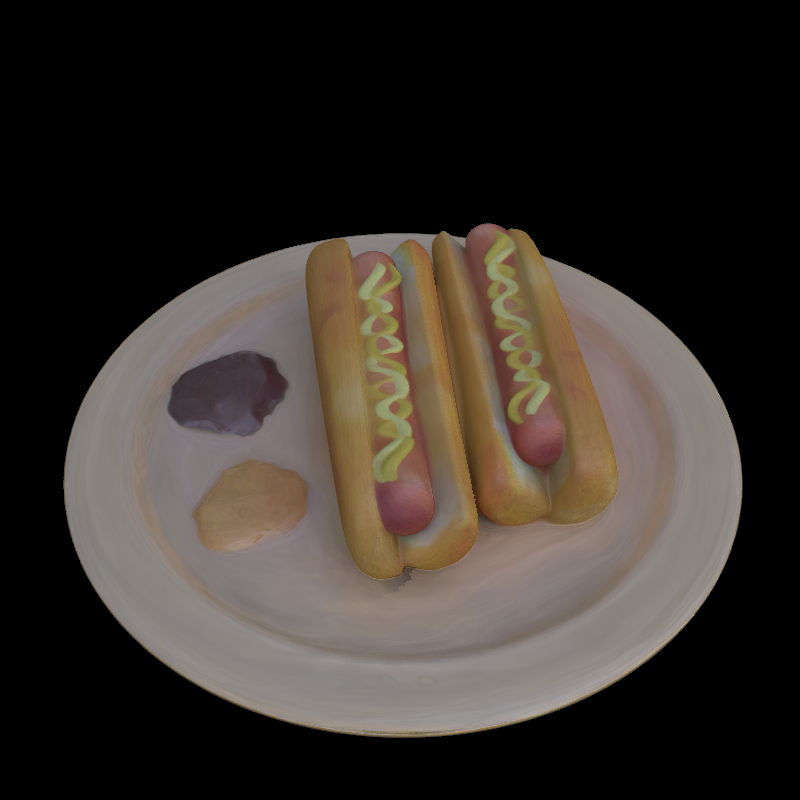} &
\includegraphics[width=\width]{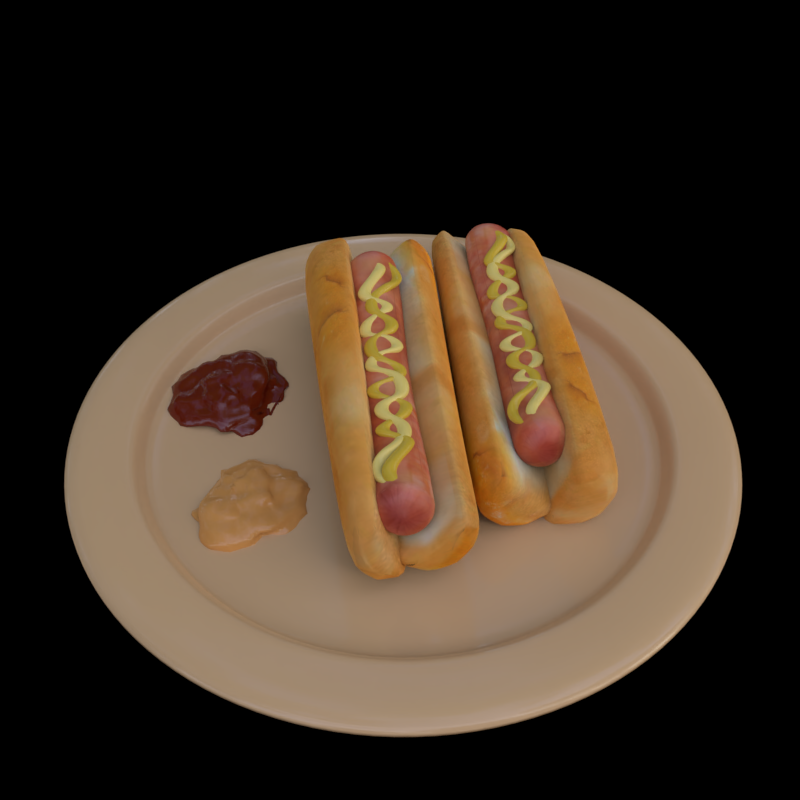} &
&
\includegraphics[width=\width]{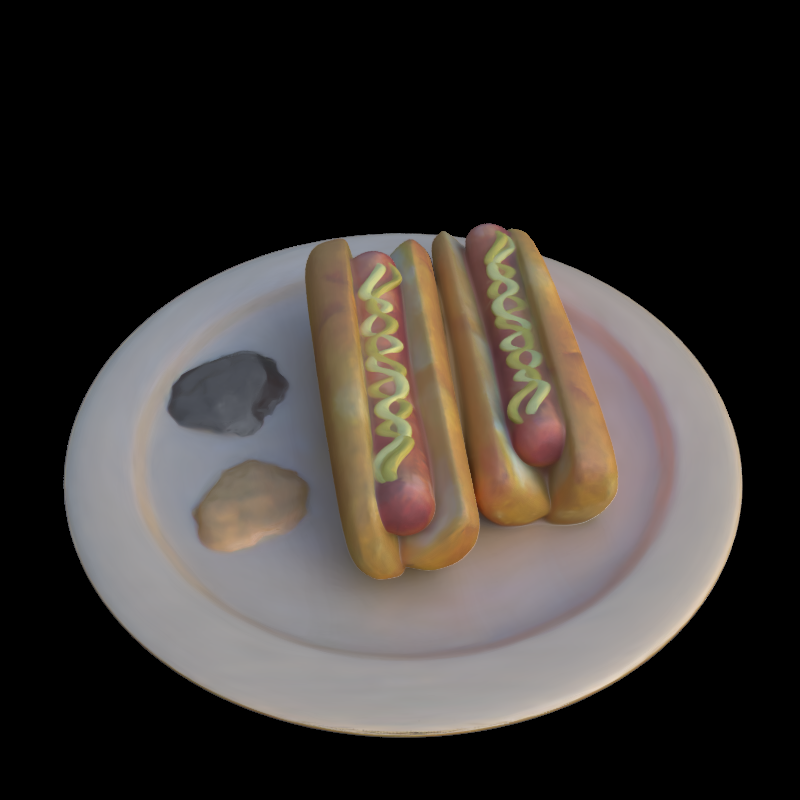} &
\includegraphics[width=\width]{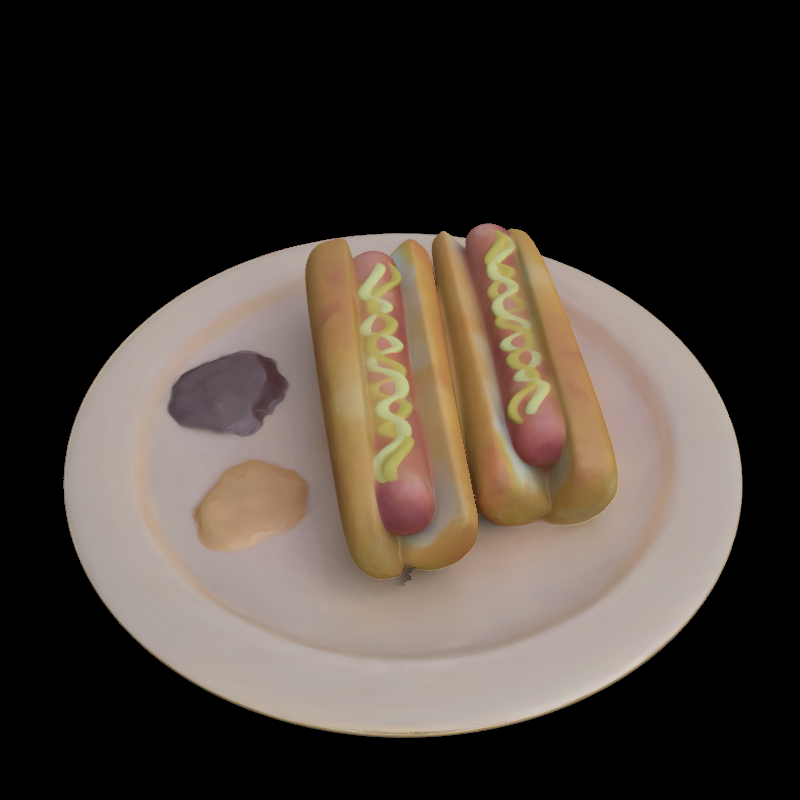} &
\includegraphics[width=\width]{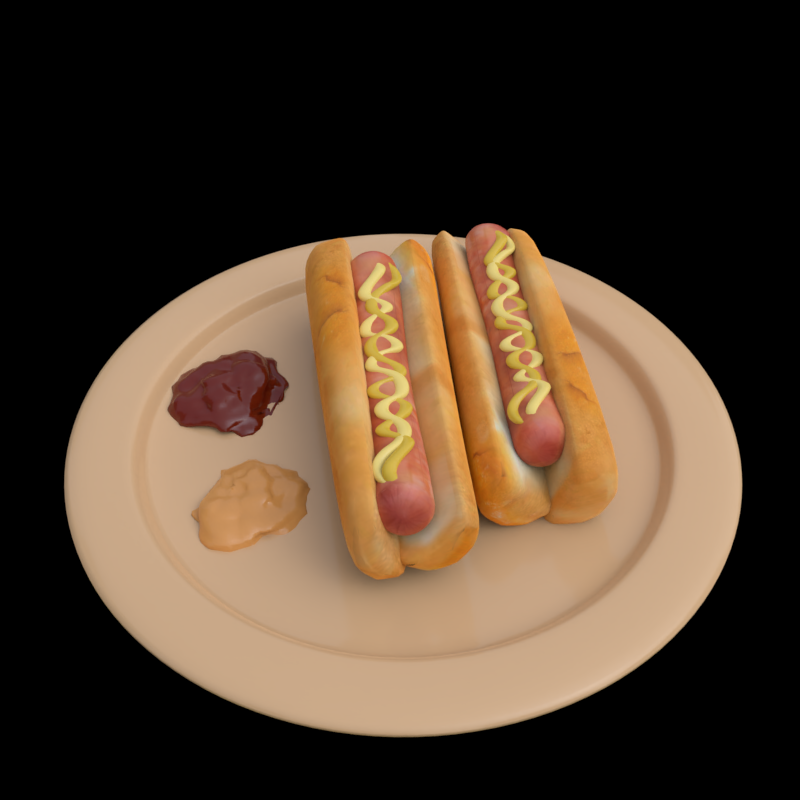} \\
\end{tabular}

\caption{Relighting comparison. We demonstrate relighting results using our recovered scene parameters and compare them with IRGS. Our method produces colors that are closer to ground truth.}
\label{fig:inverse_relight}

\end{figure*}

\setlength\tabcolsep{\oldtabcolsep}

We follow the algorithm in IRGS~\cite{gu2024IRGS} to render the recovered models with different environment maps and show the results in Figure~\ref{fig:inverse_relight}.
Overall, our method produces colors that are closer to ground truth.

\section{Known Geometry}

We experiment our method with ground truth geometry in Figure \ref{fig:inverse_gtgeo}. Our method benefits from accurate details in recovered geometry.

\providelength\width
\setlength\width{3.5cm}

\providelength\oldtabcolsep
\setlength{\oldtabcolsep}{\tabcolsep}
\setlength{\tabcolsep}{1pt}

\begin{figure*}[t]
\centering
\footnotesize

\begin{tabular}{cccc}
GT Geo. Single Light & GT Geo. Ours & Ours & GT \\
\midrule
\includegraphics[width=\width]{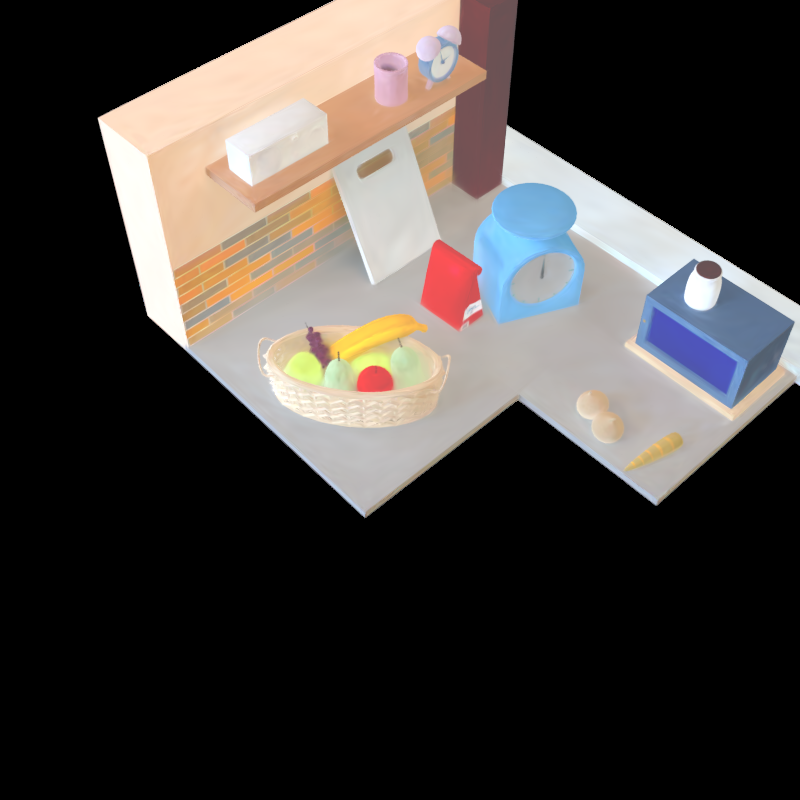} &
\includegraphics[width=\width]{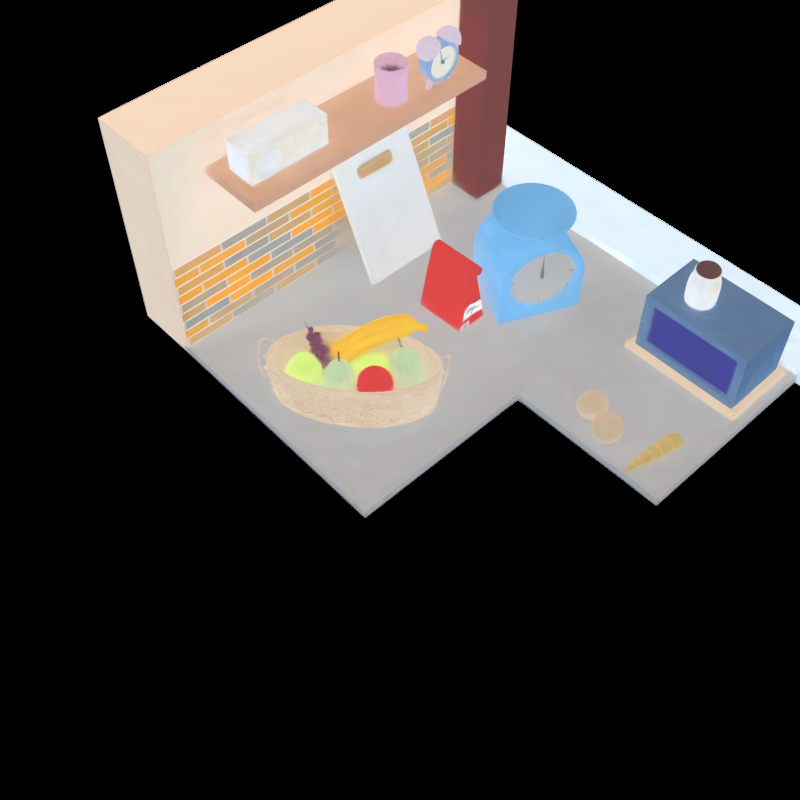} &
\includegraphics[width=\width]{image/albedo/counter/108_ours.png} &
\includegraphics[width=\width]{image/albedo/counter/108_gt.png} \\
\includegraphics[width=\width]{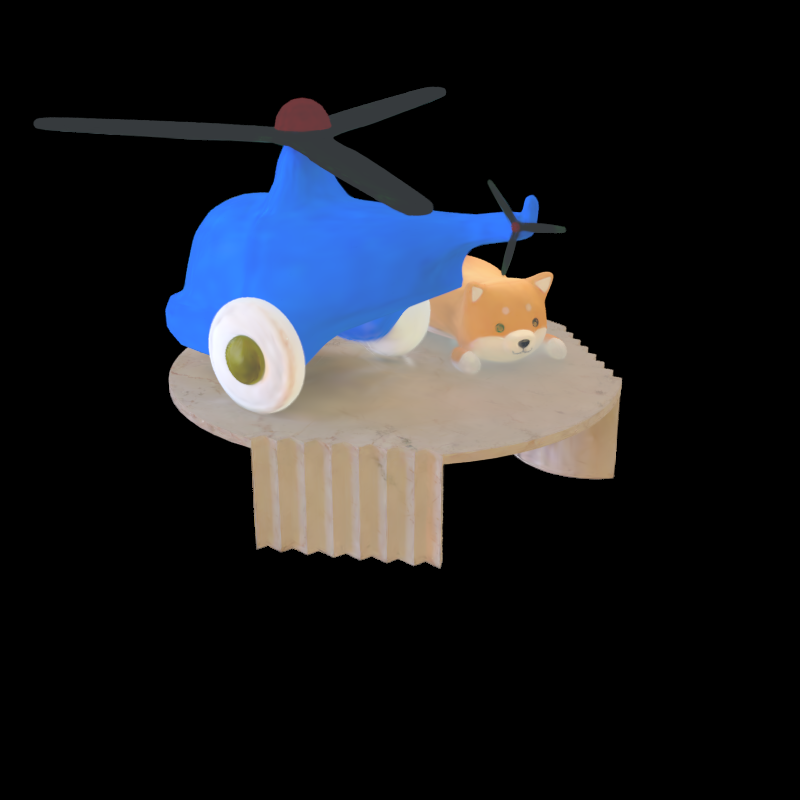} &
\includegraphics[width=\width]{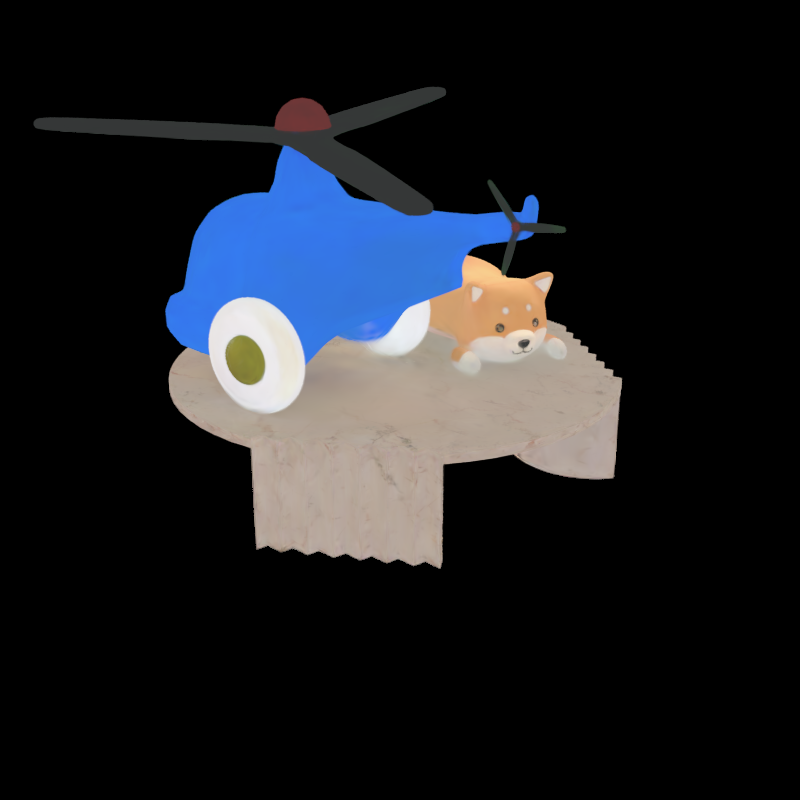} &
\includegraphics[width=\width]{image/albedo/toy/036_ours.png} &
\includegraphics[width=\width]{image/albedo/toy/036_gt.png} \\
\includegraphics[width=\width]{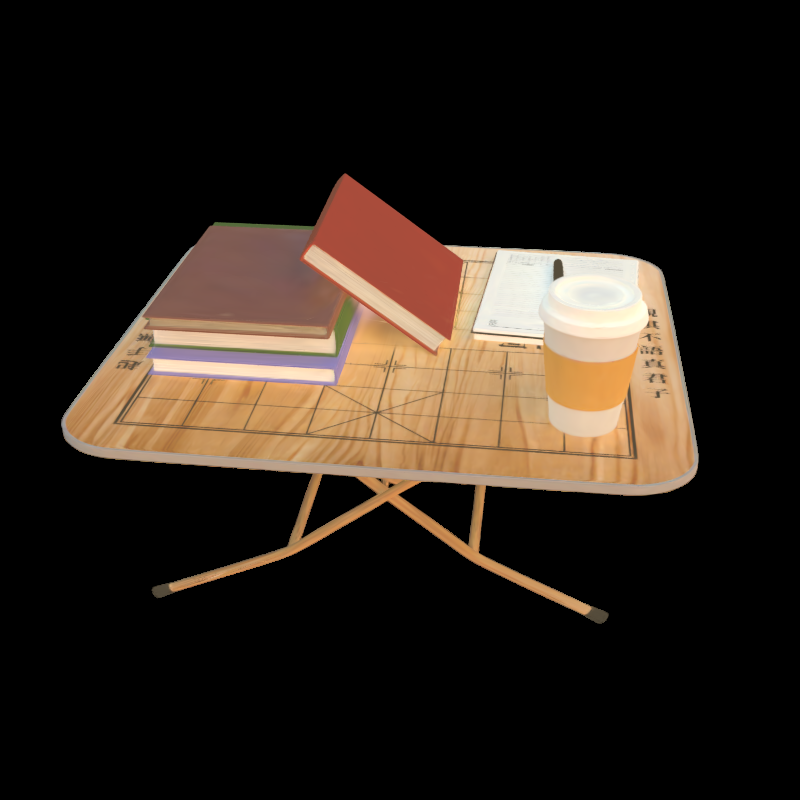} &
\includegraphics[width=\width]{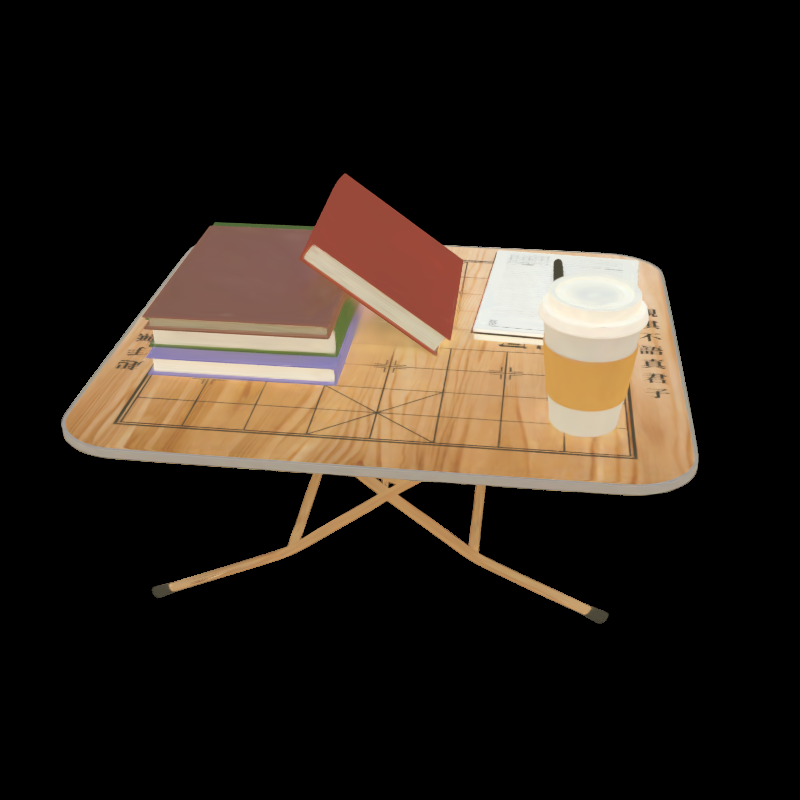} &
\includegraphics[width=\width]{image/albedo/table/057_ours.png} &
\includegraphics[width=\width]{image/albedo/table/057_gt.png} \\
\includegraphics[width=\width]{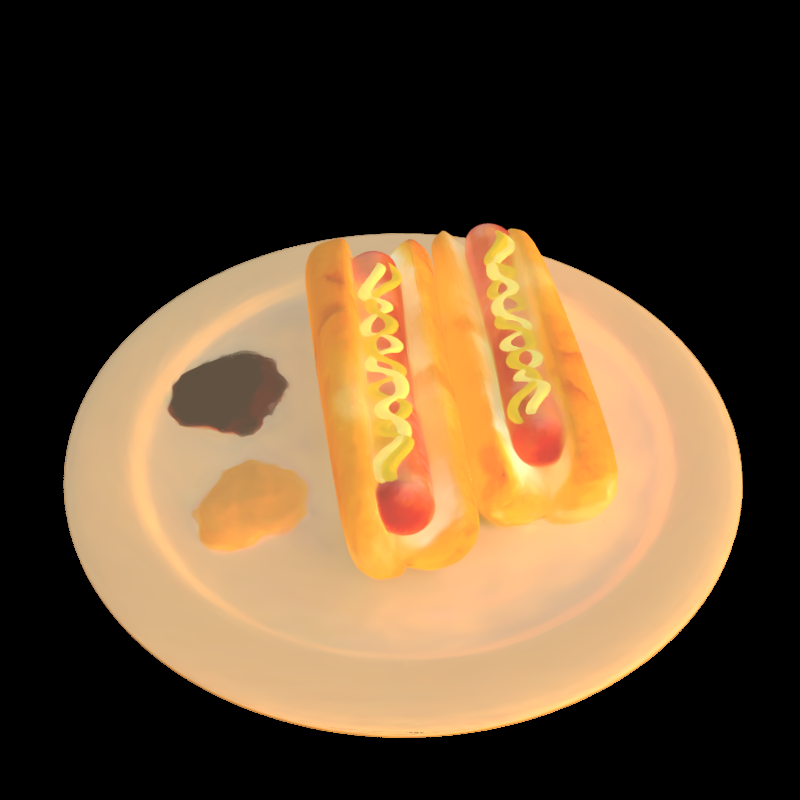} &
\includegraphics[width=\width]{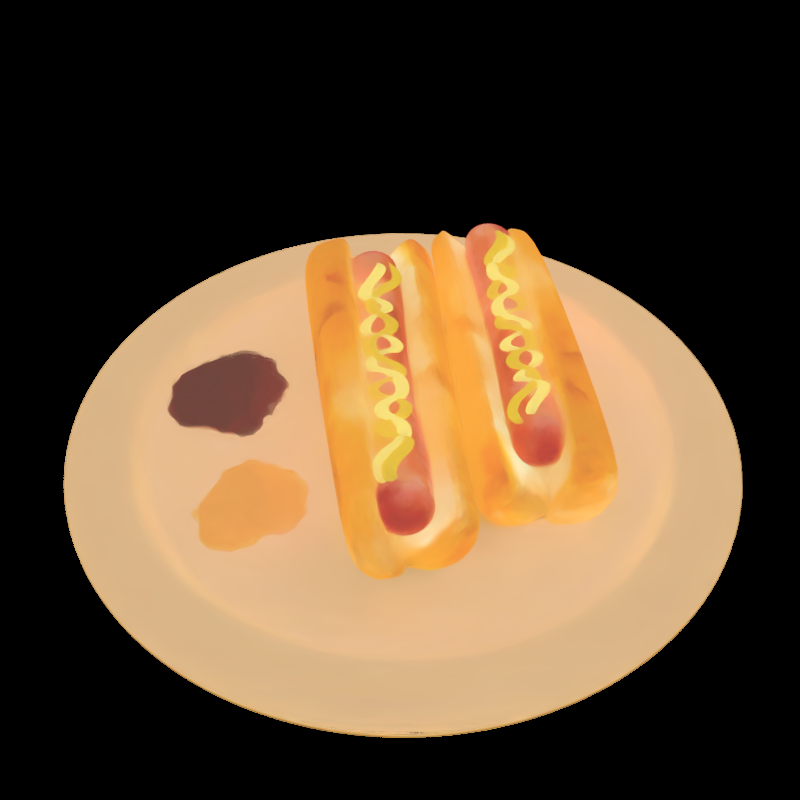} &
\includegraphics[width=\width]{image/albedo/hotdog/093_ours.png} &
\includegraphics[width=\width]{image/albedo/hotdog/093_gt.png} \\
\end{tabular}

\caption{Known geometry experiments. The first two columns are experiments run with known (GT) geometry, with single light and RotLight setups respectively. \textbf{Column 1, 2:} we show that the RotLight capture setup and other proposed components is effective with known geometry. \textbf{Column 2, 3:} the performance of our method is limited by inaccuracies in geometry, such as under the onions in \textit{counter}, under the dog in \textit{toy}, the top of the book pile in \textit{table}, and under the hotdogs in \textit{hotdog}.}
\label{fig:inverse_gtgeo}

\end{figure*}

\setlength\tabcolsep{\oldtabcolsep}

\section{Model and Training Details}

Each radiance cache MLP has two hidden layers of width 256. The inputs to an MLP are position and direction vectors. The position is encoded with the hash grid proposed in InstantNGP~\cite{muller2022instantngp}, and the direction is encoded with the NeRF positional encoding~\cite{Mildenhall2021NERF}. The output of the MLP is a 3-channel color.

In radiance cache pretraining, a random input view is chosen at each step, and the MLP for the corresponding light angle is optimized at the step. A ray is traced for every pixel to determine a 3D location to query the MLP for the radiance towards the camera. The MSE loss between the ground truth and the queried radiance is minimized.

\end{document}